\PassOptionsToPackage{table}{xcolor}

\documentclass{article}

\usepackage[sort,numbers]{natbib}
\usepackage[linesnumbered,ruled,vlined]{algorithm2e}

\newif\ifjournal
\journaltrue      

\ifjournal
    \usepackage[final,main]{neurips_2026}
\else
    \usepackage[margin=1in]{geometry}
    \usepackage{authblk}
    \usepackage{lmodern}
\fi

\usepackage{customize}

\def \figdir {.}

\newcommand{\mbf}[1]{\mathbf{#1}}              
\newcommand{\mbif}[1]{\boldsymbol{#1}}         
\newcommand{\tens}[1]{\mbf{#1}}                
\newcommand{\gtens}[1]{\mbif{#1}}              
\newcommand{\vect}[1]{\mbif{#1}}               
\newcommand{\trace}{\mathsf{trace}}

\newcommand{\um}{\underline{m}}                
\DeclareMathOperator*{\supp}{\operatorname{supp}}

\DeclareMathOperator*{\Res}{Res}

\theoremstyle{definition}
\newtheorem*{example*}{Example}

\newcommand{\freealg}{freealg}
\newcommand{\bemph}[1]{\textbf{\emph{#1}}}

\def \ie {i.e.,\ }

\newcommand{\fulltitle}{Spectral Estimation with Deformed Decompression}

\title{\fulltitle}

\newcommand{\authrone}{Siavash Ameli}
\newcommand{\affilone}{ICSI and Department of Statistics}
\newcommand{\addrsone}{University of California, Berkeley}
\newcommand{\emailone}{sameli@berkeley.edu}

\newcommand{\authrtwo}{Chris van der Heide}
\newcommand{\affiltwo}{Dept. of Electrical and Electronic Engineering}
\newcommand{\addrstwo}{University of Melbourne}
\newcommand{\emailtwo}{chris.vdh@gmail.com}

\newcommand{\authrthree}{Liam Hodgkinson}
\newcommand{\affilthree}{School of Mathematics and Statistics}
\newcommand{\addrsthree}{University of Melbourne}
\newcommand{\emailthree}{lhodgkinson@unimelb.edu.au}

\newcommand{\authrfour}{Michael W. Mahoney}
\newcommand{\affilfour}{ICSI, LBNL, and Department of Statistics}
\newcommand{\addrsfour}{University of California, Berkeley}
\newcommand{\emailfour}{mmahoney@stat.berkeley.edu}

\ifjournal

    \newcommand{\AuthorBlock}[4]{%
        {\normalfont                          
        \begin{minipage}[t]{0.49\textwidth}   
          \raggedright
          \textbf{#1}\\[2pt]                  
          #2\\                                
          #3\\                                
          \texttt{#4}                         
        \end{minipage}}%
    }

    \author{%
      \AuthorBlock{\authrone}{\affilone}{\addrsone}{\emailone}\hfill
      \AuthorBlock{\authrtwo}{\affiltwo}{\addrstwo}{\emailtwo}%
      \\[17mm]
      \AuthorBlock{\authrthree}{\affilthree}{\addrsthree}{\emailthree}\hfill
      \AuthorBlock{\authrfour}{\affilfour}{\addrsfour}{\emailfour}%
    }
    
\else
    \author[1]{\authrone}
    \author[2]{\authrtwo}
    \author[3]{\authrthree}
    \author[4]{\authrfour}
    \affil[1]{\small{\textit{\affilone, \addrsone} \protect\\ \href{mailto:\emailone}{\protect\nolinkurl{\emailone}}} \vspace{2mm}}
    \affil[2]{\small{\textit{\affiltwo, \addrstwo} \protect\\ \href{mailto:\emailtwo}{\protect\nolinkurl{\emailtwo}}} \vspace{2mm}}
    \affil[3]{\small{\textit{\affilthree, \addrsthree} \protect\\ \href{mailto:\emailthree}{\protect\nolinkurl{\emailthree}}} \vspace{2mm}}
    \affil[4]{\small{\textit{\affilfour, \addrsfour} \protect\\ \href{mailto:\emailfour}{\protect\nolinkurl{\emailfour}}} \vspace{2mm}}
    
    \date{}
\fi

\ifjournal
    \def \FreeAlgPypiUrl {\url{https://pypi.org/project/freealg}}
    \def \FreeAlgDocUrl {\url{https://ameli.github.io/freealg}}
    \def \FreeAlgGithubUrl {\url{https://github.com/ameli/freealg}}
\else
    \def \FreeAlgPypiUrl {\url{https://pypi.org/project/freealg}}
    \def \FreeAlgDocUrl {\url{https://ameli.github.io/freealg}}
    \def \FreeAlgGithubUrl {\url{https://github.com/ameli/freealg}}
\fi

\begin{document}

\addtocontents{toc}{\protect\setcounter{tocdepth}{-1}}  

\maketitle


\begin{abstract}
    Sample covariance matrices are fundamental objects in machine learning and statistics, with valuable information encoded into their eigenvalues. When treating modern problems, the scale of these matrices can become prohibitive for tractable computation. This creates a critical need for ``spectral decompression'': a method for inferring the eigenspectrum of a very large-scale model using only information from its smaller realizations. Existing methods fall into two main categories: those relying on spectral inversion; and those that rely on Free Decompression techniques. On the one hand, spectral inversion is a fundamentally ill-posed problem that is notoriously unstable, often failing to yield reliable results in finite-size settings. On the other hand, while Free Decompression was recently proposed to address this scaling, its reliance on the Nica-Speicher framework restricts its use to unitarily invariant ensembles, a symmetry condition that subsampled covariance matrices fail to satisfy. In this work, we introduce Deformed Decompression (DD), a novel framework for the stable, forward-extrapolation of spectral densities. By leveraging properties of the companion matrix, DD recasts the spectral decompression problem as a directed evolution of the Stieltjes transform of the companion matrix. This approach bypasses the instabilities of traditional inversion and provides a robust methodology for scaling covariance matrices and even their generalized variants (including Hessian matrices) that lack unitary invariance. We demonstrate the utility of our framework through experiments on a variety of random matrix examples, showing that DD successfully captures the complex spectral signatures of large-scale systems.
\end{abstract}


\section{Introduction} 
\label{sec:intro}

Covariance matrices and their generalizations are fundamental objects in data analysis, arising whenever dependence, similarity, variability, or second-order structure needs to be represented. 
Common statistical practice treats the sample covariance as a reasonable reflection of the underlying population covariance. While true when the covariates are fixed and few in quantity relative to the number of datapoints, in the setting where the covariate dimension scales with data, the behavior of the sample covariance becomes fundamentally random, and tools from random matrix theory (RMT) are required for its treatment \cite{DOBRIBAN-2015, COUILLET-2022}. 
This scenario is increasingly relevant in modern machine learning, where resource constraints can simultaneously restrict access to data and compute, and these components need to be delicately scaled accordingly \cite{KAPLAN-2020,HOFFMANN-2022}.
In this regime, a commonly encountered problem is the estimation of a population matrix; this is a notoriously difficult, ill-posed problem, but methods have been developed to provide reasonable estimates under strong regularization \cite{LEDOIT-2012,COUILLET-2014,LEDOIT-2020}. 

Our objective is different. We consider the case where the number of covariates is prohibitively large, rendering it difficult or impossible to form or extract fine-grained spectral information.
For raw data covariances, this may be encountered, for example, in the analysis of genetic datasets \cite{APARICIO-2020,SCHAFER-2005,ZHOU-2018}. 
Fisher information and Hessian matrices of large statistical and ML models have similar structure; for linear models these are precisely (generalized) covariance matrices \cite{LIAO-2021a}, providing an important starting point to model their non-linear counterparts in ML \citep{PENNINGTON-2017c,PENNINGTON-2017a,HAYASE-2021,PANDIT-2024,WANG-2024}. 
The spectra of such matrices are of primary interest for downstream tasks, determining the concentration of variance across directions, presence of near-null modes, numerical stability of inverse problems, and the behavior of determinant-based objectives used in likelihoods and Bayesian inference \citep{UBARU-2017a,HODGKINSON-2023c,HODGKINSON-2023a,HODGKINSON-2023b}. 
Mirroring the ubiquity of these matrices is their difficulty: they are often dense, highly ill-conditioned and non-isotropic,  and sensitive to sampling, regularization, and model misspecification. 
In high dimensions, their empirical spectra may contain sharp edges, spikes, heavy tails, and clusters near zero \cite{ELKAROUI-2008,ELKAROUI-2010}.
These features are often the most important for downstream analysis, e.g., log-determinants and inverse traces, and yet they remain the hardest to estimate accurately \citep{AMELI-2026}.
Classical approaches rely on low-rank approximations, including row/column sampling techniques and Nystrom-type approximations \citep{dkm_matrix3,DM16_CACM,GM15_NYSTROM_JRNL} or matrix-vector products \cite{ubaru2017fast,yao2020pyhessian, POTAPCZYNSKI-2023}.
While effective when the leading eigenspace dominates \cite{ADAMS-2018,GARDNER-2018}, these classical approaches depend on assumptions that can radically fail under the heavy-tailed spectral properties often encountered in modern ML models \citep{HODGKINSON-25,MM18_TR_JMLRversion,MM19_HTSR_ICML,MM20_SDM,MM20a_trends_NatComm,AMELI-2025a}.
The bias in low-rank methods toward large eigenvalues severely underrepresents near-singular directions that are critical for computing determinants, inverses, and for uncertainty quantification \citep{AMELI-2025a,HODGKINSON-2023a,WILSON-2025}.

Recent work on free decompression (FD) \citep{AMELI-2025a,AMELI-2025b} proposes a different strategy. By leveraging free probability and random matrix theory, FD evolves (or ``decompresses'') the empirical spectral density (ESD) of a sampled submatrix to estimate the ESD of a larger, inaccessible matrix. 
While this technique opens the door for spectral information to be obtained from very large ``impalpable'' matrices, there are classes of matrices to which FD does not apply (contrary to the original claim of \citet{AMELI-2025b}). Crucially, among them is the important class of (generalized) covariance matrices (GCMs), which do not possess the unitary invariance properties required by FD. This raises the question of whether frameworks similar to FD can be developed to handle other classes of matrices.

\begin{figure}[!t]
    \centering
    \includegraphics[width=\textwidth]{\figdir/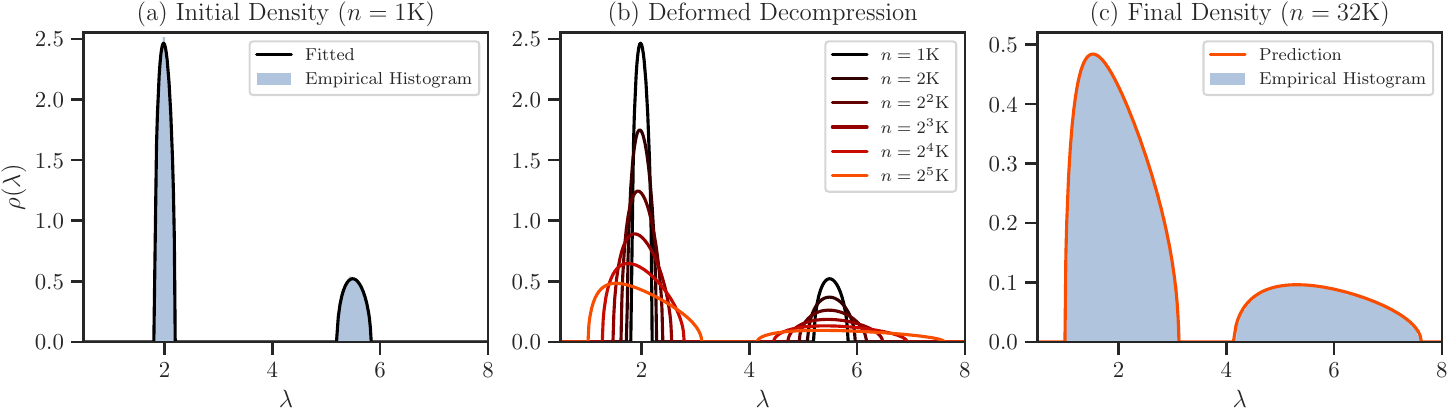}
    \caption{Example of deformed decompression for recovering the ESD of an \(n \times n\) matrix \(\tens{A}\) with \(n = \num{32000}\), generated from a deformed Marchenko--Pastur law with atoms \((t_1, t_2)=(2, \frac{11}{2})\), weights \((w_1, w_2) = \left(\frac{3}{4}, \frac{1}{4}\right)\), and target aspect ratio \(\frac{1}{10}\). (a) ESD of the sampled \(n_0 \times n_0\) matrix \(\tens{A}_0\) with \(n_0 = \num{1000}\) (histogram) together with the fitted density (solid curve). (b) DD flow from \(n_0\) to \(n\). (c) Predicted ESD at size \(n = \num{32000}\), compared with the benchmark ESD of the original matrix.}
    \label{fig:dmp-flow}
    \ifjournal
        \vspace{-4mm}
    \fi
\end{figure}

\paragraph{Contributions.} 
By directly appealing to the structure of covariance matrices and to the behavior of their \emph{companion} Stieltjes and R-transforms, we derive \emph{Deformed Decompression} (DD), a dual formulation of FD governed by a different but related partial differential equation (PDE). We show that DD unlocks the decompressibility of another wide class of matrices, including GCMs, and that it enables analogues of tools developed in \cite{AMELI-2026}; this is illustrated in \Cref{fig:dmp-flow}. In doing so, we





\begin{itemize}[leftmargin=*]
    \item 
    derive \emph{Deformed Decompression} (DD), a new method for spectral expansion of~GCMs;
    \item 
    develop tools to track the dynamics of spectral edges, merging and splitting of bulks, as well as the evolution of spike components, without access to the full distribution;
    \item 
    provide stability and error bounds for spectral expansion and approximation under our approach;
    \item 
    give empirical examples that demonstrate the utility and accuracy of our methods;
    \item 
    publish our implementation in the open source software project \texttt{freealg}.\footnote{Available at \FreeAlgGithubUrl, with documentation at \FreeAlgDocUrl.}
\end{itemize}

The remainder of this work is structured as follows. \Cref{sec:related} briefly discusses some related work and its limitations. \Cref{sec:dd-intro} introduces GCMs and their associated invariance property, develops DD flow, and gives stability and error bounds. \Cref{sec:dd-spectral} describes how this expansion can be achieved under a polynomial assumption, and provides related formulae for the behavior of spectral edges and atoms. \Cref{sec:results} applies this method on a range of numerical examples. We conclude and discuss limitations and future work in \Cref{sec:conclusion}. Additional information is provided in a set of appendices.


\section{Background and Related Work}

\label{sec:related}
As the scale of scientific computing and machine learning problems continues to grow, it is becoming increasingly important to analyze matrices so large that they cannot be formed, in order to extract information for training, evaluation and diagnostics \cite{HAYASE-2021,PENNINGTON-2017c,PENNINGTON-2017a,GARDNER-2018,HODGKINSON-2023b}. For example, Hessian matrices for modern language models contain on the order of $10^{24}$ elements \cite{deepseek2026v4}.
Observing that heavy-tailed structures pose significant challenges for matrix-vector product strategies, \citet{AMELI-2025a} proposed a statistical extrapolation approach to estimate log-determinants of large-scale kernel matrices using linear regression on small principal submatrices. 
While surprisingly accurate for neural tangent kernels of neural network models, its bespoke construction relies heavily on the mathematical structure of log-determinants of kernel matrices. Its rigidity necessitates new parallel methods for each desired quantity and each matrix class. 

For the more general case, methods have been recently proposed to obtain approximations of the \emph{entire spectral distribution} for arbitrary random matrices \citep{AMELI-2025b}. For example, using approximate Stieltjes transforms of submatrices as initial data, the FD method solves a partial differential equation to estimate the spectral distribution of the full matrix. While successful for simple and regularized problems, the fitting procedure and PDE evolution initially posed numerical challenges. These issues were largely overcome in followup work \citep{AMELI-2026}. Drawing from the polynomial method \citep{RAO-2008} and knowledge of ML random matrix models \citep{PENNINGTON-2017a,PENNINGTON-2017c}, the authors impose a polynomial ansatz on the Stieltjes transforms. This assumption accommodates a wider class of matrices, including multi-scale distributions with multiple bulks and spikes, and allows for tracking of moments, spectral edges, and the locations and weights of atoms. Such tools are particularly relevant for GCMs, where near-null directions and multi-scale eigenvalue structures are intrinsic features rather than numerical artifacts. However, this work also identified an imprecise assumption in \cite{AMELI-2025b}: the submatrix sampling used to seed the process can be too structured to guarantee recovery of the original matrix. Consequently, while the tools in \cite{AMELI-2025b, AMELI-2026} apply broadly, certain matrix classes, notably GCMs, remain beyond their reach.

While not explicitly intended for these purposes, several analytic tools for GCMs exist, and could, in principle, be composed to obtain a similar procedure to the one presented here. 
Notably, \citet{DOBRIBAN-2015} leverages the same Silverstein equation we rely on to obtain a procedure for estimating the spectrum of large GCMs. Unlike our method, however, this procedure requires a-priori knowledge of the limiting covariance matrix, severely restricting its direct applicability. To estimate this population covariance matrix, \citet{LEDOIT-2012,LEDOIT-2020} provide mechanisms based on shrinkage estimators. However, the underlying problem is itself notoriously ill-posed (this is seen in \Cref{sec:free-binom}) and sensitive to minor noise fluctuations. 
Their methods consequently rely on strong regularization which, when used for preprocessing, introduces significant smoothing error, eliminating many of the structural features that make these matrices interesting. 

Finally, we mention the foundational work of \citet{MARCHENKO-1967}, who derived a PDE for the Stieltjes transform of deformed Marchenko--Pastur distributions, parameterizing the underlying aspect ratio as a ``time'' parameter. This is closely related to our own, more malleable equations. 


\paragraph{Connection with Free Decompression.}
FD relies on a key invariance given by the Nica-Speicher theorem \citep{NICA-1996}, 
which requires the matrix of interest to be embedded in an asymptotically free sequence of matrices. 
\citet{AMELI-2025b} assert that uniform random subsampling of rows and columns ensures this property, but this claim seems to stem from an erroneous reading of \citet{NICA-1993}. In reality, this invariance is only guaranteed if the projection is Haar-uniform \citep{VOICULESCU-1991}. 
This misconception creates a gap: the techniques in \citet{AMELI-2026} work well when projected permutations exhibit asymptotic freeness, yet numerical tools to verify this condition are currently lacking. 
Similarly, while traffic probability \citep{MALE-2020,CEBRON-2024} offers analytic criteria for subsampling invariance, a clean analogue of FD within that framework remains unclear. 
Instead, we propose a technique targeting covariance matrices by exploiting a different invariance under the \emph{companion R-transform}, providing an analogous toolkit to algebraic free decompression (AFD).
It turns out that this invariance and Nica-Speicher can both hold only for Marchenko-Pastur laws, and so submatrices of nontrivial covariance matrices cannot satisfy the Nica-Speicher property.


\section{Deformed Decompression}
\label{sec:dd-intro}

To develop our method, we consider the class of sample covariance matrices of the form
\begin{equation*}
    \tens{S} \coloneqq \frac{1}{n} \tens{X} \tens{X}^{\intercal},
    \qquad \tens{X} \coloneqq \gtens{\Sigma}^{\frac{1}{2}}_p \tens{Z},
\end{equation*}
where \(\tens{Z} \in \mathbb{R}^{p \times n}\) has i.i.d. centered entries with unit variance, and \(\gtens{\Sigma}_p \in \mathbb{R}^{p \times p}\) is symmetric positive semidefinite.
As our goal is to decompress the matrix $\tens{S}$ in a stable way, we consider it as embedded in a family of matrices $\tens{S}_p$ (see, e.g. \citet{AMELI-2025b}) with the following common property. 

\begin{assumption}[Convergence of Population Covariance Structure]
\label{ass:pop}
    The matrix $\gtens{\Sigma}_p$ is a deterministic or independent population matrix whose empirical spectral distribution (ESD) converges weakly as $n \to \infty$ to a measure $\nu_{\tens{\Sigma}}$ with support on $[0,\infty)$.
\end{assumption}

To obtain a smaller realization of this covariance family, consider a larger matrix $\tens{S}_q$ with $q > p$, holding the number of samples $n$ fixed, and let $\tens{P}_I \in \mathbb{R}^{p \times q}$ select $p$ coordinates. The sampled matrix and its restricted population covariance are
\begin{equation*}
    \tens{S}_I = \tens{P}_I \tens{S}_q \tens{P}_I^\intercal,
    \qquad
    \gtens{\Sigma}_I = \tens{P}_I \gtens{\Sigma}_q \tens{P}_I^\intercal.
\end{equation*}
When $\tens{Z}$ has standard Gaussian entries, or when $\gtens{\Sigma}_q$ is diagonal, $\tens{S}_I$ is again a sample covariance matrix of the same form, with population covariance $\gtens{\Sigma}_I$. For the sampled matrix to belong to the same limiting covariance family, we additionally require that the ESD of $\gtens{\Sigma}_I$ converges weakly to $\nu_{\tens{\Sigma}}$. This holds, for example, under uniform random coordinate sampling of diagonal population covariances with a convergent ESD and a nonvanishing sampling fraction, since the sampled population eigenvalues preserve their limiting proportions. Thus, under this condition, principal-submatrix sampling preserves the limiting population covariance law while changing the dimension of the covariance matrix.

The scalar parameter of interest is the limiting aspect ratio $c \coloneqq \lim_{n \to \infty} \frac{p}{n} \in (0, \infty)$, controlling the geometry of the underlying data matrix $\tens{X}$. In this limit, the ESD of \(\tens{S}\) converges to a deterministic limit $\nu_c$ described by a \emph{deformed Marchenko-Pastur law} \citep{MARCHENKO-1967}, given in terms of the free multiplicative convolution $\nu_c = \nu_{\tens{\Sigma}} \boxtimes \mathrm{MP}_c$. Here, \(\mathrm{MP}_c\) denotes the Marchenko--Pastur law with aspect ratio \(c > 0\), and we remark that the mass at the origin is \(\max\{\nu_{\tens{\Sigma}}(\{0\}), 1 - 1/c\}\). If $\nu_{\tens{\Sigma}}$ consists of a finite number of atoms, \(\nu_c\) may exhibit multiple deformed Marchenko--Pastur bulk components.

For a fixed initial aspect ratio $c > 0$, consider a target aspect ratio $\tau c$ with $\tau>1$ (the \emph{decompression ratio}). The measure $\nu_{\tau c}$ corresponds to the limiting spectrum for a matrix in the same form as $\tens{S}$, but with $\tau$-times the number of rows and columns. Our objective is to construct a map, which we refer to as \bemph{deformed decompression}\footnote{The name is derived from the involvement of the multiplicative convolution, commonly considered a form of \emph{deformation}, as in ``deformed Marchenko-Pastur law''.} (DD), sending $\nu_c \mapsto \nu_{\tau c}$ without requiring knowledge of the population matrix $\tens{\Sigma}_p$ or its limiting spectral distribution $\nu_{\tens{\Sigma}}$. Provided this can be achieved, we can extrapolate from the spectrum of subsampled covariance matrices to the spectrum of the original.

To construct this map, let $m$ denote the \emph{Stieltjes transform} of $\nu_c$ and consider its \emph{companion Stieltjes transform} $\underline{m}(z) = c m + \frac{c-1}{z}$. The R-transform corresponding to $m$ is formally defined as the function $R$ satisfying $R(-m(z)) = z + m(z)^{-1}$. Let $\underline{R}$ be the corresponding R-transform for the companion Stieltjes transform $\underline{m}$. Likewise, we let $\underline{R}_\tau$ denote the corresponding object for $\nu_{\tau c}$. 
In terms of this companion R-transform, we are able to obtain the following characterization of deformed decompression, proved in \Cref{sec:dd-algebraic}.

\begin{tcolorbox}[mymathbox]
\begin{theorem}
    \label{thm:main}
    The deformed decompression map is prescribed by  $\underline{R}_\tau(z) = \tau \underline{R}(z)$.
\end{theorem}
\end{tcolorbox}

This is to be compared to the Nica--Speicher Theorem for free decompression \cite{NICA-1996,AMELI-2025b}, which is defined by a similar operation $R_\tau(z) = R(\tau z)$ in the R-transform.
The origins of \Cref{thm:main} can be understood in two ways. The first considers the \emph{S-transform}, which is defined in \Cref{sec:transforms} and uniquely determines a probability measure. It has the convenient property of factorizing through free multiplicative convolution: if $S_c$ is the S-transform of $\nu_c$, $S_{\tens{\Sigma}}$ is the S-transform of $\tens{\Sigma}$, and $S_{\mathrm{MP}_c}$ is the S-transform of $\mathrm{MP}_c$, then $S_c(w) = S_{\tens{\Sigma}}(w) S_{\mathrm{MP}_c}(w)$. Since $S_{\mathrm{MP}_c}(w) = (1 + c w)^{-1}$ is known explicitly, this relation effectively solves for the S-transform of the population matrix. However, performing deconvolution in this way is infamously ill-posed. Instead, taking advantage of the invariance of $S_{\tens{\Sigma}}$ to the aspect ratio, we may relate the S-transforms at two different aspect ratios $c$ and $c\tau$ by $S_{\tau c}(w)(1 + c \tau w) = S_c(w) (1 + c w)$. This turns out to be better behaved, as seen in \Cref{sec:free-binom}. Through a series of algebraic manipulations (see \Cref{prop:r_companion}), the S-transform may be related to the companion R-transform $\underline{R}$, revealing \Cref{thm:main}. 

The second, perhaps more convenient, interpretation of \Cref{thm:main} arises by considering the companion matrix itself. The companion matrix for $\tens{S}$ is given by $\underline{\tens{S}} = \frac1n \tens{X}^\intercal \tens{X} = \frac1n \tens{Z}^\intercal \tens{\Sigma}_p \tens{Z}$, and its limiting spectral distribution is prescribed in terms of its Stieltjes transform $\underline{m}$. If $\tens{Z}$ is Gaussian, we may consider $\tens{\Sigma}_p = \mathrm{diag}(\sigma_1,\dots,\sigma_p)$ with random ordering, due to the unitary invariance of $\tens{Z}$. Since $\underline{\tens{S}} = \frac1n \sum_{i=1}^p \sigma_i \vect{z}_i \vect{z}_i^\intercal$, where $\vect{z}_i$ is the $i$th row of $\tens{Z}$, increasing the aspect ratio $c$ by increasing $p$ is tantamount to adding rank-1 updates. In the large-dimensional limit, the companion spectral law is compound free Poisson with rate $c$ and jump distribution $\nu_{\tens{\Sigma}}$, whose R-transform scales linearly with $c$, revealing \Cref{thm:main}.



\paragraph{The Deformed Decompression Flow.}
With a clearer picture of how the limiting spectral laws of covariance matrices vary with aspect ratio, we can now derive the flow equations for DD. Fix an initial aspect ratio $c>0$, and let $m_1$ and $\underline{m}_1$ be the regular and companion Stieltjes transforms, respectively. Letting $m_\tau$ and $\underline{m}_\tau$ denote the corresponding transforms for the deformed decompression $\nu_{\tau c}$ under a ratio $\tau$, we obtain inviscid Burgers' equations governing their dynamics, analogous to those for FD \citep{AMELI-2025b}. The proof of \Cref{thm:pde-all} is presented in \Cref{sec:key}. 

\begin{theorem}
\label{thm:pde-all}
    Let $m_\tau(z)$ denote the Stieltjes transform under the deformed decompression flow for initial aspect ratio $c$, and let $\underline{m}_\tau(z)$ denote its companion. Then
    \begin{align}\label{eqn:dd-flow}
        \tau \underline{m}_\tau\frac{\partial}{\partial \tau} \underline{m}_\tau + (z \underline{m}_\tau + 1)\frac{\partial}{\partial z} \underline{m}_\tau = 0.
    \end{align}
    In terms of the Stieltjes transform, this becomes
    \begin{align*}
        (c z \tau m_\tau + c \tau - 1) \frac{\partial m_\tau}{\partial \tau} + c (m_\tau z + 1)\left(z \frac{\partial m_\tau}{\partial z} + m_\tau\right) = 0. 
    \end{align*}
\end{theorem}


While we avoid the PDEs for computing $m_\tau$ in our implementation, they prove useful as analytic tools. In practice, the initial Stieltjes transform is approximated from a sampled matrix, raising the question of how small approximation errors propagate under DD. By linearizing the flow, we show that, under suitable conditions, these perturbations grow at most polynomially in $\tau$ over finite decompression intervals. The following proposition makes this precise, with proof given in \Cref{app:stability}.

\begin{proposition}
    \label{prop:stability_bound}
    Let $\underline{m}_\tau$ be a companion Stieltjes transform of a density $\rho$ such that $\|\rho\|_{L^\infty} < +\infty$. Let $\eta_1$ denote an initial perturbation to $\underline{m}_1$ at $\tau=1$, and let $\eta_\tau$ denote its evolution under the linearization of the deformed decompression flow \eqref{eqn:dd-flow}. Under the conditions of \Cref{prop:stability} and the uniform support bound of \Cref{rem:stability}, for any fixed $y > 0$ and finite $T > 1$, there exists a constant $\nu > 0$ such that for any $1 < \tau \leq T$, it holds $\|\eta_\tau\|_{L^2(\Gamma_y)} \leq \tau^\nu \|\eta_{1}\|_{L^2(\Gamma_y)}$.
\end{proposition}


\paragraph{Deformed Decompression is not Free Decompression.}
Both deformed decompression and free decompression obey related first-order complex transport flows. While obtained by distinct derivations, it is natural to consider deeper connections between them. It turns out that the two processes are virtually complementary. The following proposition formalizes this, stating that the only nontrivial distributions for which FD and DD are valid simultaneously belong to the Marchenko-Pastur family, reflecting the self-dual nature of these limiting distributions under the operation $\tens{X} \tens{X}^{\intercal} \mapsto \tens{X}^{\intercal} \tens{X}$.

\begin{proposition}
    \label{prop:mp-only}
    Let $\{m(\tau,z)\}_{\tau\geq 1}$ be a family of Stieltjes transforms of probability measures $\{\nu_\tau\}_{\tau\geq 1}$ satisfying the PDEs for both the free decompression and deformed decompression flows for initial ratio $c>0$. Then for any $\tau \geq 1$, $\nu_\tau$ is a scaled Marchenko-Pastur distribution. 
\end{proposition}

\Cref{prop:mp-only} contrasts with the intended universality of free decompression implied in \citet{AMELI-2025b}; for performing decompression of an arbitrary matrix $\tens{A}$, either free decompression \emph{or} deformed decompression may succeed. Both tools are necessary for handling general impalpable matrices.


\paragraph{Generalized Covariance Matrices.}
Assume that $\tens{D}_n \in \mathbb{R}^{n\times n}$ is a diagonal matrix comprised of non-negative elements, and consider the matrix $\tens{H}$ and its companion given by 
\begin{align}\label{eqn:hessian}
    \tens{H} = \frac{1}{n} \tens{\Sigma}_p^{1/2} \tens{Z} \tens{D}_n \tens{Z}^\intercal \tens{\Sigma}_p^{1/2},\qquad \underline{\tens{H}} = \frac{1}{n} \tens{D}_n^{1/2} \tens{Z}^{\intercal} \tens{\Sigma}_p \tens{Z} \tens{D}_n^{1/2}.
\end{align}

Note that, unlike regular covariance matrices, the companion matrix of $\tens{H}$ is of the same general form. Matrices in this generalized class commonly appear as Hessians under non-quadratic loss functions \citep{LIAO-2021a}. GCMs are conceptually distinct from the covariance matrix model treated so far where $\tens{D} = \tens{I}$. Unfortunately, there exists no single map that can send $\nu_c \mapsto \nu_{\tau c}$ for any arbitrary $\tens{D}$ (see discussion in \Cref{sec:sep-covariance}). Fortunately, in many cases, including Hessian matrices for models with cross-entropy loss \citep[p. 9]{LIAO-2021a}, the eigenvalues of $\tens{D}$ are bounded within a small interval or exhibit limited dispersion. The following proposition asserts that deformed decompression may still perform very well in this regime, as the limiting Stieltjes transform $m_{\tens{\Sigma},\tens{D}}$ of $\tens{H}$, can be well-approximated by the corresponding transform of a covariance matrix model $m_{\tens{\tilde\Sigma},\tens{I}}$. 

\begin{proposition}
    \label{thm:variance-bound}
    Let $\Omega \subset \mathbb{C}^+$ be a bounded region on the upper half of the complex plane with positive distance from the real axis. There exists another limiting family of covariance matrices $\tilde{\tens{\Sigma}}$ such that
    \[
    \sup_{z \in \Omega}|m_{\tens{\Sigma},\tens{D}}(z) - m_{\tilde{\tens{\Sigma}},\tens{I}}(z)| = \mathcal{O}(\sigma_{\tens{D}}^2),
    \]
    after taking $n \to \infty$, where $\sigma_{\tens{D}}^2$ is the limiting variance in the spectral distribution of $\tens{D}$.
\end{proposition}


\section{Deformed Decompression on Spectral Curves} \label{sec:dd-spectral}

The strong analogy between DD and FD turns out to be highly useful for implementing DD. While the original FD implementation in \citet{AMELI-2025b} demonstrated its utility across an impressive number of tasks, it still relied on an implicit assumption that the Stieltjes transform is a root of a quadratic polynomial. \citet{AMELI-2026} extended their earlier implementation by allowing for Stieltjes transforms that satisfy any polynomial relation. Their strategy casts the decompression process along a \emph{spectral curve}, circumventing the need to solve a PDE or perform analytic continuation. Importantly, this enabled practical FD on a far wider array of matrices relevant to machine learning tasks, and provided tools to approximate a number of important spectral quantities. Fortunately, under this same assumption a parallel toolkit can be developed for DD. In order to do so, as in \citet{AMELI-2026}, we impose the following ansatz. 

\begin{assumption}\label{ass:polynomial}
    The Stieltjes transform \(m\) is \emph{algebraic}: there exists a nonzero bivariate polynomial \(P: \mathbb{C} \times \mathbb{C} \to \mathbb{C}\) such that for any $z \in \mathbb{C} \setminus \mathcal{K}$ where $\mathcal{K}$ is a finite set of singularities,
\begin{equation}
    P(z, m(z)) = 0. \label{eq:polynomial}
\end{equation}
\end{assumption}

Let $d_z$ denote the degree of $P$ in $z$, and $s$ its degree in $m$. While the Stieltjes transform encodes the underlying spectral distribution, the invariance that we exploit to perform DD is in terms of its companion Stieltjes transform. 
The co-ordinate change $(\zeta,\underline{m}_1(\zeta))\mapsto (z,\underline{m}_\tau(z))$ given by the characteristic curves of \eqref{eqn:dd-flow} provides a more natural spectral curve to follow the effect of DD on the underlying polynomial \eqref{eq:polynomial}. 
This is captured in our core computational \Cref{thm:SpectralCurve}, whose proof can be found as \Cref{thm:formal-decomp} in \Cref{sec:dd-algebraic}.

\begin{tcolorbox}[mymathbox]
\begin{theorem}
    \label{thm:SpectralCurve}
    Let $m$ be a Stieltjes transform satisfying \Cref{ass:polynomial}, $\tau\geq1$ be a decompression ratio, and $m_\tau$ be the deformed decompression of $m$ by the factor $\tau$. Then every point $(z, m_\tau(z)) \in \mathbb{C}^2$ satisfies
    \begin{align}\label{eq:dd-polynomial}
        P\left(\zeta, c^{-1}\left(\underline m_\tau - \frac{c-1}{\zeta}\right)\right) = 0, \quad 
        \text{with}\quad \zeta = \tau^{-1}\left(z + {\underline{m}_\tau^{-1}}\right) - \underline{m}_\tau^{-1}.    
    \end{align}
\end{theorem}
\end{tcolorbox}



\subsection{Computing and Evolving the Companion Stieltjes Transform}

The downside to working with \eqref{eq:polynomial} is that at each $z\in\mathbb{C}$, $P(z, \cdot)$ has $s$ possible solutions for $m$ up to multiplicity. The \emph{physical branch} corresponding to the Stieltjes transform of interest satisfies the Herglotz property \(\Im m(z) > 0\) for \(z \in \mathbb{C}^{+}\) and the normalization \(m(z) = -z^{-1} + O(z^{-2})\) as $|z|\to\infty$. Fitting $P$ from a real matrix $\tens{A}$ and identifying this correct branch was a key challenge overcome in \citet[Appendices~D and E]{AMELI-2026}, and we deploy their methods here.


\paragraph{Evolving the Physical Branch.}
DD shares significant similarities and differences with FD. On the one hand, casting DD as evolution on a spectral curve allows us to adapt many AFD methods. However, this evolution is arguably simpler, as Biane's Subordination Theorem \citep{BIANE-1998} guarantees that transport on the DD-induced curve maps the upper-half plane to itself. Consequently, while the spectral curve for FD provides automatic analytic continuation, DD computations need only consider the physical branch of the companion Stieltjes transform. 

The evolved algebraic relation in \Cref{eq:dd-polynomial} gives a characterization of candidate roots \(m_{\tau}(z)\); however, it does not distinguish between the physical and non-physical branches. To obtain the physical root, we follow an analogous procedure to AFD \citep{AMELI-2026} by tracking the known physical solution at \(\tau = 1\) along the spectral curves. For a fixed query point \(z\), we work in the spectral-curve coordinates \((\zeta, w)\) where $w = \underline{m}_\tau$, $\underline{P}(\zeta,w) = P(\zeta, c^{-1}(w-\zeta^{-1}(c-1))) = 0$, and view the characteristic relation $\zeta - \tau^{-1} (z + w^{-1}) + w^{-1} = 0$ as a constraint whose derivative in $\tau$ for fixed $z$ is \(\dot\zeta + \tau^{-2}(z + w^{-1}) + (\tau^{-1} - 1)w^{-2}\dot w = 0\). This recasts \((\zeta, w)\) as the coupled system
\begin{equation*}
    F_1(\zeta, w) \coloneqq \underline{P}(\zeta, w) = 0,
    \qquad
    F_2(\zeta, w; \tau, z) \coloneqq \zeta - \tau^{-1}z - (\tau^{-1} -1)w^{-1} = 0.
\end{equation*}
Note that at \(\tau = 1\), we have \(\zeta = z\) and \(w = \um_1(z)\), where \(\um_1(z)\) has previously been obtained by a physical-branch selection procedure. Fixing $z$ and taking the derivative in \(\tau\) yields the tangent equations
\begin{equation*}
    \begin{bmatrix}
        \partial_{\zeta} \underline{P}(\zeta, w) & \partial_w \underline{P}(\zeta, w) \\
        1 & (\tau^{-1} - 1)w^{-2}
    \end{bmatrix}
    \begin{bmatrix}
        \dot{\zeta} \\
        \dot{w}
    \end{bmatrix}
    =
    \begin{bmatrix}
        0 \\
        -\tau^{-2}(z + w^{-1})
    \end{bmatrix}.
\end{equation*}
We can generate a prediction of the next decompression-step by approximating this system of geometric ordinary differential equations. A Newton projection is then used to correct the predictor back onto the constrained set $(F_1,F_2) = (0,0)$. When the local geometry becomes stiff, the steps are appropriately subdivided. Following this corrector step, the decompressed Stieltjes transform is recovered from its companion via $ m_{\tau}(z) = \frac{1}{\tau c}\left(w - \frac{\tau c - 1}{z}\right)$. A full description of this geometric predictor--corrector scheme is provided in \Cref{sec:dd-on-sc}.


\begin{wrapfigure}{R}{0.62\textwidth}
    \ifjournal
        \vspace{-6mm}
    \else
        \vspace{-5mm}
    \fi
    \centering
    \includegraphics[width=0.62\textwidth]{\figdir/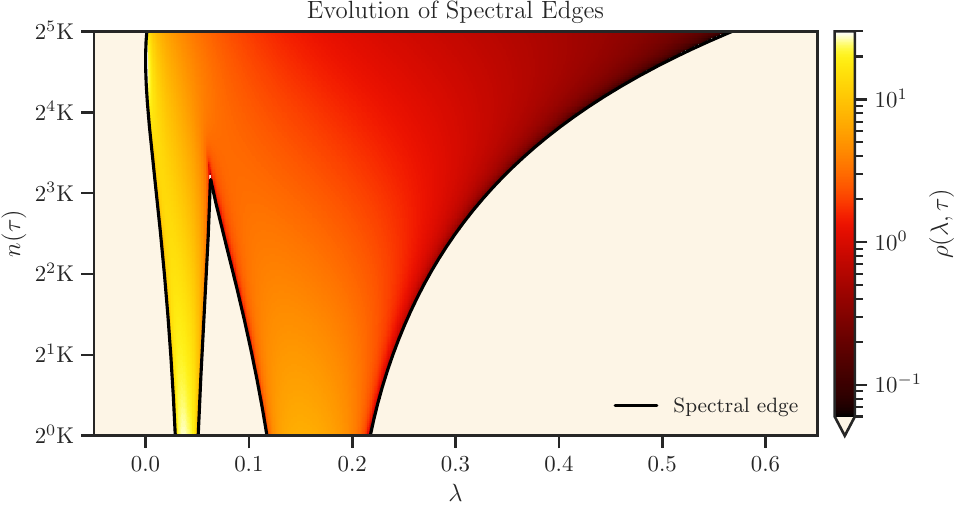}
    \ifjournal
        \vspace{-6mm}
    \else
    \fi
    \caption{Evolution of spectral edges (solid curve) corresponding to the density in \Cref{fig:hessian2-flow}. Color in log scale represents the evolved density, with beige background having zero density (spectral gaps). 
    }
    \label{fig:hessian2-edge}
    \ifjournal
        \vspace{-6mm}
    \else
        \vspace{-4mm}
    \fi
\end{wrapfigure}
\subsection{Spectral Quantities}

DD also allows for the tracking of the evolution of a number of spectral quantities of interest. In particular, the atomic weights, distributional edge locations, as well as moments can all be computed independently from the flow of the underlying distribution. 


\paragraph{Atoms.}
An atom located at $x_{\circ}$ is characterized as a simple pole of the Stieltjes transform $m(z)$ at $z=x_{\circ}$, whose weight is given in terms of its residue
$w = 
\lim_{z\to x_{\circ}}(x_{\circ} - z)m(z).$
Under DD, nonzero atom locations are pushed away from zero, and we can track their weights explicitly as functions of the decompression ratio $\tau$. 
The precise dynamics are encoded in the following proposition, whose proof is given in \Cref{sec:atoms}.

\begin{proposition}[Dynamics of Atoms under Deformed Decompression]
    \label{prop:dd-atom-short}
    Let \(x_\circ\in\mathbb R\) be an atom of the spectral measure \(\nu\) with initial mass \(w_\circ\in(0,1)\). Under deformed decompression with parameter \(\tau\), the following hold. If \(x_\circ\neq 0\), then the atom location is pushed forward to
    \[
        x_\circ(\tau)=\tau x_\circ,
        \quad \text{and mass evolves as} \quad
        w(\tau) = w_\circ - \frac{\tau - 1}{c \tau}.
    \]
    If \(x_\circ=0\), then $x_\circ(\tau) = 0$, and its mass evolves as $w(\tau) = \max \left\{w_\circ, 1 - \frac{1}{c \tau}\right\}$. For \(x_\circ \neq 0\), the formula describes the atom while $w(\tau)>0$. When the predicted mass reaches zero, the pole disappears from the physical branch.
\end{proposition}


\paragraph{Spectral Edges.} 
To track the support of the spectral density, whose bulk regions may split or merge as the spectral density evolves, we will assume that the spectral distribution can be decomposed into a finite set of atoms and absolutely-continuous bulk components. The bulk support can then be written as $I(\tau) = \bigcup_{j=1}^k I_j(\tau)$, where each $I_j(\tau) = [a_j(\tau),b_j(\tau)]$, with spectral edges being the boundary points $\{a_j(\tau),b_j(\tau)\}_{j=1}^k$. 
\Cref{ass:polynomial} characterizes these points in terms of the \emph{branch points} of the corresponding spectral curve, where different branches meet, satisfying the constraint equations
\begin{align*}
    &P(\zeta_\ast,y_\ast)=0, \\
    &\tau \left(c\zeta_\ast y_\ast+c-1\right)^2 \partial_y P(\zeta_\ast,y_\ast) + (\tau-1) \left(c\zeta_\ast^2\,\partial_\zeta P(\zeta_\ast,y_\ast) + (c-1)\partial_y P(\zeta_\ast,y_\ast) \right) = 0,
\end{align*}
where $y$ belongs to the physical branch. This is made precise as \Cref{prop:dd-branch-point} in \Cref{sec:edge}. 
Our procedure for finding the spectral edges is described in \Cref{alg:edge}. 


\paragraph{Merging and Splitting Bulks.}
Under deformed decompression, neighboring bulk regions may merge. Viewed in reverse, this corresponds to a bulk splitting into multiple bulks. These transitions correspond to second-order critical points of the spectral projection. These points are identified in \Cref{prop:dd-cusp} in \Cref{sec:merge} as satisfying
\begin{align*}
    &P(\zeta_{\ast}, y_{\ast}) = 0, \\
    &\tau_\ast (c \zeta_\ast y_\ast + c - 1)^2 \partial_y P + (\tau_\ast - 1) \left(c \zeta_\ast^2 \partial_\zeta P + (c - 1) \partial_y P\right) = 0, \\
    &(c\zeta_\ast y_\ast+c-1) \left[2\zeta_\ast(\partial_\zeta P)(\partial_yP)^2 + \zeta_\ast^2 \left(\partial_{\zeta\zeta}P(\partial_yP)^2 - 2\partial_{\zeta y}P\,\partial_\zeta P\,\partial_yP + \partial_{yy}P(\partial_\zeta P)^2 \right) \right] \\
    & \qquad - 2\partial_y P \left((c-1)\partial_yP + c\zeta_\ast^2\partial_\zeta P \right) \left(y_\ast\partial_yP-\zeta_\ast\partial_\zeta P \right) =0.
\end{align*}



\paragraph{Moments.}
Spectral moments can also be evolved without reconstructing the density. As shown in \Cref{sec:evo-moments}, the first moment remains constant, \(\mu_1(\tau) = \mu_1(1)\), while the variance grows linearly with the decompression ratio: \(\operatorname{Var}(\nu_{\tau c}) = \operatorname{Var}(\nu_c) + c(\tau - 1)\mu_1(1)^2\). More generally, for $k \geq 1$, the $k$-th moment is a polynomial in $\tau$ of degree at most $k-1$ (see \Cref{prop:moment-decompression}).


\section{Numerical Experiments}
\label{sec:results}

To demonstrate DD, we will consider three examples. The first mirrors the canonical example we used to derive our method in \Cref{sec:dd-intro}, an approximation of a deformed Marchenko--Pastur law.  Then, we consider a generalized covariance matrix of the form given in \eqref{eqn:hessian}, the Hessian of a logistic regression model. Finally, we treat the covariance matrix of the STL-10 image dataset \cite{COATES-2011}.


\paragraph{Deformed Marchenko--Pastur.}
The deformed Marchenko--Pastur law is the canonical model for deformed decompression explicitly used in its construction as the limiting spectral distribution of covariance matrices. Detailed background and description on this distribution are given in \Cref{sec:deformed-mp}. To generate the matrix of interest, we generated an \(n \times n\) matrix \(\tens{A}\) with \(n = \num{32000}\) from a deformed Marchenko--Pastur law with atoms \((t_1, t_2)=(2, \frac{11}{2})\), weights \((w_1, w_2) = \left(\frac{3}{4}, \frac{1}{4}\right)\), and target aspect ratio \(\frac{1}{10}\). A \(n_0 \times n_0\) matrix \(\tens{A}_0\) with \(n_0 = \num{1000}\) was subsampled from the original matrix, and its empirical Stieltjes transform generated from its spectral decomposition, fit with $d_z=2$ and $s=3$. DD was performed, with visualization of the flow given in \Cref{fig:dmp-flow}.



\begin{figure}[!t]
    \centering
    \includegraphics[width=\textwidth]{\figdir/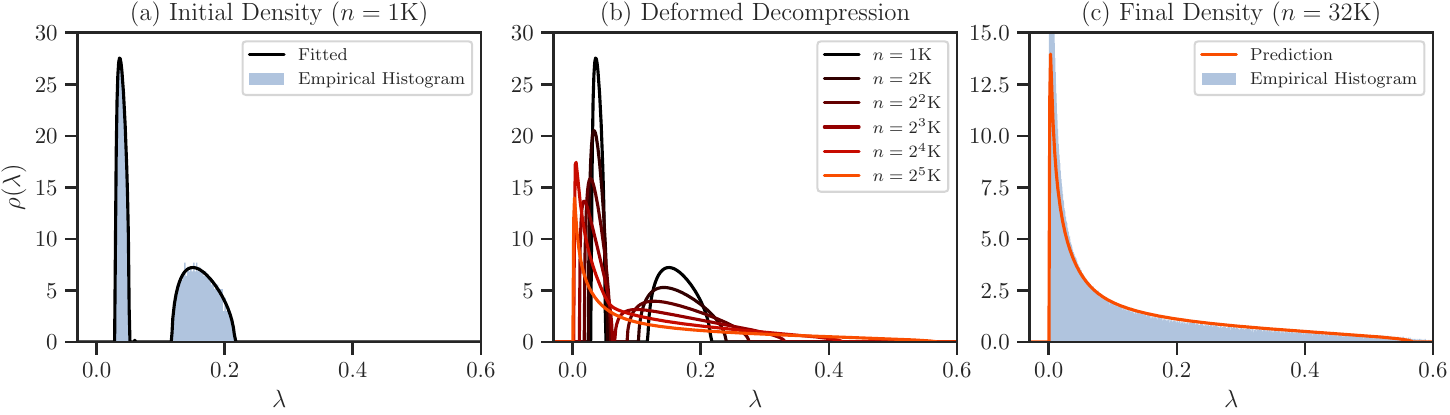}
    \ifjournal
        \vspace{-5mm}
    \fi
    \caption{Example of deformed decompression for recovering the ESD of an \(n \times n\) Hessian matrix \(\tens{A}\) following \citep{LIAO-2021a}, here with \(n = \num{32000}\). (a) ESD of the sampled \(n_0 \times n_0\) matrix \(\tens{A}_0\) with \(n_0 = \num{1000}\) (histogram) and the fitted density (solid curve). (b) Deformed decompression flow from \(n_0\) to \(n\). (c) Final predicted ESD at \(n = \num{32000}\), compared with the benchmark histogram of the original matrix.}
    \label{fig:hessian2-flow}
    \ifjournal
        \vspace{-3mm}
    \fi
\end{figure}

\paragraph{Hessian Matrices.}
We now consider the Hessian of a logistic regression loss function, where the features have non-uniform covariance. This model is taken from \citep{LIAO-2021a} and is a bona-fide GCM with $\tens{D}$ a diagonal matrix comprised of the second derivatives of the loss. The procedure used for training and Hessian formation is given in \Cref{sec:hessian-example}. To perform DD, once the \(n \times n\) matrix \(\tens{H}\) with \(n = \num{32000}\) was obtained, a \(n_0 \times n_0\) submatrix \(\tens{H}_0\) with \(n_0 = \num{1000}\) was sampled from the original matrix. The initial Stieltjes transform was fit from its empirical analogue with $d_z=2$ and $s = 4$. DD was performed, and its flow is visualized in \Cref{fig:hessian2-flow}. The evolution of its edges, demonstrating the merging of two bulk regions, is demonstrated in \Cref{fig:hessian2-edge}. \Cref{tab:hessian-complexity} provides distributional distances in terms of normalized Wasserstein-1 and Gaussian-kernel maximum mean discrepancy, as well as relative error of estimates on the first two moments of distributions obtained during the flow. Although not a covariance matrix, DD still appears to perform well. This is consistent with the small variance in the entries of $\tens{D}$, as seen in \Cref{fig:hessian-diag}. A further example of a Hessian matrix with $n=\num{16000}$ is given in \Cref{sec:hessian-example}.

\begin{figure}[t]
    \centering
    \includegraphics[width=0.54\textwidth]{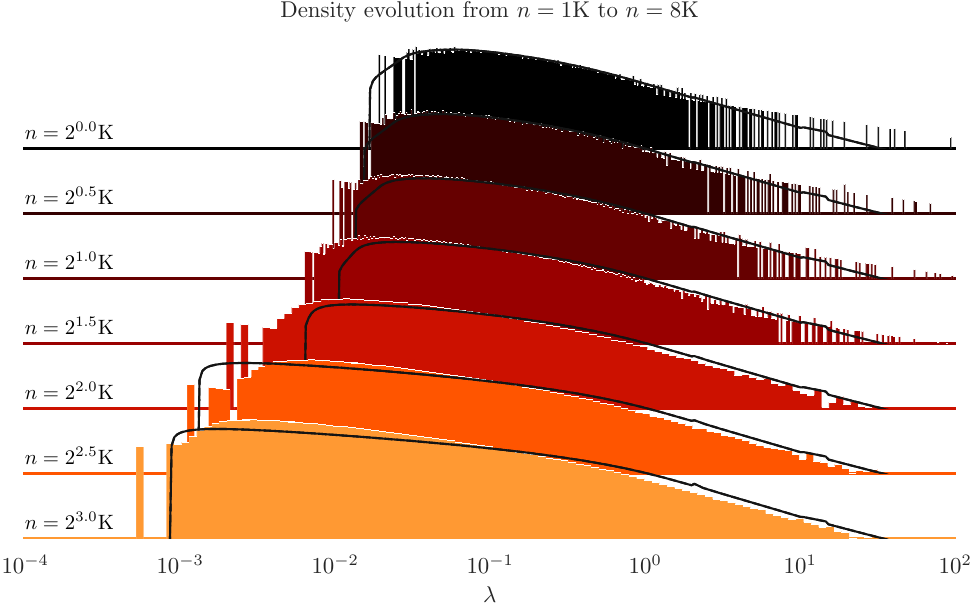}%
    \hfill%
    \includegraphics[width=0.45\textwidth]{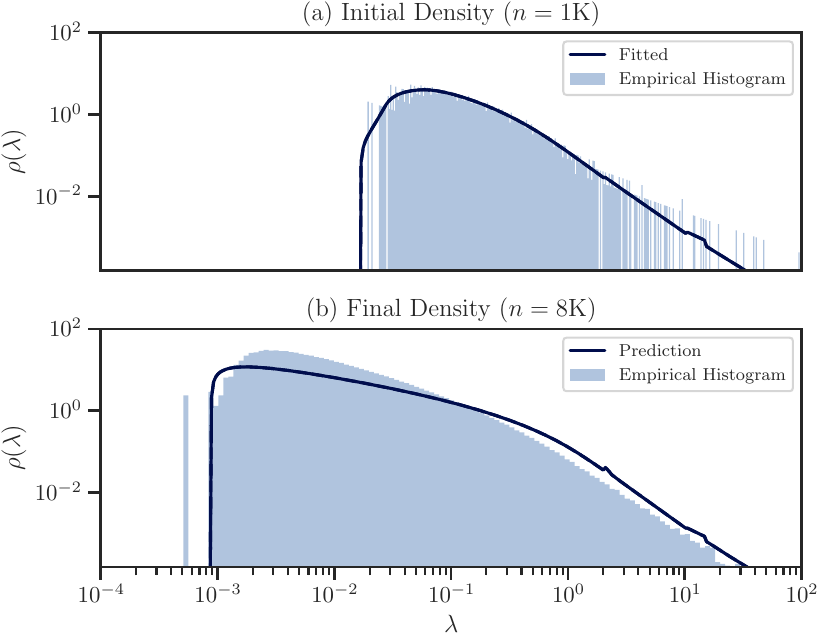}
    \ifjournal
        \vspace{-2mm}
    \fi
    \caption{DD recovering the ESD of an \(\num{8000} \times \num{8000}\) covariance matrix \(\tens{A}\) of the STL-10 dataset. \emph{Left:} ridge plot of the DD density evolution on log scale (solid lines), compared with ESD of sampled submatrices. \emph{Right:} (a) ESD of the sampled \(\num{1000}\times\num{1000}\) submatrix, with density fit; and (b) final predicted ESD, compared with the ESD of the original matrix.}
    \label{fig:diff-ridge}
    \ifjournal
        \vspace{-4mm}
    \fi
\end{figure}

\paragraph{STL-10 Dataset Covariance Matrix.}\label{sec:stl}
Finally, we consider a Gram covariance matrix obtained from a real dataset, the STL-10 image dataset. Details on this dataset and the procedure used to form the matrix, as well as additional figures and information on its edge behavior, are given in \Cref{sec:stl10}. To perform DD, we formed the Gram covariance matrix \(\tens{C} = p^{-1} \tens{X} \tens{X}^{\intercal}\) from \(n=\num{8000}\) STL-10 images with feature dimension \(p=\num{27648}\), and then sampled an \(n_0 \times n_0\) principal submatrix \(\tens{C}_0\) with \(n_0=\num{1000}\). DD was performed by fitting the initial Stieltjes transform with $d_z = 3$ and $s = 12$, and the distribution was evolved up to $\tau = 8$. This is visualized in \Cref{fig:diff-ridge}, which depicts ridge-plots of the estimated densities during the flow on log-scale, compared to the ESDs of subsamples.

\begin{wraptable}{R}{0.61\textwidth}
\ifjournal
    \vspace{-6.5mm}
\else
    \vspace{-7.5mm}
\fi
\caption{Process times for obtaining eigenvalues, as well as corresponding distributional distances and relative error on the first moment for the STL-10 covariance matrix considered in \Cref{sec:stl}.} 
\label{tab:stl-complexity}
\vspace{1.5mm}
\centering
\begin{tabular}{lrrrrr}
    \toprule
    Size & \multicolumn{2}{c}{Process Time (sec)} & \multicolumn{3}{c}{Accuracy Metrics} \\
    \cmidrule(lr){2-3} \cmidrule(lr){4-6}
    \(n\) & Direct & DD (ours) & \(W_1/L\) & MMD & \(\Delta \mu_1 /\mu_{1}\) \\
    \midrule
    \(2^{0}\)K & \(  4.2\) & \(4.2 + 0.0\) & \(0.00\%\) & \(0.23\%\) & \(0.51\%\) \\
    \(2^{1}\)K & \( 14.0\) & \(4.2 + 2.1\) & \(0.58\%\) & \(2.84\%\) & \(0.25\%\) \\
    \(2^{2}\)K & \( 69.5\) & \(4.2 + 2.7\) & \(1.42\%\) & \(1.24\%\) & \(0.05\%\) \\
    \(2^{3}\)K & \(579.6\) & \(4.2 + 3.7\) & \(3.70\%\) & \(2.34\%\) & \(0.03\%\) \\
    \bottomrule
\end{tabular}
\end{wraptable}

\Cref{tab:stl-complexity} summarizes comparisons of runtime of eigenvalue computation both under DD and direct calculation of subsamples. Distributional errors for log-eigenvalues were also computed in terms of normalized Wasserstein-1 distance \(W_1/L\), along with Gaussian-kernel maximum mean discrepancy \cite{GIBBS-2002,PANARETOS-2019}, and first-moment errors. Despite the substantial decrease in cost, the incurred errors are relatively small.



\section{Conclusions}
\label{sec:conclusion}
We introduce deformed decompression, a novel method for approximating spectral distributions of large generalized covariance matrices from smaller sampled submatrices. Due to their inherent structure, these matrices were previously not treatable by existing decompression methods. We also develop tools to track spectral edges and bulk mergers, atomic locations and weights, and arbitrary moments. DD opens the door for the extraction of spectral information in settings where the matrices themselves cannot explicitly be formed. These include covariance matrices of datasets with extremely large numbers of covariates, and covariance-like structures, such as the Fisher information or Hessian, for large models. 

\paragraph{Limitations.}
While our method broadens the class of matrices that can be treated using decompression techniques, it only provides approximations for GCMs in terms of a nearby covariance matrix. In particular, our theory applies to positive-semidefinite covariance and curvature matrices, but does not extend to indefinite Hessians in general. As such, there still exist broad classes of matrices that cannot be treated using either FD or DD. Moreover, without prior analytical knowledge of the subsampled matrix, it remains unclear how to determine whether FD or DD may apply. While the reverse operation of deformed \emph{compression}, where $0 < \tau < 1$, may be of independent interest for its potential recovery of the population covariance matrix, we found it to be too unstable in preliminary experiments.


\newcommand{\acknowledgmenttext}
{
    LH is supported by the Australian Research Council through a Discovery Early Career Researcher Award (DE240100144). MWM acknowledges partial support from DARPA, DOE, NSF, and~ONR. This work was supported in part by the U.S. Department of Energy, Office of Science, Office of Advanced Scientific Computing Research's Applied Mathematics Competitive Portfolios program under Contract No. AC02-05CH11231.
}

\newcommand{\competinginterest}
{
    The authors declare no competing interests.
}


\ifjournal
   \section*{Acknowledgments}
   \acknowledgmenttext

   \section*{Competing Interests}
   \competinginterest
\else
    \vspace{0.4cm}
    \noindent\textbf{Acknowledgments.}
    \acknowledgmenttext

    \vspace{0.4cm}
    \noindent\textbf{Competing interests.}
    \competinginterest
\fi


\newpage

\ifjournal
    \bibliographystyle{apalike2}
\else
    \bibliographystyle{apalike2}
\fi
{
\bibliography{refs}
}


\newpage

\ifjournal
    \onecolumn
\fi

\appendix
\begin{appendices}

\addtocontents{toc}{\protect\setcounter{tocdepth}{2}}  
\tableofcontents

\section{Notation and Conventions} \label{sec:notation}

\paragraph{Nomenclature.}
We use boldface lowercase letters for vectors, boldface upper case letters for matrices, and normal face letters for scalars, including the components of vectors and matrices. \Cref{tab:nomenclature} summarizes the main symbols and notations used throughout the paper, organized by context.

\begin{table}[hpt!]
\centering

\caption{Common notations used throughout the manuscript.} 
\label{tab:nomenclature}

\begin{tabular}{lll}
    
    \toprule
    Context & Symbol & Description \\
    \midrule
    \multirow{2}{*}{\makecell[l]{Matrices \& \\ Models}}
                  & \(\mathbf{A}, \mathbf{A}_n\) & General \(n \times n\) matrix (often symmetric or Hermitian) \\
                  & \(\mathbf{A}_n^{(k)}\) & \(k \times k\) principal submatrix of \(\mathbf{A}_n\) \\
                  & \(\mathbf{I}\) & The identity matrix \\
                  & \(\lambda_i\) & The \(i\)-th eigenvalue of a matrix \\
                  & \(\operatorname{trace}(\cdot)\) & Standard trace of a matrix \\
                  & \(\operatorname{tr}_n(\cdot)\) & Normalized trace, \(n^{-1} \operatorname{trace}(\cdot)\) \\
                  & \(\mathbf{Q}_n\) & Haar-distributed random orthogonal matrix \\
                  & \(\mathbf{H}, \mathbf{J}\) & Hessian matrix (and subcomponents) and Jacobian matrix \\
                  & \(\mathbf{\Sigma}\) & Population covariance matrix \\
                  & \(\mathbf{W}, \mathbf{X}, \mathbf{Z}\) & Random matrices with i.i.d. entries (e.g., Wigner, Gaussian) \\
                  & \(\lambda, c\) & Aspect ratios (\(p/n\)) or intensities/rates for random matrix models \\
    \addlinespace[1mm]  
    \arrayrulecolor{gray!30}\cline{1-3} 
    \addlinespace[1mm]
    \arrayrulecolor{black}
    Free Probability
                  & \(\nu\) & Spectral probability measure on \(\mathbb{R}\), possibly including atoms \\
                  & \(\rho\) & Density of the absolutely continuous part of \(\nu\), \(\rho = \mathrm{d}\nu_{\mathrm{ac}}/\mathrm{d}x\), \(\mathbb{R} \to \mathbb{R}^{+}\) \\
                  & \(I\) & Compact support of the density \(\rho\), \(I = \supp(\rho) \subsetneq \mathbb{R}\) \\
                  & \(m\) & Stieltjes transform of measure, \(m(z) = \int_{\mathbb{R}} \frac{\nu(\mathrm{d} y)}{y - z}\), \(\mathbb{C} \setminus \supp(\nu) \to \mathbb{C}\) \\
                  & \(\omega\) & Functional inverse of the Stieltjes transform, \(\omega(m(z)) = z\) \\
                  & \(\mathcal{H}\{\nu\}\) & Hilbert transform of measure, \(\mathcal{H}\{\nu\}(x) = \frac{1}{\pi}\operatorname{p.v.}\int_{\mathbb{R}} \frac{\nu(\mathrm{d}y)}{x-y}\), \(\mathbb{R} \to \mathbb{R}\) \\
                  & \(R\) & Voiculescu's R-transform, \(R(z) = \omega(-z) - \frac{1}{z}\) with \(\omega(m(z)) = z\) \\
                  & \(S\) & S-transform, \(S(z) = \frac{1+z}{z}\psi^{-1}(z)\), where \(\psi(z) = \int_{\mathbb{R}} \frac{zy}{1 - zy} \, \nu(\mathrm{d}y)\) \\
                  & \(\nu^{\boxplus s}\) & \(s\)-fold free additive convolution power of \(\nu\) \\
                  & \(D_c\nu\) & Dilation of measure \(\nu\) by \(c\), i.e., push-forward under \(x \mapsto cx\) \\
                  & \(K(z)\) & K-transform, $K(z) = R(z) + z^{-1}$ \\
                  & \(M(z)\) & Truncated moment-generating function \\
                  & \(\chi(z)\) & Functional inverse of the truncated moment-generating function \\
                  & \(r_k(\nu)\) & \(k\)-th free cumulant of the measure \(\nu\) \\
                  & \(\mu_p, \mu_p(t)\) & \(p\)-th moment of a probability measure (and its time evolution) \\
                  & \(\boxplus\) & Free additive convolution \\
                  & \(\boxtimes\) & Free multiplicative convolution \\
    \addlinespace[1mm]  
    \arrayrulecolor{gray!30}\cline{1-3} 
    \addlinespace[1mm]
    \arrayrulecolor{black}
    \multirow{2}{*}{\makecell[l]{Algebraic\\Curve}}
                  & \(P(z,m)\) & Algebraic equation for the Stieltjes transform, \(P(z,m(z)) = 0\) \\
                  & \(\mathcal{C}\) & Complex 1D spectral curve \(\mathcal{C} = \{(z, m) \in \mathbb{C}^2 \mid P(z, m) = 0\}\) \\
    \addlinespace[1mm]  
    \arrayrulecolor{gray!30}\cline{1-3} 
    \addlinespace[1mm]
    \arrayrulecolor{black}
    \multirow{2}{*}{\makecell[l]{Deformed\\Decompression}}
                  & \(n\) & Size of the target (large) matrix \\
                  & \(n_0\) & Size of the sampled (small) sub-matrix, \(n_0 < n\) \\
                  & \(\tau, t\) & Decompression scale, \(\tau = \frac{n}{n_0}\), with \(t = \log(\tau)\) \\
                  & \(c\) & Initial aspect ratio; the target aspect ratio at scale \(\tau\) is \(\tau c\) \\
                  & \(m_1\), \(\underline{m}_1\) & Initial ordinary and companion Stieltjes transforms at \(\tau = 1\) \\
                  & \(m_{\tau}\), \(\underline{m}_{\tau}\) & Evolved ordinary and companion Stieltjes transforms at scale \(\tau\) \\
    \addlinespace[1mm]  
    \arrayrulecolor{gray!30}\cline{1-3} 
    \addlinespace[1mm]
    \arrayrulecolor{black}
    Notation     & \(\mathbb{C}^{\pm}\) & Upper (lower) half complex plane \\
                 & \(\Re, \Im\) & Real and imaginary part of complex variable \\
                 & \(\operatorname{p.v.}\) & Cauchy principal value \\
    \bottomrule
\end{tabular}
\end{table}

\begin{figure}[ht!]
    \centering
    \includegraphics[width=\textwidth]{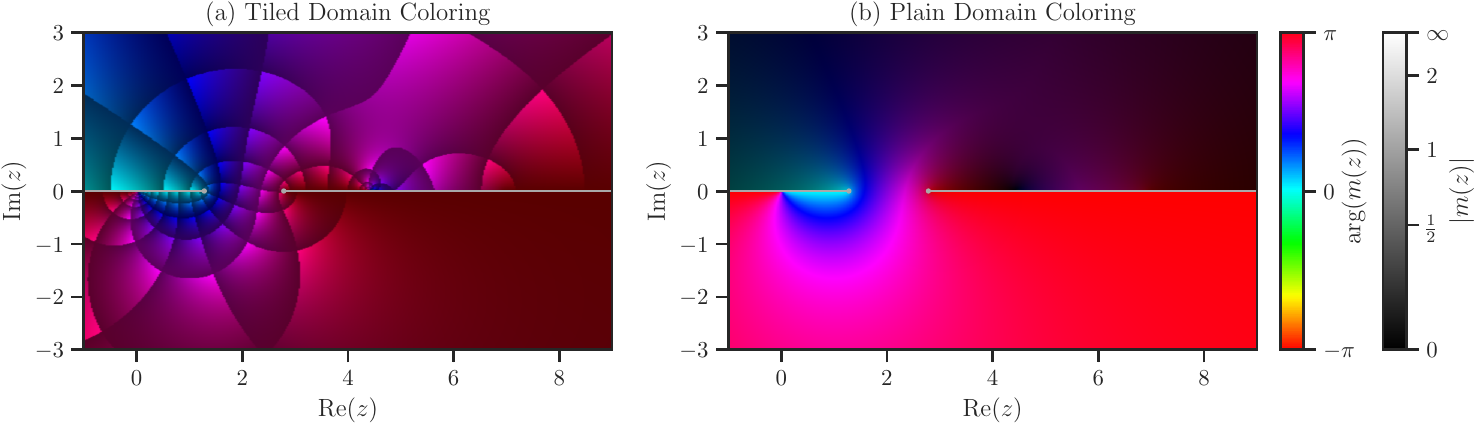}
    \caption{Domain coloring of a Stieltjes-transform branch for the deformed Marchenko--Pastur example in \Cref{fig:dmp-branches}. Right: plain domain coloring, with hue encoding \(\arg m(z)\) and brightness encoding \(|m(z)|\). Left: tiled domain coloring, where phase and logarithmic magnitude are periodically reset to make high-dynamic-range structure more visible.}
    \label{fig:domain-coloring-convention}
\end{figure}

\paragraph{Color code for plots.}
In our complex-plane plots, colors encode the values of a complex quantity using \emph{domain coloring}: hue represents phase, while brightness represents magnitude; see the plain version in \Cref{fig:domain-coloring-convention}(b). For high-dynamic-range fields, we also use the tiled version shown in \Cref{fig:domain-coloring-convention}(a), in which phase and logarithmic magnitude are periodically reset, following the phase-portrait convention of Wegert~\cite[Section~2.5]{WEGERT-2012}.

\section{Transforms in Random Matrix Theory} \label{sec:transforms}


Let \(\nu\) be a probability measure on \(\mathbb{R}\) with compact support \(\supp(\nu) \subset \mathbb{R}\). Let \(\nu_{\mathrm{ac}}\) denote its absolutely continuous part with respect to Lebesgue measure, and define \(\rho(x) \coloneqq \nu_{\mathrm{ac}}(\mathrm{d} x) / \mathrm{d} x\) as its density, given by the Radon--Nikodym derivative. In this section, we outline the various transforms of $\nu$ that each uniquely identify it. Our discussion will proceed formally without attention to domains of convergence. In the following, $\nu$ will be a fixed, but arbitrary, probability measure.

\begin{definition}[Stieltjes Transform]
    The \emph{Stieltjes transform} of \(\nu\) is the function \(m: \mathbb{C} \setminus \supp(\nu) \to \mathbb{C}\) defined by
    \begin{equation}
        m(z) \coloneqq \int_{\mathbb{R}} \frac{1}{x - z} \, \nu(\mathrm{d} x), \quad z \in \mathbb{C} \setminus \supp(\nu). \label{eq:def-stieltjes}
    \end{equation}
\end{definition}

The Stieltjes transform is a \emph{Herglotz} function, that is, \(\Im m(z) > 0\) for \(z \in \mathbb{C}^{+}\) where \(\mathbb{C}^{+} \coloneqq \{z \in \mathbb{C} \mid \Im z > 0\}\), and \(\Im w\) denotes the imaginary part for \(w \in \mathbb{C}\). Moreover, \(m\) satisfies the \emph{normalization} at infinity: \(m(z) = -z^{-1} + O(z^{-2})\) as \(\lvert z \rvert \to \infty\).


\paragraph{Recovering the Density.}
Whenever the limits exist, let \(m^{\pm}(x) \coloneqq \lim_{y \to 0^{\pm}} m(x + i y)\). If \(\nu(\mathrm{d} x) = \rho(x) \, \mathrm{d} x\) in a neighborhood of \(x\), the Sokhotski--Plemelj formula \citep[p.~42]{MUSKHELISHVILI-1992} yields
\begin{equation*}
    m^{\pm}(x) = -\pi \mathcal{H}\{\nu\}(x) \pm i \pi \rho(x),
\end{equation*}
where \(\mathcal{H}\{\nu\}(x) \coloneqq \frac{1}{\pi}\operatorname{p.v.}\int_{\mathbb{R}}\frac{\nu(\mathrm{d}t)}{x-t}\) is the \emph{Hilbert transform} of the full measure \(\nu\) and \(\mathrm{p.v.}\) denotes the Cauchy's principal value of a singular integral. The density can be recovered by
\begin{equation}
    \rho(x) = \pi^{-1} \Im m^{+}(x) = - \pi^{-1} \Im m^{-}(x),
    \label{eq:inv-stieltjes}
\end{equation}
which is the inversion of the Stieltjes transform. In the remainder, the support of the absolutely continuous part of \(\nu\) is denoted by \(I \coloneqq \supp(\rho)\), and is assumed to be comprised of finitely many disjoint compact intervals,
\begin{equation*}
    I = \bigcup_{j=1}^{k} I_j, \qquad I_j \coloneqq [a_j, b_j],
\end{equation*}
where \(k \geq 1\), and \(a_j < b_j < a_{j+1}\). Here, \(k\) is the number of spectrum bulks. Atomic components of \(\nu\) produce isolated poles of \(m\) on the real axis and must be treated separately.


The Stieltjes transform $m$ is one example of a transform that uniquely prescribes the underlying probability measure $\nu$. There are several other important transforms that are all related to the Stieltjes transform by algebraic relations. All transforms are to be discussed with reference to the same measure $\nu$, hence the dependence on the measure is dropped from the notation. In other sections, it will be convenient to state transforms with respect to particular measures using subscripts.

The first transform is the \emph{R-transform} which is required for the statement of \Cref{thm:main}. Our presentation follows \citet[Definition 5]{COUILLET-2022}. 

\begin{definition}[R-transform]
    The \emph{R-transform} of $\nu$ is defined as the solution $R(z)$ to
    \[
    m(R(z) + z^{-1}) = -z,
    \]
    for all $z$ where such a solution exists. Since $m(z) \sim -1/z$ as $|z|\to\infty$, $R(z)$ is defined in a neighbourhood of $z = 0$ where it has a power series expansion in terms of the \emph{free cumulants} $r_k$:
    \[
        R(z) = \sum_{k=1}^{\infty} r_k z^{k-1}.
    \]
\end{definition}

The primary significance of the R-transform in random matrix theory is its behavior under free additive convolution. For two measures $\nu_1$ and $\nu_2$, the free additive convolution $\nu_1 \boxplus \nu_2$ (representing the distribution of the eigenvalues of the sum of freely independent matrices with spectral distributions $\nu_1$ and $\nu_2$) has R-transform satisfying \citep[Lemma 2.19]{COUILLET-2022}
\[
    R_{\nu_1 \boxplus \nu_2}(z) = R_{\nu_1}(z) + R_{\nu_2}(z).
\]
On some occasions, the algebraic manipulations render directly working with the R-transform notationally inconvenient. This is often rectified by working with the K-transform instead.

\begin{definition}[K-transform]
    The \emph{K-transform} is the functional inverse of $-m$, that is, $-m(K(w)) = w$. It is defined in terms of the R-transform by
    \[
        K(w) = \frac{1}{w} + R(w).
    \]
\end{definition}

The Stieltjes transform has a convenient Laurent series expansion for large $|z|$ in terms of the moments $\mu_n = \int x^n \nu(\mathrm{d}x)$ of $\nu$:
\begin{equation}
    \label{eq:stieltjes-moment-expansion}
    m(z) = -\frac{1}{z} \sum_{n=0}^{\infty} \frac{\mu_n}{z^n},
\end{equation}

Therefore, a closely related transform to the Stieltjes transform is the (truncated) moment generating function, which is a direct power series in the moments.

\begin{definition}[Moment Generating Function]
    For a fixed probability measure $\nu$, we let $\mu_n = \int x^n \nu(\mathrm{d} x)$ denote the $n$-th moment of $\nu$ for integer $n$. The \emph{(truncated) moment generating function} of $\nu$ is given by
    \[
        M(z) = \sum_{n=1}^\infty \mu_n z^n = \int_{\mathbb{R}} \frac{t z}{1 - t z} \, \nu(\mathrm{d} t),
    \]
\end{definition}

Here, the integral representation follows from the geometric series. 
From \eqref{eq:stieltjes-moment-expansion}, it is clear that
\begin{equation}
    \label{eq:stieltjes-to-mgf}
    m(z) = -\frac{1}{z}(1 + M(z^{-1})), \qquad M(z) = -1 - z^{-1}m(z^{-1}).
\end{equation}

The moment generating function is the first transform on the path to the S-transform. The intermediate stage is referred to as the chi transform.

\begin{definition}[Chi Transform]
    The chi transform $\chi(z)$ is the functional inverse of the moment generating function $M(z)$ so that $M(\chi(z)) = \chi(M(z)) = z$. As with the R-transform, this is defined formally, and is to be understood as defined over the domain where $M$ is invertible. 
\end{definition}
    
\begin{definition}[S-transform]
    The S-transform $S(z)$ is given in terms of the chi transform by
    \[
        S(z) = \frac{z+1}{z}\chi(z).
    \]
\end{definition}

The S-transform is related to the moment generating function $M(z)$ by
\begin{equation}
    \label{eq:s-to-m-transform}
    z = \frac{M(z)S(M(z))}{1 + M(z)}.
\end{equation}
Analogous to the R-transform, the significance of the S-transform is its behavior for multiplicative free convolution. For two measures $\nu_1$ and $\nu_2$, one of which has support entirely on $[0,\infty)$, the free multiplicative convolution $\nu_1 \boxtimes \nu_2$ (representing the distribution of the eigenvalues of the symmetric product $\tens{A}^{1/2} \tens{B} \tens{A}^{1/2}$ of freely independent matrices $\tens{A}$, $\tens{B}$ with spectral distributions $\nu_1$ and $\nu_2$, respectively) has S-transform satisfying \citep[Lemma 2.19]{COUILLET-2022}
\[
    S_{\nu_1 \boxtimes \nu_2}(z) = S_{\nu_1}(z) S_{\nu_2}(z).
\]
The R- and S-transforms are related by the following \citep[Lemma F.1]{AMELI-2026}: 
\begin{equation}
    \label{eq:r-to-s}
    y = z S(z) \iff z = y R(y).
\end{equation}


\paragraph{Companion Matrix.}
A critical object in our exposition is the \emph{companion matrix}: for a matrix $\tens{S} = \tens{X} \tens{X}^{\intercal}$, the companion matrix of $\tens{S}$ is $\underline{\tens{S}} = \tens{X}^{\intercal} \tens{X}$. For rectangular $\tens{X}$, the shape of $\underline{\tens{S}}$ will differ from $\tens{S}$, but by the Weinstein--Aronszajn identity, their eigenspectra only differ in the number of zeros. In terms of the Stieltjes transform, this corresponds to the inclusion/removal of an atom at zero. The weight of this atom is determined by the aspect ratio of the sides of $\tens{X}$, denoted by $c > 0$. 

\begin{definition}[Companion Stieltjes Transform]
    For a fixed aspect ratio $c > 0$, the Stieltjes transform of the \emph{companion} $\underline{m}$ is given in terms of the Stieltjes transform $m$ by
    \[
        \underline{m}(z) = c m(z) + \frac{c - 1}{z}.
    \]
\end{definition}

Here and throughout, the underline notation will be used to represent the corresponding object for the companion.

There is no straightforward relationship between the R-transform or K-transform and that of the companion. However, for the other transforms we have the following lemma.

\begin{lemma}
    \label{lem:companion-transforms}
    For a given aspect ratio $c > 0$, the companion moment generating function $\underline{M}$, chi transform $\underline{\chi}$, and S-transform $\underline{S}$ satisfy
    \[
        \underline{M}(z) = c M(z),\qquad \underline{\chi}(z) = \chi(z/c),\qquad \underline{S}(z) = \frac{z+1}{z+c}S\left(\frac{z}{c}\right).
    \]
\end{lemma}

\begin{proof}
    Since $m(z) = -z^{-1} (1 + M(z^{-1}))$ and likewise for $\underline{m}(z)$,
    \[
        -\frac{1}{z}(1+\underline{M}(z^{-1}))=-\frac{c}{z}(1+M(z^{-1}))+\frac{c-1}{z},
    \]
    and by simplifying terms, we find that $\underline{M}(z^{-1}) = cM(z^{-1})$. For the chi transform, since $z = \underline{M}(\underline{\chi}(z)) = c M(\underline{\chi}(z))$, it follows that $\chi(z/c) = \underline{\chi}(z)$. Finally,
    \[
        \underline{S}(z)=\frac{z+1}{z}\underline{\chi}(z)=\frac{z+1}{z}\chi\left(\frac{z}{c}\right)=\frac{z+1}{z}\cdot\frac{z/c}{z/c+1}S\left(\frac{z}{c}\right)=\frac{z+1}{z+c}S\left(\frac{z}{c}\right).
    \]
\end{proof}


\section{Deformed Decompression on Algebraic Spectral Curves}
\label{sec:dd-algebraic}

In this section, we cover the mathematical fundamentals for deformed decompression. Our exposition begins in \Cref{sec:key} by proving our key \Cref{thm:main}, both in terms of free probability, and more rigorously using the Silverstein equation. \Cref{sec:free-binom} proves that, for $0 < c \leq 1$, every probability measure with a well-defined S-transform admits forward deformed decompression for any $\tau \geq 1$. The PDEs in \Cref{thm:pde-all} are derived in \Cref{sec:pde-deriv}. Using these PDEs, \Cref{sec:mp-only} shows that the Marchenko--Pastur distribution is the only probability measure that correctly evolves under both free decompression and deformed decompression, implying that the two operations are fundamentally distinct. 

Our more practical exposition begins in \Cref{sec:formal} with a derivation of the spectral curves along which the evolved companion Stieltjes transform $\underline{m}_\tau$ lie. \Cref{sec:deformedcoeffmethod} develops a polynomial relation for $\underline{m}_\tau$ similar to \citet{AMELI-2026}, although we find this to be primarily a theoretical curiosity with limited practical utility. Instead, solving for $\underline{m}_\tau$ is achieved using a predictor-corrector procedure described in \Cref{sec:dd-on-sc}.


\subsection{The Key Theorem}
\label{sec:key}

Our exposition begins with a complete proof of the key S-transform relationship which underpins the deformed decompression procedure. Recall that our generalized covariance matrix $\tens{S}$ and its companion $\underline{\tens{S}}$ satisfy
\[
    \tens{S} = \frac{1}{n}\gtens{\Sigma}_p^{1/2} \tens{Z}\tens{Z}^\intercal \gtens{\Sigma}_p^{1/2},\qquad \underline{\tens{S}} = \frac{1}{n} \tens{Z}^\intercal \gtens{\Sigma}_p \tens{Z}.
\]

It is a classical result in free probability theory that if $\tens{Z}$ is comprised of independent standard normal random variables, under \Cref{ass:pop}, the spectral distribution of $\tens{S}$ converges to the free multiplicative convolution of $\nu_{\tens{\Sigma}}$ and a Marchenko-Pastur distribution, that is,
\[
    \nu_{\tens{S}} = \nu_{\tens{\Sigma}} \boxtimes \nu_{\mathrm{MP}_c}.
\]
In terms of the S-transform which factorizes free multiplicative convolution,
\[
    S_{\tens{S}}(w) = S_{\tens{\Sigma}}(w) S_{\mathrm{MP}_c}(w) = \frac{1}{1 + c w} S_{\tens{\Sigma}}(w).
\]
The free probability proofs of this result rely strongly on the unitary invariance of $\tens{Z}$ to establish that $\{\{\tens{Z}_n\}_{n=1}^{\infty},\{\tens{Z}_n^\intercal\}_{n=1}^{\infty}\}$ and $\{\tens{\Sigma}_{p(n)}\}_{n=1}^{\infty}$ are asymptotically free \citep[Theorem 2.3]{VOICULESCU-1991}. However, it turns out that the asymptotic distribution does not rely on the precise nature of the random variables which comprise $\tens{Z}$. If the elements of $\tens{Z}$ are instead only independent and identically distributed with unit variance, universality takes hold and the resulting limiting spectral distribution of $\tens{S}$ turns out the same. The proof of this particular finding is due to \citet{SILVERSTEIN-1995} and follows a much different approach involving careful bounds on deterministic equivalents. A modern discussion of the proof is presented in \citet{COUILLET-2022}. We present the complete result below in the form most convenient for our purposes.

\begin{theorem}[Silverstein]
    Let $\tens{Z}_n$ be a $p \times n$ matrix for $p \equiv p(n)$ comprised of independent and identically distributed real-valued entries with unit variance. Let $\tens{\Sigma}_p$ be a $p \times p$ Hermitian non-negative definite deterministic matrix. Define $\tens{X}_n = \frac{1}{\sqrt{n}}\tens{\Sigma}_p^{1/2}\tens{Z}_n$, and let the sample covariance matrix be $\tens{S}_n = \tens{X}_n \tens{X}_n^\intercal$. Under the limit $n \to \infty$ with $p(n) / n \to c \in (0,\infty)$ and \Cref{ass:pop}, almost surely, the ESD of $\tens{S}_n$ converges weakly to a deterministic probability measure $\nu$ with S-transform satisfying
    \[
        S(w) = \frac{1}{1 + c w} S_{\tens{\Sigma}}(w). 
    \]
\end{theorem}

\begin{proof}
    This theorem is a restatement of the original equation of \citet{SILVERSTEIN-1995}. Indeed, by \citet[Theorem 1.1]{SILVERSTEIN-1995}, under the stated hypotheses, the ESD of $\tens{S}_n$ converges weakly to a deterministic probability measure $\nu$ whose Stieltjes transform satisfies the fixed point equation
    \begin{equation}
        \label{eq:silverstein-orig}
        m(z) = \int \frac{1}{t(1-c-czm(z)) - z} \nu_{\tens{\Sigma}}(\mathrm{d}t).
    \end{equation}
    Now, since $m(z) = -z^{-1}(1+M(z^{-1}))$ where $M$ is the moment generating function, $1 - c - czm(z) = 1 + cM(z^{-1})$. Substituting this back into \eqref{eq:silverstein-orig} gives
    \[
        m(z)=\int\frac{1}{t(1+cM(z^{-1}))-z}\nu_{\tens{\Sigma}}(\mathrm{d}t).
    \]
    Performing the change of variables $z \mapsto z^{-1}$ and using the identity $M(z) = -1 - z^{-1}m(z^{-1})$ reveals
    \begin{align*}
        M(z) &= -1+\int\frac{1}{1-tz(1+cM(z))}\nu_{\tens{\Sigma}}(\mathrm{d}t)\\
        &= \int\frac{tz(1+cM(z))}{1-tz(1+cM(z))}\nu_{\tens{\Sigma}}(\mathrm{d}t).
    \end{align*}
    But recalling the integral representation for $M_{\tens{\Sigma}}$, this implies
    \[
        M(z) = M_{\tens{\Sigma}}(z (1 + cM(z))).
    \]
    Now, let $w = M(z)$ so that $z = \chi(w)$. Then $w = M_{\tens{\Sigma}}(\chi(w)(1+cw))$ which implies $\chi_{\tens{\Sigma}}(w) = \chi(w)(1+cw)$. Immediately, we find that $S_{\tens{\Sigma}}(w) = S(w)(1+cw)$, which implies the result.
\end{proof}

Since $S_{\tens{\Sigma}}$ is independent of the aspect ratio, the S-transform for the measure $\nu_c$ (the spectral distribution of $\tens{S}$ with aspect ratio $c$), which we shall denote by $S$, is related to the S-transform for the measure $\nu_{\tau c}$ for some decompression ratio $\tau > 1$ by
\[
    S(w)(1+cw) = S_\tau(w) (1 + c \tau w).
\]
Consequently, solving for $S_\tau$ gives \bemph{the S-transform relation for deformed decompression}:
\begin{equation}
    \label{eq:deformed-dec-s}
    S_\tau(w) = \frac{1+cw}{1+c\tau w} S(w).
\end{equation}
Finally, to get to \Cref{thm:main} from \eqref{eq:deformed-dec-s}, we make use of the following proposition.

\begin{proposition}\label{prop:r_companion}
    The following three relationships are equivalent.
    \begin{enumerate}[label=(\roman*)]
        \item $S_\tau(w) = \displaystyle\frac{1 + cw}{1 + c\tau w} S(w)$
        \item $\underline{S}_\tau(w) = \tau^{-1} \underline{S}(w / \tau)$
        \item $\underline{R}_\tau(z) = \tau \underline{R}(z)$
    \end{enumerate}
\end{proposition}

\begin{proof}
    First, we show that (i) $\implies$ (ii). From \Cref{lem:companion-transforms}, 
    \[
        \underline{S}_{\tau}(w)=\frac{w+1}{w+c\tau}S_{\tau}\left(\frac{w}{c\tau}\right)=\frac{w+1}{w+c\tau}\cdot\frac{1+w/\tau}{1+w}S\left(\frac{w}{c\tau}\right)=\frac{1}{\tau}\cdot\frac{w/\tau+1}{w/\tau+c}S\left(\frac{w}{c\tau}\right),
    \]
    but since $\underline{S}_{1}(w)=\frac{w+1}{w+c}S(w/c)$, we obtain (ii). Now, (ii) implies that $z \coloneqq w \underline{S}_\tau(w) = w/\tau \underline{S}(w /\tau)$. But recall from \eqref{eq:r-to-s} that this implies $z \underline{R}_\tau(z) = w$ and $z \underline{R}(z) = w/\tau$. Equating in $w$,
    $z\underline{R}(z) = \frac{1}{\tau}z\underline{R}_\tau(z)$, from whence (iii) is immediate. Since all operations presented are formally reversible, this also implies that (iii) $\implies$ (ii) $\implies$ (i).
\end{proof}


\subsection{The Free Binomial Distribution}
\label{sec:free-binom}

In this section, we show that for $0 < c \leq 1$, \emph{every} probability measure with a well-defined S-transform admits forward deformed decompression for any $\tau \geq 1$. This is in stark contrast to free decompression outlined in \citet{AMELI-2025b,AMELI-2026}. Indeed, a probability measure is only infinitely freely decompressible if it is freely infinitely divisible \citep[Proposition F.3]{AMELI-2026}, which is a fairly restrictive class \citep{BERCOVICI-1993a}. The main tool in this proof is the construction of the \emph{free binomial distribution} which has precisely the S-transform required to ensure that deformed decompression is simply free multiplicative convolution with the free binomial distribution. 

\begin{definition}
Let $c \in (0,1]$ and $\tau \geq 1$. The \emph{free binomial distribution} $\beta_{c,\tau}$ is the probability measure with S-transform 
\[
    S_{c,\tau}(w) = \frac{1+cw}{1+c\tau w}.
\]
\end{definition}

In \Cref{prop:free-binom} we show that the free binomial distribution exists and describe its properties.

\begin{proposition}
    \label{prop:free-binom}
    The free binomial distribution is well-defined and is given by
    \[
        \mathrm{d}\beta_{c,\tau} = \rho_{c,\tau} \mathrm{d} x + \max\{0, 1 - (c\tau)^{-1}\} \delta_0 + \max\{0, 1 - c^{-1} + (c\tau)^{-1}\} \delta_\tau,
    \]
    where $\delta_x$ is an atomic measure at $x$ and $\rho$ is the density
    \[
        \rho(x) = \frac{\sqrt{(x_{+} - x)(x - x_{-})}}{2 \pi c x (\tau - x)},\quad x \in [x_{-}, x_{+}],
    \]
    where $x_{\pm} = 1 + c \tau - 2 c \pm 2 \sqrt{c(1-c)(\tau-1)}$ prescribe the edges of the support. It also has Stieltjes transform $m_{c,\tau}$ satisfying
    \begin{equation}
        \label{eq:fbd-stieltjes}
        c z(z-\tau) \, m_{c,\tau}(z)^2 + ((2c - 1)z + (1 - c \tau)) \, m_{c,\tau}(z) - (1-c) = 0.
    \end{equation}
\end{proposition}

\begin{proof}
    To establish existence, let \(P\) and \(Q\) be freely independent projections with traces \(c\) and \(\tau^{-1}\). Let \(\beta_{c,\tau}\) be the spectral law of \(\tau P Q P\) on the range of \(P\), equipped with the normalized trace. Since the S-transform of a projection with trace \(a\) is \((1+w)/(a+w)\), multiplicativity and the compression relation give
    \begin{equation*}
        S_{\beta_{c,\tau}}(w)
        = \frac{c(1+w)}{\tau(1+cw)} S_P(cw)S_Q(cw)
        = \frac{1+cw}{1+c\tau w}.
    \end{equation*}
    Thus, \(\beta_{c,\tau}\) is a probability measure with the required S-transform. 

    We next derive its density and atoms. Here, we drop the $c, \tau$ subscript for brevity. Using \eqref{eq:s-to-m-transform}, we can derive an algebraic relation for the moment generating function. Indeed,
    \[
        M(z)(1+cM(z)) = z(1 + M(z))(1+c\tau M(z)),
    \]
    and so
    \[
        M(z^{-1})(1 + cM(z^{-1})) = z^{-1}(1 + M(z^{-1}))(1 + c\tau M(z^{-1})).
    \]
    From \eqref{eq:stieltjes-to-mgf}, it follows that
    \[
        (zm(z)+1)(1-czm(z)-c)=m(z)(1-c\tau zm(z)-c\tau).
    \]
    Expanding reveals that
    \[
        m(z)+1-cz^{2}m(z)^{2}-2czm(z)-c=m(z)-c\tau zm(z)^{2}-c\tau m(z),
    \]
    and rearranging terms arrives at \eqref{eq:fbd-stieltjes}. Solving using the quadratic formula reveals two possible branches for the Stieltjes transform:
    \[
        m_{c,\tau}(z)=\frac{-((2c-1)z+(1-c\tau))\pm\sqrt{\Delta(z)}}{2cz(z-\tau)},
    \]
    where $\Delta(z) = ((2c-1)z+(1-c\tau))^{2}+4c(1-c)z(z-\tau)$ is the discriminant. Simplifying,
    \[
        \Delta(z) = z^{2}+2(2c-1-c\tau)z+(1-c\tau)^{2}.
    \]
    For $z$ on the real line, $\Delta(z) < 0$ if $z \in [x_{-}, x_{+}]$ where
    \begin{align*}
        x_{\pm}&=1+c\tau-2c\pm\sqrt{(2c-1-c\tau)^{2}-(1-c\tau)^{2}}\\
        &= 1+c\tau-2c\pm2\sqrt{c(1-c)(\tau-1)}
    \end{align*}
    This prescribes the support of the density. Possible atoms occur where the highest-order coefficient in $m$ vanishes, namely at $z = 0$ and $z = \tau$. Their masses, when positive, are given by $((2c-1)z+(1-c\tau))/(2cz-c\tau)$ evaluated at the corresponding locations. By the Stieltjes inversion formula, the density $\rho$ in the interval $[x_{-},x_{+}]$ is given by
    \[
        \rho(x) = \frac{\sqrt{-\Delta(x)}}{2\pi c x(\tau - x)},
    \]
    which completes the proof. 
\end{proof}

Using the free binomial distribution, it is straightforward to show that all measures are infinitely deformed decompressible.

\begin{proposition}[All Measures are Decompressible]
    \label{prop:all-decomp}
    Let $\nu$ be an arbitrary probability measure with S-transform $S$. For any $c \in (0,1]$ and $\tau \geq 1$, there exists a probability measure $\nu_\tau$ with S-transform $S_\tau$ such that $(1 + c \tau w) S_\tau(w) = (1 + c w)S(w)$.
\end{proposition}

\begin{proof}
    From \Cref{prop:free-binom}, for any $c \in (0,1]$ and $\tau \geq 1$, there exists a probability measure $\beta_{c,\tau}$ such that the S-transform of $\beta_{c, \tau}$ is given by $S_{c, \tau}(w) = \frac{1+cw}{1+c\tau w}$. The construction $\nu_\tau = \beta_{c, \tau} \boxtimes \nu$ is always well-defined \citep{ARIZMENDI-2026} and has S-transform of the desired form. 
\end{proof}

The construction above guarantees forward evolution for $0 < c \leq 1$ and $\tau \geq 1$; outside this range, $S_{c,\tau}$ need not correspond to a probability measure. This does not imply that $S_\tau(w)$ is necessarily invalid (for example, if $\nu$ is a Marchenko--Pastur distribution, then $S_\tau(w)$ is a valid S-transform), but suggests that additional conditions on the initial law are required for the reverse operation. This is consistent with the relative ill-posedness of recovering the underlying population spectral distribution by taking $\tau \to 0$.


\subsection{Deriving the Partial Differential Equations} 
\label{sec:pde-deriv}

Let $\nu$ be a probability measure and let $\{\nu_\tau\}_{\tau \geq 1}$ be a family of probability measures obtained by deformed decompression such that for each $\tau$, denoting by $\underline{R}_\tau$ the companion R-transform of $\nu_\tau$, $\underline{R}_\tau(z) = \tau \underline{R}(z)$, where $\underline{R}$ is the companion R-transform of $\nu$. For each $\tau \geq 1$, let $m_\tau$ denote the Stieltjes transform corresponding to $\nu_\tau$, and $\underline{m}_\tau$ its companion Stieltjes transform. In the following, we look to obtain PDEs to describe the evolution of the deformed decompression flow. Our derivations begin with the PDE for the companion Stieltjes transform. Similar to \citet{AMELI-2025b} in the case of free decompression, it is most convenient to express this first equation under the change of variables $\tau = e^t \in [1, \infty)$. 

\begin{proposition}
    \label{prop:companion-pde}
    The companion Stieltjes transforms $\underline{m}(t, z) = \underline{m}_{e^t}(z)$ satisfy the PDE
    \[
        \frac{\partial\underline{m}}{\partial t} + \frac{\partial \underline{m}}{\partial z}\left(z + \frac{1}{\underline{m}}\right) = 0.
    \]
\end{proposition}

\begin{proof}
    For brevity, we let $\underline{R}(t,z) = \underline{R}_{e^t}(z)$ and consider the K-transform $K(t,z) = \underline{R}(t,z) + z^{-1}$. Since $\underline{R}(t,z) = e^t \underline{R}(0,z)$, it follows that
    \[
        \frac{\partial K}{\partial t}(t, z) = e^t \underline{R}(0, z) = K(t, z) - \frac{1}{z}.
    \]
    Since $-\underline{m}(t, K(t,w)) = w$,
    \[
        \frac{\partial\underline{m}}{\partial t}(t,K(t,w))+\frac{\partial\underline{m}}{\partial z}(t,K(t,w))\left(K(t,w)-\frac{1}{w}\right)=0.
    \]
    Letting $w = -\underline{m}(t,z)$ recovers the result.
\end{proof}

Next, we consider the Stieltjes transform itself. This particular equation is best expressed in the original $\tau$ parameterization. A change of variables shows that
\begin{equation}
    \label{eq:companion-pde-tau}
    \tau \underline{m} \frac{\partial \underline{m}}{\partial \tau} + \frac{\partial \underline{m}}{\partial z}(\underline{m} z + 1) = 0
\end{equation}

\begin{proposition}
    \label{prop:stieltjes-pde}
    The Stieltjes transforms $m_\tau(z)$ satisfy the PDE
    \begin{equation}
        \label{eq:stieltjes-pde-prop}
        (c\tau z m + c \tau - 1)\frac{\partial m}{\partial \tau} + c (z m + 1) \left(z \frac{\partial m}{\partial z} + m\right) = 0.
    \end{equation}
\end{proposition}

\begin{proof}
    Recalling that $\underline{m} = c \tau m + (c \tau - 1) / z$,
    \begin{align*}
        \frac{\partial \underline{m}}{\partial \tau} &= c m + c \tau \frac{\partial m}{\partial \tau} + \frac{c}{z} \\
        \frac{\partial \underline{m}}{\partial z} &= c\tau \frac{\partial m}{\partial z} - \frac{c\tau - 1}{z^2} \\
        z \underline{m} + 1 &= c \tau (z m + 1).
    \end{align*}
    For the terms in the PDE \eqref{eq:companion-pde-tau}, we have
    \begin{align*}
        \tau \underline{m}\frac{\partial \underline{m}}{\partial \tau} &= c \tau\left(c \tau m + \frac{c \tau - 1}{z}\right)\left(m + \tau \frac{\partial m}{\partial \tau} + \frac{1}{z}\right)\\
        \frac{\partial \underline{m}}{\partial z} (z \underline{m} + 1) &= \left(c\tau \frac{\partial m}{\partial z} - \frac{c\tau - 1}{z^2}\right) c\tau (zm + 1).
    \end{align*}
    Substituting into the PDE from \eqref{eq:companion-pde-tau} and dividing by $c \tau$ gives
    \begin{equation*}
        \left(c \tau m + \frac{c \tau - 1}{z}\right)\left(m + \tau \frac{\partial m}{\partial \tau} + \frac{1}{z}\right) + \left(c\tau \frac{\partial m}{\partial z} - \frac{c\tau - 1}{z^2}\right) (zm + 1) = 0.
    \end{equation*}
    Multiplying by $z$ to clear the fractions, we obtain
    \begin{equation*}
        (c \tau z m + c \tau - 1)\left(m + \tau \frac{\partial m}{\partial \tau} + \frac{1}{z}\right) + \left(c\tau z \frac{\partial m}{\partial z} - \frac{c\tau - 1}{z}\right)(zm + 1) = 0.
    \end{equation*}
    Expanding the products yields
    \begin{align*}
        &(c \tau z m + c \tau - 1)m + (c \tau z m + c \tau - 1)\tau \frac{\partial m}{\partial \tau} + c \tau m + \frac{c\tau - 1}{z} \\
        &+ c\tau z (zm + 1) \frac{\partial m}{\partial z} - (c\tau - 1)m - \frac{c\tau - 1}{z} = 0.
    \end{align*}
    Observe that the terms $(c\tau - 1)m$ and $\frac{c\tau - 1}{z}$ cancel and the remaining non-derivative terms simplify to $c \tau z m^2 + c \tau m = c \tau m(zm + 1)$. After these reductions, we find
    \begin{equation*}
        (c \tau z m + c \tau - 1)\tau \frac{\partial m}{\partial \tau} + c \tau z(zm + 1) \frac{\partial m}{\partial z} + c \tau m(zm + 1) = 0.
    \end{equation*}
    Finally, dividing by $\tau$ and factoring $c(zm + 1)$ from the last two terms, we arrive at the result.
\end{proof}


\subsection{Deformed Decompression and Free Decompression}
\label{sec:mp-only}

Given that deformed decompression appears to display similarities in form to free decompression given in \citet{AMELI-2025b}, in this section, we discuss when the two operations coincide. It turns out that deformed decompression and free decompression coincide only in the special case where the initial distribution $\nu$ is a scaled Marchenko--Pastur distribution. 

\begin{proposition}
    Suppose that $m_\tau(z)$ satisfies both the deformed decompression flow \eqref{eq:stieltjes-pde-prop} and the free decompression flow given by the PDE
    \begin{equation}
        \label{eq:free-decomp-pde}
    \tau m \frac{\partial m}{\partial \tau} = -m^2 + \frac{\partial m}{\partial z}.
    \end{equation}
    Then there exists some $\sigma^2 \geq 0$ and some $c$ such that for $\lambda = c \tau$,
    \begin{equation}
        \label{eq:mp-law-quad}
        \lambda \sigma^2 z m_\tau^2 + (z - \sigma^2(1-\lambda)) m_\tau + 1 = 0,
    \end{equation}
    which is the defining relation for the Stieltjes transform of the Marchenko-Pastur distribution with aspect ratio $\lambda$ and variance parameter $\sigma^2$.
\end{proposition}

\begin{proof}
    From \eqref{eq:free-decomp-pde}, we find that $\frac{\partial m}{\partial \tau} = \frac{1}{\tau m} (\frac{\partial m}{\partial z} - m^2)$. Substituting this into \eqref{eq:stieltjes-pde-prop} yields the ordinary differential equation
    \[
        (c\tau z m + c \tau - 1)(m' - m^2) + \tau m c(m z + 1)(z m' + m) = 0.
    \]
    Expanding and simplifying terms reveals
    \[
        \frac{\mathrm{d}m}{\mathrm{d}z} = \frac{m^2}{1 - c \tau(mz + 1)^2},
    \]
    which is separable. However, observe that it will be far more convenient for the numerator and denominator in this expression to be reversed. To allow for this, consider instead the inverse function $z(m)$ which satisfies
    \[
        \frac{\mathrm{d}z}{\mathrm{d}m} = \frac{1 - c\tau (mz+1)^2}{m^2}.
    \]
    Now let $v(m) = mz(m) + 1$ so that
    \[
        \frac{\mathrm{d}v}{\mathrm{d}m} = z + m \frac{\mathrm{d}z}{\mathrm{d}m} = z+\frac{1-c\tau v^{2}}{m} =\frac{v}{m}(1-c\tau v).
    \]
    Since the $m \equiv 0$ case is ruled out, consider the $v \equiv 0$ case, which implies that $m(z) = -1/z$, satisfying \eqref{eq:mp-law-quad} with $\sigma^2 = 0$. If $c \tau v = 1$, then $m(z) = -\frac{1 - (c\tau)^{-1}}{z}$,
    which fails to be a Stieltjes transform of a probability measure. With these cases ruled out, we may solve for $v$. Since
    \[
        \frac{1}{v(1-c\tau v)} = \frac{1}{v} + \frac{c\tau}{1 - c\tau v},
    \]
    it follows that, for some constant $C$,
    \[
        \log v - \log (1 - c\tau v) = \log m + C.
    \]
    Letting $C = \log \sigma^2$ for some $\sigma^2 > 0$, we find that
    \[
        \frac{v}{1-\lambda v} = \sigma^2 m.
    \]
    Substituting $v = mz + 1$ and rearranging yields \eqref{eq:mp-law-quad}. 
\end{proof}


\subsection{Formal Deformed Decompression along Spectral Curves}
\label{sec:formal}

Now, recall that the functional relation for the R-transform in terms of the Stieltjes transform is given by \citet[Definition 5]{COUILLET-2022} as
\[
    R(-m(z)) = z + \frac{1}{m(z)}.
\]
From \Cref{thm:main}, the operation of deformed decompression coincides with the map $\nu \mapsto \nu_\tau$ where the companion R-transforms $\underline{R}$ and $\underline{R}_\tau$ of $\nu$ and $\nu_\tau$, respectively, satisfy
\[ 
    \underline{R}_\tau(z) = \tau \underline{R}(z),\qquad \tau > 1.
\]
If $m$ is defined to satisfy the formal algebraic relation $P(z,m) = 0$ for some polynomial $P$, then, formally, we can track the behavior of the Stieltjes transform of $\nu_\tau$ under deformed decompression. In doing so, we can track solutions to deformed decompression while avoiding any potential issues with $m$ and/or $R$ being multivalued. This is captured in \Cref{thm:formal-decomp}.

\begin{theorem} \label{thm:formal-decomp}
    For a polynomial $P(z,m)$, define the following spectral curves:
    \begin{align*}
        \mathcal{C}_{m} & =\left\{(z,m) \in \mathbb{C}^2 \mid P(z,m)=0\right\}, & \text{(graph of } m \text{)} \\ \
        \mathcal{C}_{\underline{m}} & = \left\{(z,w) \in \mathbb{C}^2 \mid (z,c^{-1}(w-(c-1)z^{-1}))\in\mathcal{C}_{m} \right\}, & \text{(graph of companion } \underline{m} \text{)}\\ \
        \mathcal{C}_{\underline{R}} &= \left\{(w,r) \in \mathbb{C}^2 \mid (r-1/w,w)\in\mathcal{C}_{\underline{m}} \right\}, & \text{(R-transform of } \underline{m} \text{)}\\ \
        \mathcal{C}_{\underline{R}_{\tau}} &= \left\{(w,r) \in \mathbb{C}^2 \mid (w,r/\tau)\in\mathcal{C}_{\underline{R}}\right\}, & \text{(R-transform of } \underline{m}_\tau \text{)}\\ \
        \mathcal{C}_{\underline{m}_{\tau}} &= \left\{(z,\omega) \in \mathbb{C}^2 \mid \omega\neq0\text{ and }(\omega,z+1/\omega)\in\mathcal{C}_{\underline{R}_{\tau}}\right\}. & \text{(graph of } \underline{m}_\tau \text{)}
    \end{align*}
    Then $$\mathcal{C}_{\underline{m}_\tau} \subseteq \mathcal{C}_{P_\tau} \coloneqq \left\{(z,\omega) \in \mathbb{C}^2 \,\middle|\, P\left(\zeta, \frac{1}{c}\left(\omega - \frac{c-1}{\zeta}\right)\right) = 0\quad\text{ where } \zeta = \frac{1}{\tau}\left(z+\frac{1}{\omega}\right) - \frac{1}{\omega}\right\}.$$
\end{theorem}

\begin{proof}
    We begin by unpacking the conditions for a point $(z,\omega)$ to belong to $\mathcal{C}_{\underline{m}_\tau}$. 
       By definition (excluding the $\omega = 0$ case), $(z, \omega) \in \mathcal{C}_{\underline{m}_\tau}$ if and only if
    \[
        (\omega, z + 1/\omega) \in \mathcal{C}_{\underline{R}_\tau}.
    \]
    According to the definition of $\mathcal{C}_{\underline{R}_\tau}$, this implies that
    \[
        \left(\omega, \frac{1}{\tau}\left(z + \frac{1}{\omega}\right)\right) \in \mathcal{C}_{\underline{R}}.
    \]
    By the definition of the companion R-transform curve $\mathcal{C}_{\underline{R}}$, setting $w = \omega$ and $r = \frac{1}{\tau}(z + \frac{1}{\omega})$, this implies that
    \[
        \left(\frac{1}{\tau}\left(z + \frac{1}{\omega}\right) - \frac{1}{\omega}, \omega\right) \in \mathcal{C}_{\underline{m}}.
    \]
    Denote the first coordinate as $\zeta$, so $\zeta = \frac{1}{\tau}(z + \frac{1}{\omega}) - \frac{1}{\omega}$. We now have $(\zeta, \omega) \in \mathcal{C}_{\underline{m}}$. 
    By the definition of the companion Stieltjes curve $\mathcal{C}_{\underline{m}}$, this means that there exists an underlying point in the original curve $\mathcal{C}_m$ satisfying
    \[
        \left(\zeta, \frac{1}{c}\left(\omega - \frac{c-1}{\zeta}\right)\right) \in \mathcal{C}_m.
    \]
    Finally, since belonging to $\mathcal{C}_m$ implies being a root of the annihilating polynomial $P$, we conclude that
    \[
        P\left(\zeta, \frac{1}{c}\left(\omega - \frac{c-1}{\zeta}\right)\right) = 0.
    \]
    Consequently, any $(z,\omega) \in \mathcal{C}_{\underline{m}_\tau}$ must satisfy this polynomial relation, meaning $(z,\omega) \in \mathcal{C}_{P_\tau}$.
\end{proof}

\Cref{thm:formal-decomp} immediately implies \Cref{thm:SpectralCurve}, as we identify $m_\tau$ to be the deformed decompression of $m$ by the factor $\tau$. 


\subsection{Polynomial Evolution under Deformed Decompression}
\label{sec:deformedcoeffmethod}

In this section, we use \Cref{thm:formal-decomp} to derive a bivariate polynomial $Q_\tau$ such that the companion Stieltjes transform $\underline{m}_\tau$ under deformed decompression satisfies $Q_\tau(z, \underline{m}_\tau(z)) = 0$ for almost all $z$. This process is performed analogously to \citet[Corollary F.5]{AMELI-2026} as a theoretical curiosity; we will find the resulting polynomial to exhibit high numerical instability in its roots. 

\begin{corollary} \label{cor:DeformedCoefficientMethod}
    Suppose that the standard Stieltjes transform $m: \mathbb{C}^+ \to \mathbb{C}^+$ satisfies an algebraic relation of the form $P(z, m(z)) = 0$ for $z \in \mathbb{C}^+$ where
    \[
        P(z,m) = \sum_{j=0}^{d_z} \sum_{k=0}^s c_{jk} z^j m^k.
    \]
    Then for $\mathcal{C}_{\underline{m}_\tau}$ defined in \Cref{thm:formal-decomp},
    \[
        \mathcal{C}_{\underline{m}_\tau} \subseteq \left\{ (z,\omega) \in \mathbb{C}^2 \, \middle|\, Q_\tau(z,\omega) = 0 \right\},
    \]
    where $Q_\tau(z,\omega)$ is the polynomial given by
    \[
        Q_\tau(z,\omega) \coloneqq \sum_{p=0}^{d_z+s} \sum_{q=0}^{d_z+2s} c_{pq}^{(\tau)} z^p \omega^q,
    \]
    and the coefficients $c_{pq}^{(\tau)}$ are defined as
    \[
        c_{pq}^{(\tau)} \coloneqq \sum_{j=0}^{d_z} c_{j,k} \tau^{-j} c^{-k} \sum_{u=0}^p \binom{s-k+j}{u} \binom{k}{p-u} (1-\tau)^{s-k+j-u} (1-\tau c)^{k-p+u},
    \]
    where $k = q - p + j - d_z$, and we adopt the convention that $c_{jk} = 0$ if $k \notin \{0,\dots,s\}$ and $\binom{n}{r} = 0$ if $r < 0$ or $r > n$.
\end{corollary}

\begin{proof}
    From \Cref{thm:formal-decomp}, we have the inclusion
    \[
        \mathcal{C}_{\underline{m}_\tau} \subseteq \left\{(z,\omega) \in \mathbb{C}^2 \,\middle|\, P\left(\zeta, \frac{1}{c}\left(\omega - \frac{c-1}{\zeta}\right)\right) = 0 \quad\text{where}\quad \zeta = \frac{1}{\tau}\left(z+\frac{1}{\omega}\right) - \frac{1}{\omega}\right\}.
    \]
    For $(z,\omega) \in \mathcal{C}_{\underline{m}_\tau}$, we have $\omega \neq 0$. We first simplify the arguments evaluated by the original polynomial $P$. The first argument $\zeta$ can be rewritten with a common denominator as
    \[
        \zeta = \frac{z}{\tau} + \frac{1-\tau}{\tau \omega} = \frac{z\omega + 1 - \tau}{\tau \omega}.
    \]
    Let $M$ denote the second argument of the polynomial $P$, which effectively maps back to the standard Stieltjes transform domain. Substituting our expression for $\zeta$ into $M$, we obtain
    \begin{align*}
        M &= \frac{1}{c}\left(\omega - \frac{c-1}{\frac{z\omega + 1 - \tau}{\tau \omega}}\right) = \frac{1}{c}\left(\omega - \frac{\tau \omega (c-1)}{z\omega + 1 - \tau}\right) \\
        &= \frac{\omega(z\omega + 1 - \tau) - \tau \omega(c-1)}{c(z\omega + 1 - \tau)} \\
        &= \frac{\omega(z\omega + 1 - \tau c)}{c(z\omega + 1 - \tau)}.
    \end{align*}
    Since $(z,\omega)$ originates from the curve ensuring these mapped coordinates are a root of $P$, substituting these rational functions into the polynomial expansion yields
    \[
        P(\zeta, M) = \sum_{j=0}^{d_z} \sum_{k=0}^s c_{jk} \left(\frac{z\omega + 1 - \tau}{\tau \omega}\right)^j \left(\frac{\omega(z\omega + 1 - \tau c)}{c(z\omega + 1 - \tau)}\right)^k = 0.
    \]
    To clear the denominators, we multiply the entire equation by $\omega^{d_z}(z\omega + 1 - \tau)^s$. Because the summation indices satisfy $j \le d_z$ and $k \le s$, this guarantees all terms become proper bivariate polynomials in $z$ and $\omega$:
    \[
        \omega^{d_z}(z\omega + 1 - \tau)^s P(\zeta, M) = \sum_{j=0}^{d_z} \sum_{k=0}^s c_{jk} \tau^{-j} c^{-k} \omega^{d_z-j+k} (z\omega + 1 - \tau)^{s-k+j} (z\omega + 1 - \tau c)^k.
    \]
    We define the left-hand side of this expression as the unnormalized polynomial $Q_\tau(z,\omega)$. To extract its explicit coefficients, we expand the binomial products in terms of $z\omega$. Using the binomial theorem on both factors,
    \begin{align*}
        (z\omega + 1 - \tau)^{s-k+j} (z\omega + 1 - \tau c)^k &= \left( \sum_{u=0}^{s-k+j} \binom{s-k+j}{u} (z\omega)^u (1-\tau)^{s-k+j-u} \right) \\&\times\left( \sum_{v=0}^k \binom{k}{v} (z\omega)^v (1-\tau c)^{k-v} \right).
    \end{align*}
    Taking the Cauchy product of these two sums, the coefficient of $(z\omega)^p$ where $p = u+v$ is given by
    \[
    C_{j,k,p} = \sum_{u=0}^p \binom{s-k+j}{u} \binom{k}{p-u} (1-\tau)^{s-k+j-u} (1-\tau c)^{k-p+u}.
    \]
    Substituting this combinatorial term back into our expression for $Q_\tau(z,\omega)$ gives:
    \[
        Q_\tau(z,\omega) = \sum_{j=0}^{d_z} \sum_{k=0}^s \sum_{p=0}^{s+j} c_{jk} \tau^{-j} c^{-k} C_{j,k,p} z^p \omega^{d_z-j+k+p}.
    \]
    Finally, we change summation variables to group the terms strictly by powers of $z$ and $\omega$. Let $p$ remain the power of $z$, and let $q = d_z - j + k + p$ be the power of $\omega$. Consequently, we can isolate $k$ as $k = q - p + j - d_z$. Reindexing the sums in terms of $p$ and $q$ yields the final polynomial form:
    \[
        Q_\tau(z,\omega) = \sum_{p=0}^{d_z+s} \sum_{q=0}^{d_z+2s} c_{pq}^{(\tau)} z^p \omega^q,
    \]
    as required.
\end{proof}

Unlike free decompression \citep{AMELI-2026}, we are unable to find conditions which guarantee that the degree of $Q_\tau$, once simplified, does not exceed the degree of $P$. Indeed, the substantial increase in the degree of $Q_\tau$ over the initial polynomial $P$ often renders $Q_\tau$ unusable in practice, as small numerical errors in the coefficients result in vast discrepancies in the roots.


\subsection{Deformed Decompression on Spectral Curves} \label{sec:dd-on-sc}

Analogous to \citet{AMELI-2026}, identifying the desired solution $\underline{m}_\tau$ to the deformed decompression flow as a root of the polynomial $Q_\tau$ is not sufficient for reconstructing $\underline{m}_\tau$, as it remains unclear which root to select. Unlike free decompression, however, deformed decompression benefits from a key geometric property. Biane's Subordination Theorem \cite{BIANE-1998} guarantees that the transport on the curve induced by deformed decompression maps the upper half plane into itself. Consequently, if we start on the physical branch, the continuation remains robustly physical. To identify this correct root, we track the known physical sheet at $\tau = 1$ along the deformed decompression flow, analogous to the geometric procedure used in \citet{AMELI-2026} for free decompression.

Let the initial fitted spectral curve be given by
\begin{equation*}
	\mathcal{C} = \left\{(\zeta, y) \in \mathbb{C}^2 \mid P(\zeta, y) = 0 \right\},
\end{equation*}
where $y = m_1(\zeta)$ denotes a branch of the initial ordinary Stieltjes transform. Let $c$ denote the initial aspect ratio,  so the target aspect ratio at scale \(\tau\) is \(\tau c\). For a fixed target query point $z$, let $w = \underline{m}_\tau(z)$ be the decompressed companion Stieltjes transform value. Along the spectral curves of the deformation, this value is conserved, so $w = \underline{m}_1(\zeta)$. The ordinary and companion transforms on the initial curve are related by $w = c y + \zeta^{-1}(c-1)$. From \Cref{thm:SpectralCurve}, the change of variables is given by
\begin{equation*}
	\zeta = \tau^{-1}(z + w^{-1}) - w^{-1}.
\end{equation*}
Substituting the relation for $w$ into this inverse map yields the evolved algebraic relation characterizing the candidate solutions. However, to track the physical solution directly, we cast $(\zeta, w)$ as a coupled system:
\begin{subequations}
\label{eq:dd-spectral}
\begin{align}
    F_1(\zeta, w) &\coloneqq P\left(\zeta, c^{-1}\left(w - \frac{c-1}{\zeta}\right)\right) = 0 \label{eq:dd-spectral-f1} \\
    F_2(\zeta, w; \tau, z) &\coloneqq \zeta - \tau^{-1} z - (\tau^{-1} - 1)w^{-1} = 0. \label{eq:dd-spectral-f2}
\end{align}
\end{subequations}
At $\tau = 1$, the system \eqref{eq:dd-spectral} reduces to $\zeta = z$ and $w = \underline{m}_1(z)$, where the initial physical point is already known from the branch selection procedure adopted from \citet{AMELI-2026}. The task is to continue this solution in $\tau$ while keeping $z$ fixed, using a predictor-corrector approach along the constrained curve.

Differentiating \eqref{eq:dd-spectral} with respect to $\tau$ provides the predictor direction. For $F_1 = 0$, we will define for brevity,
\begin{equation*}
	\underline{P}(\zeta, w) \coloneqq P\left(\zeta, c^{-1}\left(w - \frac{c-1}{\zeta}\right)\right).
\end{equation*}
For $F_2 = 0$, differentiation with respect to $\tau$ yields
\begin{equation*}
	\dot{\zeta} + \tau^{-2}(z + w^{-1}) + (\tau^{-1} - 1)w^{-2} \dot{w} = 0.
\end{equation*}
This reveals the $2 \times 2$ tangent system
\begin{equation}
\label{eq:tangent-system}
    \begin{bmatrix}
        \partial_{\zeta} \underline{P}(\zeta, w) & \partial_w \underline{P}(\zeta, w) \\
        1 & (\tau^{-1} - 1)w^{-2}
    \end{bmatrix}
    \begin{bmatrix}
        \dot{\zeta} \\
        \dot{w}
    \end{bmatrix}
    =
    \begin{bmatrix}
        0 \\
        -\tau^{-2}(z + w^{-1})
    \end{bmatrix}.
\end{equation}
The system \eqref{eq:tangent-system} provides the predictor direction for advancing the physical point from one decompression ratio to the next. The predicted point is then corrected by a Newton projection onto the nonlinear system $(F_1,F_2) = (0,0)$. As in \citet{AMELI-2026}, steps in $\tau$ are adaptively subdivided when the local geometry becomes stiff to prevent jumping between nearby sheets. After the correction step converges, the value of the decompressed ordinary Stieltjes transform is recovered from the companion transform via the relation
\begin{equation*}
	m_\tau(z) = (\tau c)^{-1}\left(w - \frac{\tau c-1}{z}\right).
\end{equation*}
The complete predictor-corrector procedure is summarized in \Cref{alg:deform}, and allows us to coherently evolve the density while remaining strictly on the physical branch.

\begin{algorithm}[ht!]
    \caption{Deformed decompression by geometric continuation on a spectral curve}
    \label{alg:deform}
    \SetKwInOut{Input}{Input}
    \SetKwInOut{Output}{Output}
    \DontPrintSemicolon

    \vspace{1mm}
    \Input{
        Polynomial \(P \in \mathbb{R}[z,m]\) such that \(P(z,m_1(z))=0\) (\citet[Alg. D.1]{AMELI-2026}); \\
        Initial aspect ratio \(c > 0\); \\
        Query grid \(\{z_j\}_{j=1}^{N_z}\subset\mathbb{C}^{+}\); \\
        Decompression ratios \(1=\tau_0<\tau_1<\cdots<\tau_T\); \\
        Oracle for the initial \(m_1(z)\) on \(P(z, m) = 0\) (\citet[Alg. E.1]{AMELI-2026}); \\
        Newton tolerance \(\epsilon>0\).
    }

    \vspace{2mm}
    \Output{
        Physical Stieltjes values \(m_{\tau_{\ell}}(z_j)\) for all requested \(\tau_{\ell}\) and \(z_j\).
    }

    \vspace{2mm}
    \tcp{Initialize the physical sheet at \(\tau_0=1\)}
    \For{\(j = 1,\dots,N_z\)}{
        \(m_j^{(0)} \leftarrow m_1(z_j)\) \tcp*{Select physical root using \citet[Appendix E]{AMELI-2026}}
        \(w_j^{(0)} \leftarrow c m_j^{(0)} + (c - 1)z_j^{-1}\) \tcp*{Compute initial companion transform \(\underline{m}_1\)}
        \(\zeta_j^{(0)} \leftarrow z_j\) \tcp*{At \(\tau=1\), the map is identity in \(z\)}
        \(m_{\tau_0}(z_j) \leftarrow m_j^{(0)}\) \;
    }

    \vspace{2mm}
    \tcp{Continue the physical sheet in decompression ratio \(\tau\)}
    \For{\(\ell = 1,\dots,T\)}{
        \For{\(j = 1,\dots,N_z\)}{
            Set \((\zeta,w) \leftarrow (\zeta_j^{(\ell-1)}, w_j^{(\ell-1)})\). \;
            Define \(\underline{P}(\zeta, w) \coloneqq P\left(\zeta, c^{-1}\left(w - \frac{c - 1}{\zeta}\right)\right)\). \;

            \vspace{2mm}
            \tcp{Predict along the tangent of the constrained system}
            Solve
            \begin{equation*}
                \begin{bmatrix}
                    \partial_{\zeta}\underline{P}(\zeta,w) & \partial_w \underline{P}(\zeta,w) \\
                    1 & (\tau_{\ell-1}^{-1}-1)w^{-2}
                \end{bmatrix}
                \begin{bmatrix}
                    \dot{\zeta}\\
                    \dot{w}
                \end{bmatrix}
                =
                \begin{bmatrix}
                    0\\
                    -\tau_{\ell-1}^{-2}(z_j + w^{-1})
                \end{bmatrix}.
            \end{equation*}

            \(\Delta \tau \leftarrow \tau_{\ell}-\tau_{\ell-1}\). \;
            \(\zeta_{\mathrm{pred}} \leftarrow \zeta + \Delta\tau\,\dot{\zeta}\). \;
            \(w_{\mathrm{pred}} \leftarrow w + \Delta\tau\,\dot{w}\). \;

            \vspace{2mm}
            \tcp{Correct by projecting back to the decompression system}
            Starting from \((\zeta_{\mathrm{pred}},w_{\mathrm{pred}})\), apply damped Newton to solve
            \begin{equation*}
                \underline{P}(\zeta,w)=0,
                \qquad
                \zeta - \tau_{\ell}^{-1}z_j - (\tau_{\ell}^{-1}-1)w^{-1} = 0.
            \end{equation*}

            \vspace{1mm}
            \eIf{Newton converges to \((\zeta_{\ast},w_{\ast})\) within tolerance \(\epsilon\)}{
                \(\zeta_j^{(\ell)} \leftarrow \zeta_{\ast}\), \quad
                \(w_j^{(\ell)} \leftarrow w_{\ast}\). \;
                \(c_{\ell} \leftarrow \tau_{\ell} c\). \;
                \(m_{\tau_{\ell}}(z_j) \leftarrow c_{\ell}^{-1}\left(w_{\ast} - \frac{c_{\ell}-1}{z_j}\right)\). \tcp*{Recover evolved ordinary Stieltjes value}
            }{
                Subdivide \([\tau_{\ell-1},\tau_{\ell}]\) and repeat the predictor--corrector step. \tcp*{Avoid sheet jumping in stiff regions}
            }
        }
    }

    \vspace{2mm}
    \Return{\(\{m_{\tau_{\ell}}(z_j)\}_{\ell,j}\).}\;
\end{algorithm}


\section{Stability Analysis} \label{app:stability}

To analyze the stability of \(\underline{m}_\tau\) it is convenient to reparameterize in terms of \(t = \log(\tau)\). Noting that $\frac{\mathrm{d}\tau}{\mathrm{d}t} = \tau$, our PDE \eqref{eqn:dd-flow} becomes
\begin{align}\label{eqn:ddt}
    \um_t\frac{\partial \um_t}{\partial t} + (z \um_t + 1)\frac{\partial \um_t}{\partial z} = 0.
\end{align}
We then consider the PDE operator's effect on the perturbed field \(\um_t + \delta \eta_t\), and derive a PDE governing the evolution of the perturbation direction \(\eta_t\).

We are then able to quantify the impact of this perturbation under the horizontal slice-wise \(L^2\)-norm in the upper half complex plane, defined by
\begin{equation*}
    \Vert \eta_t( \cdot + iy) \Vert^2_{L^2(\Gamma_y)} \coloneqq \int_{\mathbb{R}} \left| \eta_t(x + iy) \right|^2 \, \mathrm{d}x,
\end{equation*}
where \(\Gamma_y \coloneqq \{ z = x + iy \;|\; x \in \mathbb{R} \}\) is slice at height \(y > 0\). This is a natural choice of norm for our analysis, since recovery of the spectral distribution requires evaluation over slices slightly above the real axis.

We establish a Gr\"{o}nwall-type inequality controlling the growth of the perturbation norm from its initial value. This is made precise in the following proposition.

\begin{proposition} \label{prop:stability}
    Suppose \(\rho(t, \cdot) \in L^2(\mathbb{R})\), and let \(m_t(z)\) be its companion Stieltjes transform defined on \(\mathbb{C}^{+}\). Let \(\eta_t\) be a G\^{a}teaux perturbation direction of \(\um_t\), with perturbation \(\delta \eta_t\) for \(0 < \delta \ll 1\), and suppose \(\eta_t\) satisfies the Bernstein inequality
    \begin{equation}
        \Vert \partial_x \eta_t(z)\Vert_{L^2(\Gamma_y)} \leq M \Vert \eta_t(z)\Vert_{L^2(\Gamma_y)}, \quad z \in \Gamma_y, \forall y > 0 \label{eq:bernstein}
    \end{equation}
    meaning that \(\eta_t(\cdot+iy)\) lies in a \(M\)-band‑limited Paley--Wiener space on each slice, ensuring appropriate perturbation decay. Then, for each \(y > 0\), there holds
    \begin{equation}
        \frac{\mathrm{d}}{\mathrm{d} t} \Vert \eta_t \Vert_{L^2(\Gamma_y)}^2 \leq (1 + C_y(t)) \; \Vert \eta_t \Vert_{L^2(\Gamma_y)}^2, \label{eq:gronwall}
    \end{equation}
    where
    \begin{equation}
        C_y(t) \coloneqq \sup_{z \in \Gamma_y}  \Re \, \frac{\partial_z \um_t(z)}{\um_t(z)^2}  + 2 M \sup_{z \in \Gamma_y} \left| \Im \, \left(z + \frac{1}{\um_t(z)}\right) \right|.
        \label{eq:esssup-C}
    \end{equation}
\end{proposition}

\begin{proof}
    Write \(\underline{m}'_t = \um_t + \delta \eta_t\), where \( \eta_t \in H^2(\mathbb{C}^{+})\) is the usual Hardy space, for a small perturbation with the boundary condition \(\eta_t(z) = 0\) at \(\vert z \vert \to \infty\). Plugging \(\underline{m}'_t\) into the PDE \eqref{eqn:ddt} and discarding the $\mathcal{O}(\delta^2)$ terms as $\delta\to0$ gives the linear perturbation equation
    \begin{equation*}
        \frac{\partial \eta_t}{\partial t} = - \left(z + \frac{1}{\um_t}\right)\frac{\partial \eta_t}{\partial z} + \frac{1}{\um_t^2}\frac{\partial \um_t}{\partial z}\eta_t.
    \end{equation*}
    This is a linear transport PDE, with transport velocity $\big(z + \frac{1}{\um_t}\big)$ and growth coefficient $\frac{1}{\um_t^2} \frac{\partial \um_t}{\partial z}$.

    Since $\eta_t$ is holomorphic, $\frac{\partial \eta_t}{\partial z} = \frac{\partial \eta_t}{\partial x}$ along $\Gamma_y$, so on these slices we have
    \begin{equation*}
        \frac{\partial \eta_t}{\partial t} = - \left(z + \frac{1}{\um_t}\right)\frac{\partial \eta_t}{\partial x} + \frac{1}{\um_t^2}\frac{\partial \um_t}{\partial z}\eta_t.
    \end{equation*}

    To obtain an energy estimate, we multiply by the complex conjugate $\overline{\eta_t}$, integrate over $\Gamma_y$ and take the real part:
    \begin{equation*}
        \frac{1}{2}\frac{\mathrm{d}}{\mathrm{d}t}\Vert\eta_t\Vert^2_{L^2(\Gamma_y)} = \Re\int_{\Gamma_y}- \left(z + \frac{1}{\um_t}\right)\frac{\partial \eta_t}{\partial x}\overline{\eta_t}\mathrm{d}x + \int_{\Gamma_y}\Re\left(\frac{1}{\um_t^2}\frac{\partial \um_t}{\partial z}\right)\vert\eta_t\vert^2\mathrm{d}x.
    \end{equation*}

    Decomposing the velocity into its real and imaginary components as $z + \frac{1}{\um_t} = A + i B$, we observe
        \begin{equation*}
        \Re\left(- \left(z + \frac{1}{\um_t}\right)\frac{\partial \eta_t}{\partial x}\overline{\eta_t}\right)
         = -\frac{1}{2}A\frac{\partial |\eta_t|^2}{\partial x} + B\Im\left(\frac{\partial \eta_t}{\partial x}\overline{\eta_t}\right)
    \end{equation*}
    Integrating this by parts, keeping in mind that the boundary terms vanish, we find
    \begin{equation*}
        \Re\int_{\Gamma_y}- \left(z + \frac{1}{\um_t}\right)\frac{\partial \eta_t}{\partial x}\overline{\eta_t}\mathrm{d}x = \frac{1}{2}\int_{\Gamma_y}\frac{\partial A}{\partial x}|\eta_t|^2\mathrm{d}x +  \int_{\Gamma_y}B\Im\left(\frac{\partial \eta_t}{\partial x}\overline{\eta_t}\right)\mathrm{d}x.
    \end{equation*}
    The Bernstein inequality \eqref{eq:bernstein} estimates the second term as
    \begin{equation*}
        \left\vert\int_{\Gamma_y}B\Im\left(\frac{\partial \eta_t}{\partial x}\overline{\eta_t}\right)\mathrm{d}x\right\vert \leq M \sup_{\Gamma_y} |B| \Vert \eta_t \Vert^2_{L^2(\Gamma_y)}.
    \end{equation*}
    Noting that $\frac{\partial A}{\partial x} = 1 - \Re \big(\frac{1}{\um_t^2}\frac{\partial \um_t}{\partial z}\big)$, we find that
    \begin{equation*}
        \frac{1}{2}\frac{\partial A}{\partial x} + \Re\left(\frac{1}{\um_t^2}\frac{\partial \um_t}{\partial z}\right) = \frac{1}{2} + \frac{1}{2}\Re\left(\frac{1}{\um_t^2}\frac{\partial \um_t}{\partial z}\right).
    \end{equation*}

    Combining these terms we obtain the required result
    \begin{equation*}
        \frac{\mathrm{d}}{\mathrm{d} t} \Vert \eta_t \Vert_{L^2(\Gamma_y)}^2 \leq
        \left(1 + \sup_{z \in \Gamma_y} \Re \, \frac{\partial_z \um_t(z)}{\um_t(z)^2}  + 2 M \sup_{z \in \Gamma_y} \left| \Im \, \left(z + \frac{1}{\um_t(z)}\right) \right|\right) \; \Vert \eta_t \Vert_{L^2(\Gamma_y)}^2.
    \end{equation*}
    Using \eqref{eq:esssup-C} and applying Gr\"onwall's inequality we can further estimate
    \begin{equation*}
        \Vert \eta_t \Vert_{L^2(\Gamma_y)}^2 \leq \Vert \eta_0 \Vert_{L^2(\Gamma_y)}^2 \exp \left(\int_0^t (1 + C_y(s)) \, \mathrm{d}s\right).
\end{equation*}
\end{proof}

\begin{remark} \label{rem:stability}
    For compactly supported companion spectral measures, \(C_y(t)\) is finite for every fixed \(y > 0\). Indeed, \(\Im \underline{m}_t(z) > 0\) on \(\mathbb{C}^+\), and the Stieltjes expansion \(\underline{m}_t(z) = -z^{-1} + O(z^{-2})\) implies \(\partial_z \underline{m}_t(z) / \underline{m}_t(z)^2 \to 1\) and \(z + 1/\underline{m}_t(z) = O(1)\) as \(\vert z \vert \to \infty\). Thus, both suprema in \eqref{eq:esssup-C} are finite on \(\Gamma_y\). If the companion supports remain uniformly bounded over a finite decompression interval, so does \(C_y(t)\). The bound need not remain finite as \(y \to 0^{+}\).
\end{remark}


\section{Evolution of Spectral Quantities}
\label{sec:evo-features}

Deformed decompression acts on spectral densities by evolving their Stieltjes transforms in a controlled way. While this gives access to the entire spectral distribution of a larger matrix, it is often the case that we only care about certain spectral quantities, such as condition numbers, spectral moments, atomic locations and weights, or cusp points where the bulk densities either merge or split. The spectral-curve viewpoint and its corresponding change-of-variable formulae allow for convenient calculation of these quantities.


\subsection{Evolving Spectral Edges} 
\label{sec:edge}

Spectral edges are important quantities when modeling the spectrum of large matrices, with the most obvious applications being the estimation of the condition number or detecting when a family of matrices is becoming singular. More recently, the location of internal edges in multi-bulk distributions have been linked to training dynamics of diffusion models \citep{BONNAIRE-2025}. Writing the support of a distribution with $k$ separate bulk components as 
\begin{equation*}
    I(\tau) \coloneqq \bigcup_{j=1}^{k} I_j(\tau), \qquad I_j(\tau) \coloneqq[a_j(\tau), b_j(\tau)],
\end{equation*}
then the edges are simply the endpoints of the support subintervals \(\{a_j(\tau), b_j(\tau)\}_{j=1}^{k}\).

Under \Cref{ass:polynomial} these edges correspond to branch points of the spectral curve
\begin{equation*}
    \mathcal{C} \coloneqq \left\{ (\zeta, y) \in \mathbb{C}^2 \mid P(\zeta, y) = 0 \right\},
\end{equation*}
corresponding to points where multiple sheets (given by roots $y = y_i(\zeta)$) meet. These points have the convenient characterization as satisfying
\begin{equation*}
    P(\zeta_{\ast}, y_{\ast}) = 0,
    \qquad \text{and} \qquad
    \partial_y P(\zeta_{\ast}, y_{\ast}) = 0,
\end{equation*}
describing how sheets glue across branch cuts. The branch points of interest to us are those that connect the physical sheet to non-physical sheets, since branch points connecting non-physical sheets do not correspond to the distribution of interest's support. This means that care must be taken to select the correct  branch points.  

With this in mind, we are able to track branch points under deformed decompression. This is formalized in the following proposition.

\begin{proposition}[Evolving Branch Points under Deformed Decompression]
    \label{prop:dd-branch-point}
    Let \(m\) be a Stieltjes transform satisfying \(\Cref{ass:polynomial}\), let
    \(\tau\geq 1\), and let \(m_\tau\) be the deformed decompression of \(m\) by
    the factor \(\tau\). Let
    \begin{equation*}
        \mathcal C \coloneqq \left\{(\zeta,y)\in\mathbb C^2 \mid P(\zeta,y)=0\right\}.
    \end{equation*}
    Then, away from the singular chart loci, the branch-point candidates of the
    deformed decompressed curve are obtained from solutions
    \((\zeta_\ast,y_\ast)\in\mathcal C\) satisfying
    \begin{subequations} \label{eq:dd-branch-point}
    \begin{align}
        &P(\zeta_\ast,y_\ast)=0, \\
        & \tau \left(c\zeta_\ast y_\ast+c-1\right)^2 \partial_y P(\zeta_\ast,y_\ast) +(\tau-1)\left(c\zeta_\ast^2\,\partial_\zeta P(\zeta_\ast,y_\ast) +(c-1)\partial_y P(\zeta_\ast,y_\ast)\right) =0.
    \end{align}
    \end{subequations}
    The corresponding point in the deformed decompressed \(z\)-plane is
    \begin{equation*}    
        z_\ast = \tau\zeta_\ast + \frac{(\tau-1)\zeta_\ast}{c\zeta_\ast y_\ast+c-1}.
    \end{equation*}
    In particular, real solutions \(z_\ast\in\mathbb R\) of \eqref{eq:dd-branch-point} include the evolving spectral edges of the real density under deformed decompression.
\end{proposition}

\begin{proof}
    Set $ u\coloneqq \underline m_\tau,$ and $y\coloneqq c^{-1} \big(u-\frac{c-1}{\zeta} \big)$. Then $u=c y+\frac{c-1}{\zeta} = \frac{c\zeta y+c-1}{\zeta}$. 
    The deformed decompression relation $P\big(\zeta,c^{-1}\big(\underline m_\tau-\frac{c-1}{\zeta}\big)\big) = 0$
    therefore pulls back to the original curve $P(\zeta,y)=0.$
    Moreover, from $\zeta = \tau^{-1} \big(z+u^{-1} \big) - u^{-1}$, we obtain
    \begin{equation*}    
        z = \tau\zeta+(\tau-1)u^{-1} = \tau\zeta + \frac{(\tau-1)\zeta}{c\zeta y+c-1}.
    \end{equation*}
    Hence the branch points of the deformed decompressed curve are the critical
    points of the projection
    $$(\zeta,y) \mapsto z(\zeta,y) = \tau\zeta + \frac{(\tau-1)\zeta}{c\zeta y+c-1}$$
    restricted to \(\mathcal C\). Along \(\mathcal C\), implicit differentiation gives
    \begin{align}\label{eqn:edge-grads}
        \partial_\zeta P(\zeta,y)
        + \partial_y P(\zeta,y)\frac{\mathrm dy}{\mathrm d\zeta} = 0,
        \qquad\text{so, whenever \(\partial_yP\neq 0\)},\qquad
        \frac{\mathrm dy}{\mathrm d\zeta} = - \frac{\partial_\zeta P}{\partial_y P}.
    \end{align}
    Writing $D\coloneqq c\zeta y+c-1$, we have $z(\zeta,y) = \tau\zeta+(\tau-1)\frac{\zeta}{D}.$ Differentiating along \(\mathcal C\),
    \begin{equation*}    
        \frac{\mathrm dz}{\mathrm d\zeta} = \tau + (\tau-1) \frac{D-\zeta D'}{D^2}.
    \end{equation*}
    Since $D'= c y+c\zeta\frac{\mathrm dy}{\mathrm d\zeta}$, we obtain
    \[
        D-\zeta D' = c\zeta y+c-1 - \zeta \left( c y+c\zeta\frac{\mathrm dy}{\mathrm d\zeta} \right)
        = c-1-c\zeta^2\frac{\mathrm dy}{\mathrm d\zeta}.
    \]
    Therefore
    \[
        \frac{\mathrm dz}{\mathrm d\zeta} = \tau + (\tau-1) \frac{c-1-c\zeta^2\frac{\mathrm dy}{\mathrm d\zeta}}{D^2}.
    \]
    Substituting $\eqref{eqn:edge-grads}$ gives
    \begin{equation*}
        \frac{\mathrm dz}{\mathrm d\zeta} = \tau + (\tau-1) \frac{ c-1 + c\zeta^2 \frac{\partial_\zeta P}{\partial_y P}}{D^2}.
    \end{equation*}
    Thus \(\mathrm dz/\mathrm d\zeta=0\) is equivalent to
    \[
        \tau D^2\partial_y P + (\tau-1) \left( c\zeta^2\partial_\zeta P + (c-1)\partial_y P  \right) = 0.
    \]
    Since \(D=c\zeta y+c-1\), the critical-point equations are precisely
    \eqref{eq:dd-branch-point}. The pushforward formula for \(z_\ast\) follows from
    the expression for \(z(\zeta,y)\).
    
    Finally, we note that the algebraic system gives branch-point candidates. The spectral edges are the real-valued candidates whose cuts connect the physical sheet to a non-physical sheet.
\end{proof}

While \Cref{prop:dd-branch-point} holds for all branch points, we are concerned only with the selection of a particular subset, the spectral edges \(\{a_j(\tau), b_j(\tau)\}_{j=1}^k\). For fixed \(\tau \geq 1\), the system \eqref{eq:dd-branch-point} admits solutions
\begin{equation*}
    \mathcal{B}(\tau) \coloneqq \left\{ z_{\ast}^{(j)}(\tau) \right\}_{j=1}^p,
\end{equation*}
where typically \(p \geq k\). The real-valued subset \(\mathcal{B}(\tau) \cap \mathbb{R}\) provides edge candidates, which can be selected as the ones connecting the physical to a non-physical sheet. This is realized as a nontrivial jump across the real axis at that point in the Stieltjes inversion formula. \Cref{alg:edge} summarizes the resulting procedure, and is a straight-forward adaptation of \citep[Algorithm~G.1]{AMELI-2026} to the DD coordinate change.

\begin{algorithm}[ht!]
    \caption{Evolution of spectral edges under algebraic deformed decompression}
    \label{alg:edge}
    \SetKwInOut{Input}{Input}
    \SetKwInOut{Output}{Output}
    \DontPrintSemicolon

    \vspace{1mm}
    \Input{
        Polynomial \(P \in \mathbb{R}[z, m]\) with \(\deg_m(P) = s\) such that \(P(z, m_1(z)) = 0\) (see \citet[Algorithm~E.1]{AMELI-2026}); \\
        Initial aspect ratio \(c > 0\); \\
        Decompression ratio grid \(1 = \tau_0 < \tau_1 < \cdots < \tau_T\); \\
        Initial support \(I(1) = \bigcup_{j=1}^{k_0} [a_j(1), b_j(1)]\) (optional, for initialization); \\
        Oracle for the physical branch \(m_1(z)\) on \(\mathbb{C} \setminus I\) (see \citep[Algorithm E.1]{AMELI-2026}); \hfill {\textit{\textcolor{gray}{\footnotesize{// Only needed if \(I(1)\) is provided.}}}} \\
        Tolerance \(\epsilon > 0\).
    }

    \vspace{2mm}
    \Output{
        Edge trajectories \(\{a_j(\tau_{\ell}), b_{j}(\tau_{\ell})\}\) stored as \(T+1\) by \(2 k_0\) with \texttt{NaN}s after merging; \\
        Active bulk count \(k(\tau_{\ell})\) (number of connected components).
    }

    \vspace{2mm}
    \tcp{Define the edge system from \Cref{prop:dd-branch-point}}
    Define the functions
    \begin{align*}
        G_1(\zeta, y; \tau) &\coloneqq P(\zeta, y) \\
        G_2(\zeta, y; \tau) &\coloneqq \tau \left(c\zeta y +c-1\right)^2 \partial_y P(\zeta,y) +(\tau-1)\left(c\zeta^2\,\partial_\zeta P(\zeta,y) +(c-1)\partial_y P(\zeta,y)\right).
    \end{align*}

    \tcp{Initialization at \(\tau_0 = 1\)}
    \If{initial edges \(I(1)\) are provided}{
        Initialize guesses \(\{(\zeta_0^{(j)}, y_0^{(j)})\}_{j=1}^{2 k_0}\) as follows: \tcp*{Any reasonable edge-local seeding works}
        \(\quad\triangleright\) Sample \(2 k_0\) points \(\zeta\) near each endpoint \;
        \(\quad\triangleright\) Choose Herglotz-consistent root \(y_0^{(j)} \leftarrow m_1(\zeta_0^{(j)})\). \;
    }
    \Else{
        Obtain initial guesses \(\{(\zeta_0^{(j)}, y_0^{(j)})\}_{j=1}^{2 k_0}\) by solving \((G_1, G_2) = (0, 0)\) at \(\tau_0\) using multiple seeds.
    }

    \vspace{2mm}
    \tcp{Refine the initial guesses}
    Solve \((G_1, G_2) = (0, 0)\) at \(\tau_0\), store the initial edge locations \(x_0^{(j)} = \zeta_0^{(j)}\). Set \(k(\tau_0) \leftarrow k_0\). \;

    \vspace{2mm}
    \tcp{Track edge preimages \((\zeta, y)\) by warm-started Newton over time}
    \For{\(\ell = 1, \dots, T\)}{
        \(\tau \leftarrow \tau_{\ell}\) \;
        \ForEach{\(j = 1, \cdots, 2 k_0\) not yet discarded}{
            Use previous solution \((\zeta_{\ell-1}^{(j)}, y_{\ell-1}^{(j)})\) as initial guess. \;
            Apply damped Newton to solve \((G_1(\zeta, y; \tau_{\ell}), G_2(\zeta, y; \tau_{\ell})) = (0, 0)\) for \((\zeta_{\ell}^{(j)}, y_{\ell}^{(j)})\). \;
            \(z_{\ell}^{(j)}  \leftarrow \tau\zeta_{\ell}^{(j)} + \frac{(\tau-1)\zeta_{\ell}^{(j)}}{c\zeta_{\ell}^{(j)} y_{\ell}^{(j)}+c-1}\) \tcp*{Pushforward to the \(z\)-plane}
            \leIf{\(\vert \Im z_{\ell}^{(j)} \vert < \epsilon\)}{
                Store \(x_{\ell}^{(j)} \leftarrow \Re z_{\ell}^{(j)}\)
            }{
                mark failure.
            }
        }

        \vspace{2mm}
        \tcp{Handle merged edges of disjoint bulks (inner edges annihilate)}
        \(\mathcal{P} \leftarrow \{a_1, b_1, \dots, a_{k_0}, b_{k_0} \}\) \;
        \ForEach{adjacent pair \((b_j, a_{j+1})\) in \(\mathcal{P}\)}{
            \lIf{\(b_j(\tau_{\ell}) > a_{j+1}(\tau_{\ell})\)}{
                \((b_{j}(\tau_{\ell}), a_{j+1}(\tau_{\ell})) \leftarrow (\text{\texttt{NaN}}, \text{ \texttt{NaN}})\)
            }
        }
        \(k(\tau_{\ell}) \leftarrow\) number of remaining connected components. 
    }
    
    \vspace{2mm}
    \Return{Spectral edges \(\left\{ \{a_{j}(\tau_{\ell}), b_{j}(\tau_{\ell})\}_{j=1}^{k_0} \right\}_{\ell=0}^{T}\) and active bulk count \(\{k(\tau_{\ell})\}_{\ell=0}^{T}\)};
\end{algorithm}


\subsection{Spectral Cusps and The Mechanism of Merging Bulks}\label{sec:merge}

Write \(I_j(\tau) \coloneqq [a_j(\tau), b_j(\tau)]\) for the \(j\)-th bulk region and assume that at \(\tau=1\) the bulks are disjoint,
\begin{equation*}
    a_1(1) < b_1(1) < a_2(1) < b_2(1) < \cdots < a_k(1) < b_k(1),
\end{equation*}
so that \(I_j(1) \cap I_{j+1}(1) = \emptyset\). After decompression to \(\tau_{\ast}\), merging or splitting of bulks can be characterized as the density developing a \emph{cusp} (corresponding to a cubic singularity at a sharp local minimum), associated with the Pearcey universality class \citep{ERDOS-2020}. These colliding (or emerging) branch points correspond to a higher-order critical point of the projection of the spectral curve. 
Similar to the tracking of edges, under \Cref{ass:polynomial} there is a set of algebraic conditions that characterize these cusp points. 
The following proposition makes this precise.

\begin{proposition}[Cusp Point at Merging Bulks under Deformed Decompression]
    \label{prop:dd-cusp}
    Suppose two consecutive real edges collide at \(\tau = \tau_\ast\), so that $x_\ast = b_j(\tau_\ast)=a_{j+1}(\tau_\ast)$.  Let \((\zeta_\ast,y_\ast)\in\mathcal C\), where $\mathcal C=\{(\zeta,y)\in\mathbb C^2:P(\zeta,y )=0\}$, be the pullback of \(x_\ast\) under the deformed-decompression projection 
    \begin{equation*}
        z(\zeta,y) = \tau_\ast\zeta + \frac{(\tau_\ast-1)\zeta}{c\zeta y+c-1}.
    \end{equation*}
    Then \((\zeta_\ast,y_\ast)\) is a higher-order critical point of this projection restricted to \(\mathcal C\). Hence, away from the singular chart loci, it satisfies
    \begin{subequations}
    \label{eq:dd-cusp}
    \begin{align}
        &P(\zeta_\ast,y_\ast)=0, \\
        &\tau_\ast (c\zeta_\ast y_\ast+c-1)^2 \partial_yP + (\tau_\ast-1) \left( c\zeta_\ast^2\partial_\zeta P + (c-1)\partial_yP \right) =0, \\
        &(c\zeta_\ast y_\ast+c-1) \left[ 2\zeta_\ast(\partial_\zeta P)(\partial_yP)^2 + \zeta_\ast^2 \left( \partial_{\zeta\zeta}P(\partial_yP)^2 - 2\partial_{\zeta y}P\,\partial_\zeta P\,\partial_yP + \partial_{yy}P(\partial_\zeta P)^2 \right)
        \right] \notag \\
        &\qquad -2\partial_yP \left( (c-1)\partial_yP + c\zeta_\ast^2\partial_\zeta P \right) \left(y_\ast\partial_yP-\zeta_\ast\partial_\zeta P \right) =0.
    \end{align}
    \end{subequations}
    Here all derivatives are evaluated at \((\zeta_\ast,y_\ast)\). Moreover,
    \[
        x_\ast = \tau_\ast\zeta_\ast + \frac{(\tau_\ast-1)\zeta_\ast}{c\zeta_\ast y_\ast+c-1}.
    \]
\end{proposition}

\begin{proof}
    Let $D(\zeta,y)=c\zeta y+c-1$. For deformed decompression, the projection to the \(z\)-plane is $ z(\zeta,y) = \tau\zeta+\frac{(\tau-1)\zeta}{D(\zeta,y)}$. 
    A cusp is a critical point of multiplicity two of this map restricted to
    \(\mathcal C\). Thus, along a local branch \(y=y(\zeta)\), we impose
    \[
        z'(\zeta_\ast)=0,
        \qquad
        z''(\zeta_\ast)=0,
        \qquad
        P(\zeta_\ast,y_\ast)=0.
    \]
    
    Implicit differentiation of \(P(\zeta,y(\zeta))=0\) gives
    \[
        y'=-\frac{\partial_\zeta P}{\partial_yP}, 
        \qquad\text{and}\qquad
        y'' = - \frac{ \partial_{\zeta\zeta}P(\partial_yP)^2 - 2\partial_{\zeta y}P\,\partial_\zeta P\,\partial_yP + \partial_{yy}P(\partial_\zeta P)^2 }{(\partial_yP)^3}.
    \]
    
    Now $ z' = \tau + (\tau-1) \frac{D-\zeta D'}{D^2}$.
    Since $ D-\zeta D'=c-1-c\zeta^2y'$, the condition \(z'=0\), after substituting the formula for \(y'\), becomes
    \[
        \tau D^2\partial_yP + (\tau-1) \left( c\zeta^2\partial_\zeta P + (c-1)\partial_yP \right) = 0.
    \]
    
    Next set $N=c-1-c\zeta^2y'$. Then $ z' = \tau+(\tau-1)\frac{N}{D^2}$, so $z'' = (\tau-1)\frac{N'D-2ND'}{D^3}$. Thus \(z''=0\) is equivalent to $N'D-2ND'=0.$ Substituting
    \[
        y'=-\frac{\partial_\zeta P}{\partial_yP}, \qquad
        y'' = - \frac{ \partial_{\zeta\zeta}P(\partial_yP)^2 - 2\partial_{\zeta y}P\,\partial_\zeta P\,\partial_yP + \partial_{yy}P(\partial_\zeta P)^2 }{(\partial_yP)^3},
    \]
    and clearing denominators yields the third equation in \eqref{eq:dd-cusp}. Finally, evaluating the projection gives
    \[
        x_\ast = \tau_\ast\zeta_\ast + \frac{(\tau_\ast-1) \zeta_\ast}{c\zeta_\ast y_\ast + c - 1}.
    \]
\end{proof}

Equation \eqref{eq:dd-cusp} allows us to solve for the triple \((\zeta_{\ast}, y_{\ast}, \tau_{\ast})\), recovering both the merging (or splitting) ratio \(\tau_{\ast}\) and the location \(x_{\ast}\). This is more accurate than using a discrete time grid to monitor edge crossing. After the identification of the cusp time $\tau_\ast$, the decompressed distribution can be treated as having one fewer (or more) bulk regions.


\subsection{Dynamics of Atoms} \label{sec:atoms}
There is also a convenient characterization of atomic components of a distribution, as simple poles of the Stieltjes transform. Around such a pole at $z = x_\circ$, the weights can be computed in terms of the residue 
$w = - \Res_{z = x_{\circ}} m(z) = \lim_{z \to x_{\circ}} (x_{\circ} - z) m(z).$

Writing the polynomial defining the spectral curve \(P(z, m) = 0\) as \(P(z, m) \coloneqq \sum_{j=0}^{s} a_j(z) m^j\), a pole at a finite real point \(x_{\circ}\)  necessarily satisfies \(x_{\circ}\) is a root of the leading coefficient, \ie \(a_s(x_{\circ}) = 0\). 
By expanding the Stieltjes transform around such a point, we are able to track changes in both the location and weights under deformed decompression in the following way.

\begin{proposition}[Dynamics of Atoms under Deformed Decompression]
    \label{prop:dd-atom}
    Let \(x_\circ\in\mathbb R\) be an atom of the spectral measure \(\nu\) with initial mass
    \(w_\circ\in(0, 1)\). Under deformed decompression with parameter \(\tau\), the following hold. If \(x_\circ \neq 0\), then the atom location is pushed forward to
    \begin{equation*}
        x_\circ(\tau)=\tau x_\circ,
    \end{equation*}
    and mass evolves as
    \begin{equation*}
        w(\tau) = w_\circ - \frac{\tau - 1}{c \tau}.
    \end{equation*}
    If \(x_\circ = 0\), then $x_\circ(\tau) = 0$, and its mass evolves as $w(\tau) = \max \left\{w_\circ, 1 - \frac{1}{c \tau} \right\}$. For \(x_\circ \neq 0\), the formula describes the atom while $w(\tau) > 0$. When the predicted mass reaches zero, the pole disappears from the physical branch.
\end{proposition}

\begin{remark}
    For an atom at zero, the weight remains at $w_\circ$ until the rank-deficiency threshold $1 - (c \tau)^{-1}$ exceeds $w_\circ$, after which it follows this threshold. For $x_\circ$ away from zero, the atom location is pushed away from the origin, while its weight decreases until extinction, if it occurs.
\end{remark}

\begin{proof}
    We use the deformed-decompression change of variables $u \coloneqq \underline m_1(\zeta) = \underline m_\tau(z)$, and $y=c^{-1}\big(u-\frac{c-1}{\zeta}\big)$, so that $ u=c y+\frac{c-1}{\zeta}$, and $z=\tau\zeta+(\tau-1)u^{-1}$. First suppose \(x_\circ\neq 0\). Near the initial atom,
    $y(\zeta) = -\frac{w_\circ}{\zeta-x_\circ} + \mathcal{O}(1)$. Since \(x_\circ\neq 0\), the term \((c-1)/\zeta\) is analytic at \(x_\circ\). Hence
    \begin{equation*}
        u(\zeta) = c y(\zeta)+\frac{c-1}{\zeta} = -\frac{c w_\circ}{\zeta-x_\circ}+\mathcal{O}(1),
    \end{equation*}
    and also
    \begin{equation*}
        u(\zeta)^{-1} = -\frac{\zeta-x_\circ}{c w_\circ} + \mathcal{O}\!\left((\zeta-x_\circ)^2\right).
    \end{equation*}
    Substituting into the pushforward map gives
    \begin{equation*}
        z = \tau\zeta + (\tau - 1)u^{-1} = \tau x_\circ + \left(\tau - \frac{\tau - 1}{c w_\circ}\right)(\zeta - x_\circ) + \mathcal{O}\!\left((\zeta - x_\circ)^2\right).
    \end{equation*}
    Thus the atom is pushed forward to $x_\circ(\tau)=\tau x_\circ$. Moreover, if \(\alpha \coloneqq \tau-\frac{\tau-1}{c w_\circ}\),
    then
    \begin{equation*}
        z - \tau x_\circ = \alpha (\zeta - x_\circ) + \mathcal{O}\!\left((\zeta - x_\circ)^2\right).
    \end{equation*}
    Consequently, $u(z) = -\frac{c w_\circ\alpha}{z-\tau x_\circ}+O(1)$. 
    On the decompressed side,
    \begin{equation*}
        u(z) = \underline m_\tau(z) = \tau c \, m_\tau(z) + \frac{\tau c - 1}{z}.
    \end{equation*}
    Since \(\tau x_\circ\neq 0\), the term \((\tau c - 1)/z\) is analytic at \(z = \tau x_\circ\). Hence, if the pushed-forward atom has mass \(w(\tau)\),
    \begin{equation*}
        u(z) = -\frac{\tau c \, w(\tau)}{z-\tau x_\circ} + O(1).
    \end{equation*}
    Comparing residues gives \(\tau c \, w(\tau) = c w_\circ \alpha\), and so
    \begin{equation*}
        w(\tau) = \frac{w_\circ}{\tau} \left(\tau-\frac{\tau-1}{c w_\circ} \right) = w_\circ-\frac{\tau-1}{c \tau}.
    \end{equation*}
    
    
    
    Now suppose \(x_\circ = 0\). Then \(y(\zeta) = -\frac{w_\circ}{\zeta} + O(1)\), so the initial companion transform satisfies
    \begin{equation*}
        u(\zeta) = c y(\zeta) + \frac{c-1}{\zeta} = -\frac{\kappa_\circ}{\zeta} + O(1), \qquad \kappa_\circ \coloneqq 1 - c + c w_\circ.
    \end{equation*}
    Since \(\underline R_\tau = \tau \underline R_1\), the companion law evolves by a free additive convolution power. Its atom at zero therefore has weight
    \begin{equation*}
        \kappa(\tau) = \max \left\{ \tau \kappa_\circ - (\tau - 1), 0 \right\}.
    \end{equation*}
    On the decompressed side, \(\underline m_\tau(z) = \tau c \, m_\tau(z) + \frac{\tau c - 1}{z}\), so comparing residues at zero gives \(\kappa(\tau) = 1-\tau c+\tau c \, w(\tau)\). Consequently,
    \begin{equation*}
        w(\tau) = \frac{\kappa(\tau) + \tau c-1}{\tau c} = \max \left\{w_\circ, 1 - \frac{1}{c \tau} \right\}.
    \end{equation*}
    The atom remains located at the origin.
\end{proof}

\subsection{Evolution of Moments} \label{sec:evo-moments}

We now turn to computing the moments of the evolving measures \(\nu_\tau\) under deformed decompression starting from $\nu \equiv \nu_1$. For \(n \geq 0\), define the time-dependent raw moment
\begin{equation*}
	\mu_n(\tau) \coloneqq \int x^n  \mathrm{d} \nu_\tau (x).
\end{equation*}
Much like the free decompression case in \citet{AMELI-2026}, these moments can be evolved without reconstructing $\nu_\tau$. The deformed decompression PDE in \Cref{prop:stieltjes-pde} induces a closed triangular system of ODEs for $\mu_n(\tau)$, allowing each moment to be computed from lower-order moments.

\begin{lemma}[ODEs for Moments] \label{lem:moment}
    The evolution of the moment $\mu_n(\tau)$ under deformed decompression is governed by the system of triangularly coupled ODEs
    \begin{equation}
        \label{eq:moment-ode}
        \frac{\mathrm{d}\mu_n}{\mathrm{d}\tau} = c \sum_{j=1}^{n-1} \mu_{n-j}(\tau)\left(j \mu_j(\tau) - \tau \frac{\mathrm{d}\mu_j}{\mathrm{d}\tau}\right),\qquad n \geq 1,
    \end{equation}
    and, in particular, $\mu_0(\tau) = 1$ and $\mu_1(\tau) = \mu_1(1)$ for all $\tau \geq 1$.
\end{lemma}

\begin{proof}
    We begin with the Stieltjes transform PDE for deformed decompression \eqref{eq:stieltjes-pde-prop}, so
    \[
        (cz\tau m_\tau + c\tau - 1)\frac{\partial m_\tau}{\partial \tau} + c(m_\tau z + 1)\left(z \frac{\partial m_\tau}{\partial z} + m_\tau\right) = 0.
    \]
    For large $|z|$, $m_\tau(z)$ has an expansion as a convergent Laurent series $m_\tau(z) = -\sum_{n=0}^\infty \mu_n(\tau) z^{-n-1}$. Since $\mu_0(\tau) = 1$,
    \[
        1 + zm_\tau = -\sum_{n=1}^\infty \frac{\mu_n(\tau)}{z^n}, \quad\mbox{and}\quad z\frac{\partial m_\tau}{\partial z} + m_\tau = \sum_{n=0}^\infty \frac{n\mu_n(\tau)}{z^{n+1}}.
    \]
    Therefore, by the Cauchy product, the second term of the PDE \eqref{eq:stieltjes-pde-prop} becomes
    \begin{align}
        c(m_\tau z + 1)\left(z \frac{\partial m_\tau}{\partial z} + m_\tau\right) &= -c\left(\sum_{k=1}^\infty \frac{\mu_k(\tau)}{z^k}\right)\left(\sum_{j=0}^\infty \frac{j \mu_j(\tau)}{z^{j+1}}\right) \nonumber \\
        &= -c\sum_{n=1}^\infty \left(\sum_{j=0}^{n-1} j\mu_{n-j}(\tau)\mu_j(\tau)\right)\frac{1}{z^{n+1}}. 
        \label{eq:moment-pde-part2}
    \end{align}
    For the first term of the PDE, there is $cz\tau m_\tau + c\tau - 1 = -c\tau \sum_{n=1}^\infty \mu_n(\tau) z^{-n} - 1$ and $\frac{\partial m_\tau}{\partial \tau} = -\sum_{n=0}^\infty \mu_n'(\tau) z^{-n-1}$. Their Cauchy product yields
    \begin{equation}
        \label{eq:moment-pde-part1}
        (cz\tau m_\tau + c \tau - 1)\frac{\partial m_\tau}{\partial \tau} = \sum_{n=0}^\infty \frac{\mu_n'(\tau)}{z^{n+1}} + c\tau \sum_{n=1}^\infty \left(\sum_{j=0}^{n-1}\mu_{n-j}(\tau)\mu_j'(\tau)\right)\frac{1}{z^{n+1}}.
    \end{equation}
    Combining \eqref{eq:moment-pde-part1} and \eqref{eq:moment-pde-part2} with \eqref{eq:stieltjes-pde-prop} and equating the coefficient of $z^{-n-1}$ with zero gives
    \begin{equation}
        \label{eq:moment-ode-final}
        \mu_n'(\tau) + c\tau \sum_{j=0}^{n-1}\mu_{n-j}(\tau)\mu_j'(\tau) - c \sum_{j=0}^{n-1}j\mu_{n-j}(\tau)\mu_j(\tau) = 0.
    \end{equation}
    For $n = 0$, this gives $\mu_0'(\tau) = 0$, as expected. Consequently, this implies that the lower limit in the sums of \eqref{eq:moment-ode-final} can start from $j=1$. For $n = 1$, this becomes an empty sum, and so $\mu_1'(\tau) = 0$, implying $\mu_1(\tau)$ is constant.
\end{proof}

The triangular structure of \cref{lem:moment} implies that the moments can be solved for recursively. Specifically, $\mu_n(\tau)$ takes the form of a polynomial in $\tau$:
\[
    \mu_n(\tau) = \sum_{k=0}^{n-1} \kappa_{n,k} \tau^k,
\]
where the coefficients $\kappa_{n,k}$ are determined recursively from the initial moments $\mu_k(\tau = 1)$. This recurrence is established in the following proposition. 

\begin{proposition}[Moment Recurrence] \label{prop:moment-decompression}
    For any $\tau \geq 1$ and $n \geq 1$, the $n$-th moment $\mu_n(\tau)$ is a polynomial in $\tau$ of degree at most $n - 1$ given by $\mu_n(\tau) = \sum_{k=0}^{n-1}\kappa_{n,k}\tau^k$ where $\kappa_{1,0} = \mu_1(1)$ and for $n \geq 2$,
    \[
        \kappa_{n,k} = \begin{cases}
            \displaystyle\frac{c}{k}\displaystyle\sum_{i=1}^{n-1}\displaystyle\sum_{j=0}^{k-1}(i-j)\kappa_{i,j}\kappa_{n-i,k-1-j} & \text{ if }1 \leq k \leq n-1,\\
            \mu_n(1) - \displaystyle\sum_{j=1}^{n-1}\kappa_{n,j} & \text{ if } k = 0,
        \end{cases}
    \]
    and where $\kappa_{n,k} = 0$ if $k \geq n$. 
\end{proposition}

\begin{proof}
    First, we substitute the polynomial ansatz $\mu_n(\tau) = \sum_{k=0}^{n-1}\kappa_{n,k}\tau^k$ into the system of ODEs obtained in \Cref{lem:moment}. The left-hand side gives
    \begin{equation}
        \label{eq:moment-ode-lhs}
        \frac{\mathrm{d}\mu_n}{\mathrm{d}\tau} = \sum_{k=1}^{n-1} k \kappa_{n,k} \tau^{k-1} = \sum_{p=0}^{n-2} (p+1) \kappa_{n,p+1} \tau^p.
    \end{equation}
    For the right-hand side, starting with the bracketed term,
    \[
        i \mu_i(\tau) - \tau \frac{\mathrm{d}\mu_i}{\mathrm{d}\tau} = \sum_{j=0}^{i-1} i \kappa_{i,j}\tau^j - \sum_{j=1}^{i-1} j \kappa_{i,j} \tau^j = \sum_{j=0}^{i-1} (i-j) \kappa_{i,j} \tau^j.
    \]
    Multiplying this by $\mu_{n-i}(\tau) = \sum_{m=0}^{n-i-1} \kappa_{n-i,m}\tau^m$ gives
    \begin{align}
        \mu_{n-i}(\tau)\left(i\mu_i(\tau) - \tau\frac{\mathrm{d}\mu_i}{\mathrm{d}\tau}\right) &= \left(\sum_{m=0}^{n-i-1}\kappa_{n-i,m}\tau^m\right)\left(\sum_{j=0}^{i-1}(i-j)\kappa_{i,j}\tau^j\right) \nonumber\\
        &= \sum_{p=0}^{n-2}\left(\sum_{j=0}^p (i-j)\kappa_{i,j}\kappa_{n-i,p-j}\right) \tau^p,\label{eq:moment-ode-rhs}
    \end{align}
    where we define $p = m +j$ and adopt the convention that $\kappa_{a,b} = 0$ when $b \geq a$. Substituting \eqref{eq:moment-ode-lhs} and \eqref{eq:moment-ode-rhs} into the ODE \eqref{eq:moment-ode} gives
    \[
        \sum_{p=0}^{n-2}(p+1)\kappa_{n,p+1}\tau^p = c\sum_{i=1}^{n-1}\sum_{p=0}^{n-2}\left(\sum_{j=0}^p (i - j)\kappa_{i,j}\kappa_{n-i,p-j}\right) \tau^p.
    \]
    Equating the coefficients of $\tau^p$ for $0 \leq p \leq n-2$ on both sides gives
    \[
        (p+1)\kappa_{n,p+1} = c\sum_{i=1}^{n-1}\sum_{j=0}^p (i - j) \kappa_{i,j} \kappa_{n-i,p-j}.
    \]
    Setting $k = p + 1$ yields the recursive formula for $\kappa_{n,k}$ for $1 \leq k \leq n-1$. Finally, evaluating the moment representation at the initial state $\tau = 1$ requires $\mu_n(1) = \sum_{k=0}^{n-1}\kappa_{n,k}$, which rearranges to provide the expression for $\kappa_{n,0}$.
\end{proof}


\section{Generalized Covariance Models}
\label{sec:sep-covariance}

In this section, we consider the generalized covariance matrix model and study the difference between its limiting Stieltjes transform and that of a corresponding covariance matrix model which can be treated using deformed decompression. Our exposition and tools closely follow \citet{couillet2014analysis}; for further details on the generalized covariance model, see \citet{COUILLET-2022}. Since covariance matrices are a subset of generalized covariance matrices, for the sake of brevity, we shall temporarily adopt a more general notation, and later connect our findings to the central model presented in \Cref{sec:dd-intro}. 

A \emph{generalized covariance matrix (GCM)} is of the general form
\[
    \tens{S}_n \coloneqq \frac{1}{n} \tens{A}_n^{1/2} \tens{Z}_n \tens{B}_n \tens{Z}_n^\intercal \tens{A}_n^{1/2},
\]
where $\tens{Z}_n \in \mathbb{R}^{p \times n}$ is comprised of independent and identically distributed elements with zero mean and unit variance, and $\tens{A}_n \in \mathbb{R}^{p \times p}$, $\tens{B}_n \in \mathbb{R}^{n\times n}$ are deterministic positive-semidefinite matrices. Consider the limit $n \to \infty$ with $p \equiv p(n)$ where $p(n) / n \to c > 0$, where $c$ is the aspect ratio. In this regime, we assume the following.

\begin{assumption}
    \label{ass:gcm}
    The empirical spectral measures of $\tens{A}_n$ and $\tens{B}_n$ converge weakly to the probability measures $\nu$ and $\eta$, respectively, with support on $[0,\infty)$.
\end{assumption}

Under \Cref{ass:gcm}, from \citet[Proposition 1.1]{couillet2014analysis}, the empirical spectral distribution of $\tens{S}_n$ converges weakly to a probability measure; we shall denote the Stieltjes transform of this measure by $m_\nu^\eta$. We will generally keep $\nu$ fixed, and consider the effect of changing $\eta$. For a fixed probability measure $\eta$, we shall compare $m_\nu^\Lambda$ between the following choices for $\Lambda$:
\begin{itemize}
    \item An arbitrary probability measure $\eta$ with support on $[0,\infty)$. This is the complete generalized covariance model.
    \item A Dirac measure $\delta_{\bar{\eta}}$ assigning total probability mass to the point $\bar{\eta} = \int x \eta(\mathrm{d}x)$. This corresponds to the covariance model approximation, where $\tens{B}_n$ is replaced with $\frac{1}{n}\trace(\tens{B}_n) \tens{I}$.
\end{itemize}
For each $z \in \mathbb{C}^+$, define $\delta_{1}(z),\tilde{\delta}_{1}(z) \in \mathbb{C}^+$ as the unique solution to the master equations
\[
    \delta_{1}(z) =c\int\frac{t}{-z(1+\tilde{\delta}_{1}(z)t)}\nu(\mathrm{d}t),\qquad 
    \tilde{\delta}_{1}(z) =\int\frac{t}{-z(1+\delta_{1}(z)t)}\eta(\mathrm{d}t).
\]
By \citet[eqn. (5)]{couillet2014analysis}, the Stieltjes transform $m^\eta_\nu$ is given by
\[
    m_\nu^\eta(z) = -z^{-1} - c^{-1} \delta_1(z) \tilde{\delta}_1(z).
\]
Next, we define the same objects for the Dirac case: for each $z \in \mathbb{C}^+$, let $\delta_{2}(z),\tilde{\delta}_{2}(z) \in\mathbb{C}^{+}$ satisfy
\[
    \delta_{2}(z) =c\int\frac{t}{-z(1+\tilde{\delta}_{2}(z)t)}\nu(dt), \qquad 
    \tilde{\delta}_{2}(z) =\frac{\bar{\eta}}{-z(1+\delta_{2}(z)\bar{\eta})},
\]
with corresponding Stieltjes transform $m^{\delta_{\bar{\eta}}}_\nu$ given by
\[
    m^{\delta_{\bar{\eta}}}_\nu(z) = -z^{-1} - c^{-1} \delta_2(z) \tilde{\delta}_2(z).
\]
Our main theorem for this section bounds the discrepancy between $m_\nu^\eta$ and $m_\nu^{\delta_{\bar{\eta}}}$. 

\begin{theorem}
    \label{thm:app-variance-bound}
    Let $\eta$ have finite second moment, and let $\Omega \subset \mathbb{C}^{+}$ be a bounded region on the complex plane with positive distance from the real axis. There exists $K_{\Omega}>0$
    such that for $\sigma_\eta^2$ the variance of $\eta$,
    \[
        \sup_{z\in\Omega}|m^{\eta}_{\nu}(z)-m^{\delta_{\bar{\eta}}}_{\nu}(z)|\leq K_{\Omega}\cdot\sigma^{2}_{\eta}.
    \]
\end{theorem}

\begin{proof}
    Let $z\in\mathbb{C}^{+}$ be fixed and define the function
    \[
        f_{z}(t)=\frac{t}{-z(1+\delta_{2}(z)t)},\qquad t \in [0,\infty),
    \]
    noting that $\tilde{\delta}_{2}(z)=f_{z}(\bar{\eta})$. The primary source of the error between $m_\nu^\eta$ and $m_\nu^{\delta_{\bar{\eta}}}$ is captured in the term $B(z)$, defined as the difference
    \[
        B(z)=\int f_{z}(t)\eta(\mathrm{d}t)-f_{z}(\bar{\eta}).
    \]
    To bound $B(z)$, a second-order Taylor expansion of $f_{z}(t)$ around
    $\bar{\eta}$ will suffice:
    \[
        f_{z}(t)=f_{z}(\bar{\eta})+f_{z}'(\bar{\eta})(t-\bar{\eta})+\frac{1}{2}f_{z}''(\xi_{t})(t-\bar{\eta})^{2},\qquad \xi_t \in (\bar{\eta},t).
    \]
    Integrating with respect to $\eta$,
    the linear term cancels, and
    \begin{equation}
        \label{eq:gcm-b-bound}
        \left|B(z)\right|\leq\frac{\sigma^{2}_{\eta}}{2}\sup_{t\geq0}|f_{z}''(t)|.
    \end{equation}
    The second derivative of $f_{z}$ is given by $f_{z}''(t)=\frac{2\delta_{2}(z)}{z(1+\delta_{2}(z)t)^{3}}$.
    From the proof of \citet[Proposition 2.1]{couillet2014analysis}, for any $t\geq0$ and $z\in\mathbb{C}^{+}$, $|z(1+\delta_{2}(z)t)|\geq\mathrm{Im}(z)>0$.
    Consequently, $|1+\delta_{2}(z)t|^{-3}\leq\frac{|z|^{3}}{\mathrm{Im}(z)^{3}}$.
    Furthermore, \citet[Lemma 3.1]{couillet2014analysis} guarantees that there exists $K^{(1)}_{\Omega}>0$
    such that
    \begin{equation*}
        \sup_{z\in\Omega}\max\{|\delta_{1}(z)|,|\delta_{2}(z)|,|\tilde{\delta}_{1}(z)|,|\tilde{\delta}_{2}(z)|\}\leq K^{(1)}_{\Omega}.
    \end{equation*}
    Letting $K^{(2)}_{\Omega}=\sup_{z\in\Omega}\frac{|z|^{2}}{\mathrm{Im}(z)^{3}}$,
    \begin{equation}
        \label{eq:gcm-secondderiv-bound}
        \sup_{t\geq0}|f_{z}''(t)|\leq2K^{(1)}_{\Omega}K^{(2)}_{\Omega}.
    \end{equation}
    Now let $\Delta\delta=\delta_{1}-\delta_{2}$ and $\Delta\tilde{\delta}=\tilde{\delta}_{1}-\tilde{\delta}_{2}$,
    and observe that
    \begin{align*}
        \Delta\tilde{\delta} & =\int\frac{t}{-z(1+\delta_{1}(z)t)}\eta(\mathrm{d}t)-\frac{\bar{\eta}}{-z(1+\delta_{2}(z)\bar{\eta})}\\
         & =\int\left(\frac{t}{-z(1+\delta_{1}(z)t)}-\frac{t}{-z(1+\delta_{2}(z)t)}\right)\eta(\mathrm{d}t)+B(z).
    \end{align*}
    Combining the fractions yields
    \begin{align*}
        \frac{t}{-z(1+\delta_{1}(z)t)}-\frac{t}{-z(1+\delta_{2}(z)t)} & =\frac{t(1+\delta_{2}(z)t)-t(1+\delta_{1}(z)t)}{-z(1+\delta_{1}(z)t)(1+\delta_{2}(z)t)}\\
         & =\frac{t^{2}}{z}\frac{\delta_{1}(z)-\delta_{2}(z)}{(1+\delta_{1}(z)t)(1+\delta_{2}(z)t)}.
    \end{align*}
    Consequently,
    \begin{equation}
        \label{eq:scm-deltatilde}
        \Delta\tilde{\delta}  =z\Delta\delta\int\frac{t^{2}}{z^{2}(1+\delta_{1}(z)t)(1+\delta_{2}(z)t)}\eta(\mathrm{d}t)+B(z)\eqqcolon z\Gamma_{\eta}(z)\Delta\delta+B(z).
    \end{equation}
    Applying the same fraction combination to $\Delta\delta$ now under
    the shared measure $\nu$,
    \begin{align*}
        \Delta\delta & =c\int\left[\frac{t}{-z(1+\tilde{\delta}_{1}(z)t)}-\frac{t}{-z(1+\tilde{\delta}_{2}(z)t)}\right]\nu(\mathrm{d}t)\\
         & =z\Delta\tilde{\delta}\cdot c\int\frac{t^{2}}{z^{2}(1+\tilde{\delta}_{1}(z)t)(1+\tilde{\delta}_{2}(z)t)}\nu(\mathrm{d}t)\\
         & \eqqcolon z\Gamma_{\nu}(z)\Delta\tilde{\delta}.
    \end{align*}
    Substituting $\Delta\delta$ into \eqref{eq:scm-deltatilde} reveals
    \begin{equation}
        \label{eq:gcm-deltaeq}
        \Delta\tilde{\delta}(1-z^{2}\Gamma_{\eta}(z)\Gamma_{\nu}(z))=B(z),
    \end{equation}
    where
    \begin{align*}
        \Gamma_{\eta}(z) & = \int\frac{t^{2}}{z^{2}(1+\delta_{1}(z)t)(1+\delta_{2}(z)t)}\eta(\mathrm{d}t)\\
        \Gamma_{\nu}(z) & = \int\frac{c t^{2}}{z^{2}(1+\tilde{\delta}_{1}(z)t)(1+\tilde{\delta}_{2}(z)t)}\nu(\mathrm{d}t).
    \end{align*}
    Since the difference in the Stieltjes transforms $\Delta m = m_\nu^{\eta} - m_\nu^{\delta_{\bar{\eta}}}$ satisfies
    \[
        c\left|\Delta m(z)\right|=\left|\delta_{1}\tilde{\delta}_{1}-\delta_{2}\tilde{\delta}_{2}\right|=\left|\delta_{1}\Delta\tilde{\delta}+\tilde{\delta}_{2}\Delta\delta\right|\leq\left|\Delta\tilde{\delta}\right|\left|\delta_{1}+\tilde{\delta}_{2}z\Gamma_{\nu}(z)\right|
        =\frac{|B(z)|\left|\delta_{1}+\tilde{\delta}_{2}z\Gamma_{\nu}(z)\right|}{|1-z^{2}\Gamma_{\eta}(z)\Gamma_{\nu}(z)|},
    \]
    this implies from \eqref{eq:gcm-b-bound}, \eqref{eq:gcm-secondderiv-bound}, and \eqref{eq:gcm-deltaeq}, that
    \begin{align*}
        \sup_{z\in\Omega}\left|\Delta m(z)\right| & \leq\frac{1}{c}\frac{\sigma^{2}_{\eta}K^{(1)}_{\Omega}K^{(2)}_{\Omega}}{|1-z^{2}\Gamma_{\eta}(z)\Gamma_{\nu}(z)|}\left|\delta_{1}+\tilde{\delta}_{2}z\Gamma_{\nu}(z)\right|\\
         & \leq\frac{1}{c}\sigma^{2}_{\eta}(K^{(1)}_{\Omega})^{2}K^{(2)}_{\Omega}\cdot\frac{1+\sup_{z\in\Omega}|z\Gamma_{\nu}(z)|}{\inf_{z\in\Omega}|1-z^{2}\Gamma_{\eta}(z)\Gamma_{\nu}(z)|}
    \end{align*}
    It only remains to deal with the last fraction. First, we establish
    that $\inf_{z\in\Omega}|1-z^{2}\Gamma_{\eta}(z)\Gamma_{\nu}(z)|>0$.
    To do so, let
    \begin{align*}
        \gamma_{1}(z_{1},z_{2}) & = \int\frac{c t^{2}}{z_{1}z_{2}(1+\tilde{\delta}_{1}(z_{1})t)(1+\tilde{\delta}_{1}(z_{2})t)}\nu(\mathrm{d}t)\\
        \gamma_{2}(z_{1},z_{2}) & = \int\frac{c t^{2}}{z_{1}z_{2}(1+\tilde{\delta}_{2}(z_{1})t)(1+\tilde{\delta}_{2}(z_{2})t)}\nu(\mathrm{d}t)\\
        \tilde{\gamma}_{1}(z_{1},z_{2}) & = \int\frac{t^{2}}{z_{1}z_{2}(1+\delta_{1}(z_{1})t)(1+\delta_{1}(z_{2})t)}\eta(\mathrm{d}t)\\
        \tilde{\gamma}_{2}(z_{1},z_{2}) & = \int\frac{t^{2}}{z_{1}z_{2}(1+\delta_{2}(z_{1})t)(1+\delta_{2}(z_{2})t)}\eta(\mathrm{d}t)
    \end{align*}
    By the Cauchy-Schwarz inequality,
    \begin{align*}
        |\Gamma_{\eta}(z)| & \leq\tilde{\gamma}_{1}(z,\bar{z})^{1/2}\tilde{\gamma}_{2}(z,\bar{z})^{1/2}\\
        |\Gamma_{\nu}(z)| & \leq\gamma_{1}(z,\bar{z})^{1/2}\gamma_{2}(z,\bar{z})^{1/2}.
    \end{align*}
    Together this reveals that
    \[
        |z^{2}\Gamma_{\eta}(z)\Gamma_{\nu}(z)|\leq\sqrt{|z|^{2}\gamma_{1}(z,\bar{z})\tilde{\gamma}_{1}(z,\bar{z})}\cdot\sqrt{|z|^{2}\gamma_{2}(z,\bar{z})\tilde{\gamma}_{2}(z,\bar{z})}.
    \]
    Let $h_{i}(z)=|z|^{2}\gamma_{i}(z,\bar{z})\tilde{\gamma}_{i}(z,\bar{z})$
    for $i=1,2$. From \citet[eqn. (8)]{couillet2014analysis}, $h_{i}(z)<1$ for any $z\in\mathbb{C}^{+}$.
    Since $h_{i}$ is also continuous on $\mathbb{C}^{+}$, by the Extreme
    Value Theorem, $h_{i}$ must achieve its maximum on $\bar{\Omega}$,
    and $\max_{i=1,2}\sup_{z\in\Omega}h_{i}(z)\leq1-\epsilon$ for some
    $\epsilon>0$. Consequently,
    \[
        \inf_{z\in\Omega}|1-z^{2}\Gamma_{\eta}(z)\Gamma_{\nu}(z)|\geq1-\sup_{z\in\Omega}\sqrt{h_{1}(z)h_{2}(z)}\geq\epsilon.
    \]
    Finally, note that
    \[
        |z\Gamma_{\nu}(z)|\leq|z|\gamma_{1}(z,\bar{z})^{1/2}\gamma_{2}(z,\bar{z})^{1/2}.
    \]
    Since $|z|\leq M$ for $z\in\Omega$ for some $M>0$ and \citet[Lemma 3.2]{couillet2014analysis} implies that $\gamma_{i}(z,\bar{z})$ is uniformly bounded on $\Omega$ (by $M_\gamma$, say), this completes the proof with
    \begin{equation*}
        K_\Omega = (c \epsilon)^{-1} (K_\Omega^{(1)})^2 K_\Omega^{(2)} (1 + M M_\gamma).
    \end{equation*} 
\end{proof}

\Cref{thm:app-variance-bound} immediately implies \Cref{thm:variance-bound} with $\tens{A}_n = \tens{\Sigma}_p$, $\tens{B}_n = \tens{D}$, and $\tilde{\tens{\Sigma}}_p = \frac{1}{n}\trace(\tens{D})\tens{\Sigma}_p$.

For a fixed \(y > 0\), \Cref{thm:app-variance-bound} also provides a bound on the Poisson-smoothed spectral densities
\begin{equation*}
    \rho_{\nu,y}^{\Lambda}(x) \coloneqq \pi^{-1}\Im m_\nu^\Lambda(x+iy)
    \quad \text{for} \quad
    \Lambda \in \{\eta, \delta_{\bar{\eta}}\}.
\end{equation*}
Indeed, for any compact interval \(I \subset \mathbb{R}\) and any \(\Omega\) as in \Cref{thm:app-variance-bound} containing \(\{x+iy : x \in I\}\),
\begin{equation*}
    \sup_{x \in I} \left|\rho_{\nu,y}^{\eta}(x) - \rho_{\nu,y}^{\delta_{\bar{\eta}}}(x)\right| \leq \frac{K_\Omega}{\pi}\sigma_\eta^2.
\end{equation*}
This follows directly by taking imaginary parts of the Stieltjes-transform bound. Since the bound applies at a fixed height \(y > 0\), it does not by itself provide control of the unsmoothed density, exact spectral edges, or log-determinants.

Given the fine level of granularity available for examining GCMs, one might consider whether it is feasible to develop a generalization of deformed decompression that is directly capable of extrapolating from the limiting spectral measure $\nu_c$ of $\tens{S}_n$ to that of a larger aspect ratio $\nu_{\tau c}$. This turns out to be impossible, at least, using the available techniques. To see why, assume that $\tens{Z}_n$ is comprised of i.i.d. standard normal random variables. The S-transform $S(z)$ of the limiting spectral distribution of $\tens{S}_n$ should factorize into three parts:
\[
    S(z) = S_{\tens{A}}(z) S_{\mathrm{MP}_c}(z) S_{\tens{B}}(c z),
\]
where the presence of $c$ in the S-transform for $\tens{B}$ addresses the change in scale due to the different sizes of $\tens{A}$ and $\tens{B}$. In practice, we have knowledge of $S$ and $S_{\mathrm{MP}_c}$, but not of $S_{\tens{A}}$ or $S_{\tens{B}}$. Due to the presence of $c$ in the scale factor for $S_{\tens{B}}$, to extrapolate from smaller to larger matrix sizes in the GCM setting requires solving for \emph{both} $S_{\tens{A}}$ and $S_{\tens{B}}$. With only one equation and two unknowns, this is unachievable from the information available. While it may be possible to develop some extrapolation procedure using this framework, the resulting method cannot behave as a single-parameter flow. 


\section{Random Matrix Models} \label{sec:ensemble-models}

This section describes the matrix models and Gram covariance matrices used in our numerical experiments. We first use a closed-form deformed Marchenko--Pastur model as a controlled benchmark, where the limiting spectral law is known algebraically and the effect of changing the aspect ratio can be tested directly (\Cref{sec:deformed-mp}). We then consider Hessian matrices arising from logistic regression (\Cref{sec:hessian-example}) and an empirical image Gram covariance matrix formed from the STL-10 dataset (\Cref{sec:stl10}).


\subsection{Deformed Marchenko--Pastur Law}
\label{sec:deformed-mp}

The deformed Marchenko--Pastur law is the canonical model for deformed decompression. It arises as the limiting eigenvalue distribution of a sample covariance matrix with a nontrivial population covariance \citep[Section 9.7]{BAI-2010}. Let \(\tens{Z} \in \mathbb{R}^{p \times n}\) have independent centered entries with variance one, let \(\gtens{\Sigma} \in \mathbb{R}^{p \times p}\) be a deterministic positive semidefinite matrix, and define
\begin{equation*}
	\tens{X} \coloneqq \gtens{\Sigma}^{1 / 2} \tens{Z},
	\qquad
	\tens{S} \coloneqq \frac{1}{n} \tens{X} \tens{X}^{\intercal}.
	\label{eq:dmp-matrix-model}
\end{equation*}
Assume that \(p / n \to c \in (0, \infty)\), and that the empirical spectral distribution of \(\gtens{\Sigma}\) converges weakly to a compactly supported probability measure \(\nu_{\gtens{\Sigma}}\) on \([0, \infty)\). Then, under standard assumptions, the empirical spectral distribution of \(\tens{S}\) converges weakly to a deterministic probability measure \(\nu_c\), given by the free multiplicative convolution
\begin{equation}
	\nu_c = \nu_{\gtens{\Sigma}} \boxtimes \mathrm{MP}_c,
	\label{eq:dmp-law}
\end{equation}
where \(\mathrm{MP}_c\) denotes the Marchenko--Pastur law with aspect ratio \(c > 0\). The mass at the origin is \(\max\{\nu_{\gtens{\Sigma}}(\{0\}), 1 - 1/c\}\).

For the explicit algebraic examples used in this paper, we take the population law \(\nu_{\gtens{\Sigma}}\) to have finitely many atoms,
\begin{equation}
	\nu_{\gtens{\Sigma}}(\mathrm{d} t) \coloneqq \sum_{i = 1}^{r} w_i \delta_{t_i}(\mathrm{d} t),
	\qquad
	t_i > 0,
	\qquad
	w_i > 0,
	\qquad
    \sum_{i = 1}^{r} w_i = 1.
	\label{eq:dmp-finite-atom-population}
\end{equation}

A convenient characterization of \(\nu_c\) is given in terms of the \emph{companion} Stieltjes transform \(\underline{m}\)
\begin{equation}
	\underline{m}(z) \coloneqq c m(z) + \frac{c - 1}{z}.
	\label{eq:dmp-companion-transform}
\end{equation}
This companion transform reflects the rectangular geometry of the data matrix: it is the Stieltjes transform of the limiting empirical law of \(n^{-1} \tens{X}^{\intercal} \tens{X}\), and it satisfies the Silverstein self-consistent equation \citep{SILVERSTEIN-1995}
\begin{equation}
	-\frac{1}{\underline{m}(z)} = z - c \int_{\mathbb{R}} \frac{t}{1 + t \underline{m}(z)} \, \nu_{\gtens{\Sigma}}(\mathrm{d} t).
	\label{eq:dmp-silverstein}
\end{equation}
For the finite-atom population law \eqref{eq:dmp-finite-atom-population}, this reduces to
\begin{equation}
	-\frac{1}{\underline{m}(z)} = z - c \sum_{i = 1}^{r} \frac{w_i t_i}{1 + t_i \underline{m}(z)}.
	\label{eq:dmp-silverstein-finite}
\end{equation}
Multiplying \eqref{eq:dmp-silverstein-finite} by the common denominator \(\underline{m} \prod_{i = 1}^{r} (1 + t_i \underline{m})\) yields the algebraic constraint
\begin{equation}
	P(z, \underline{m}) \coloneqq (1 + z \underline{m}) \prod_{i = 1}^{r} (1 + t_i \underline{m}) - c \underline{m} \sum_{i = 1}^{r} w_i t_i \prod_{j \neq i} (1 + t_j \underline{m}) = 0.
	\label{eq:dmp-companion-polynomial}
\end{equation}
Thus, when \(\nu_{\gtens{\Sigma}}\) has \(r\) atoms, the companion Stieltjes transform is an algebraic function of degree \(r + 1\). The physical branch is selected by the Herglotz property \(\Im \underline{m}(z) > 0\) for \(z \in \mathbb{C}^{+}\), equivalently \(\Im m(z) > 0\), together with the normalization at infinity.

\begin{example}[Two-atom deformed Marchenko--Pastur law]
	\label{ex:dmp-two-atom}
	For the two-atom population law \(\nu_{\gtens{\Sigma}} = w_1 \delta_{t_1} + w_2 \delta_{t_2}\), with \(w_1 + w_2 = 1\), the companion polynomial \eqref{eq:dmp-companion-polynomial} is cubic:
	\begin{equation}
		P(z, \underline{m}) = \underline{a}_3(z) \underline{m}^3 + \underline{a}_2(z) \underline{m}^2 + \underline{a}_1(z) \underline{m} + \underline{a}_0(z) = 0,
		\label{eq:dmp-companion-cubic}
	\end{equation}
	with coefficients
	\begin{align}
		\underline{a}_3(z) &= z t_1 t_2, \nonumber \\
		\underline{a}_2(z) &= z (t_1 + t_2) + t_1 t_2 (1 - c), \nonumber \\
		\underline{a}_1(z) &= z + (t_1 + t_2) - c (w_1 t_1 + w_2 t_2), \nonumber \\
		\underline{a}_0(z) &= 1.
		\label{eq:dmp-companion-cubic-coefficients}
	\end{align}
	Equivalently, using \eqref{eq:dmp-companion-transform}, the ordinary Stieltjes transform satisfies
	\begin{equation}
		P(z, m) = a_3(z) m^3 + a_2(z) m^2 + a_1(z) m + a_0(z) = 0,
		\label{eq:dmp-ordinary-cubic}
	\end{equation}
	where
	\begin{align}
		a_3(z) &= c^2 t_1 t_2 z^2, \nonumber \\
		a_2(z) &= c (t_1 + t_2) z^2 - 2 c t_1 t_2 (1 - c) z, \nonumber \\
		a_1(z) &= z^2 + \left(c (w_2 t_1 + w_1 t_2) + (c - 1) (t_1 + t_2)\right) z + (c - 1)^2 t_1 t_2, \nonumber \\
		a_0(z) &= z - (1 - c) (w_2 t_1 + w_1 t_2).
		\label{eq:dmp-ordinary-cubic-coefficients}
	\end{align}
	In particular, the two-atom deformed Marchenko--Pastur law has \(s = 3\) algebraic sheets and \(d_z = 2\).
\end{example}

\begin{figure}[t]
	\centering
	\includegraphics[width=\textwidth]{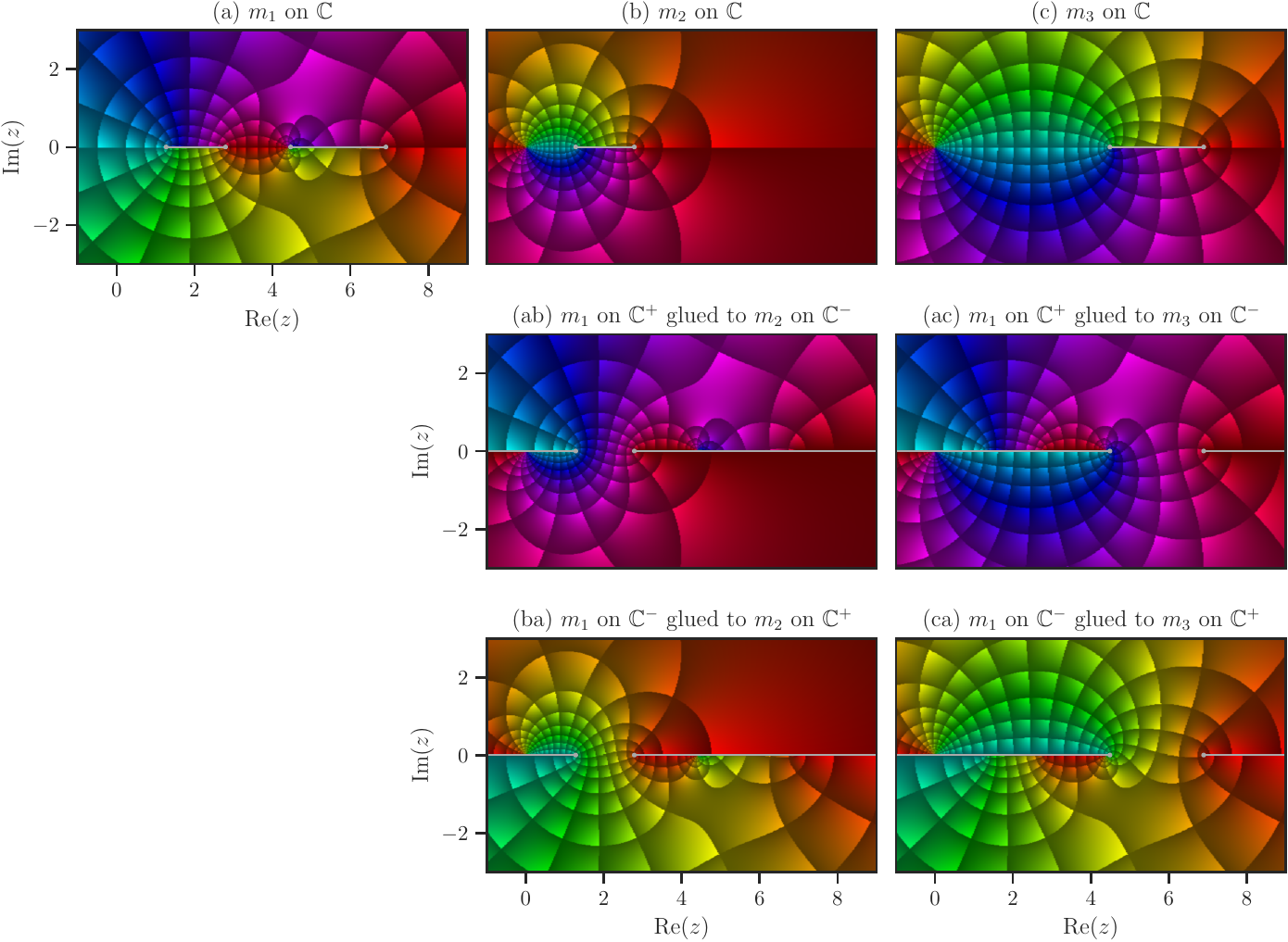}
	\caption{Branches and analytic continuations of the cubic deformed Marchenko--Pastur Stieltjes transform in \eqref{eq:dmp-ordinary-cubic}. The first row shows the three algebraic branches \(m_1, m_2, m_3\) on the \(z\)-plane, with branch cuts marked on the real axis. The physical branch is \(m_1\). The lower panels show the possible half-plane gluings involving the physical sheet: for example, \((ab)\) denotes \(m_1\) on \(\mathbb{C}^{+}\) glued to \(m_2\) on \(\mathbb{C}^{-}\), while \((ba)\) denotes the opposite gluing. Across a shared cut, the individual branches are discontinuous, but the glued half-planes form a continuous analytic continuation. Colors encode the complex value of the branch using the domain-coloring convention of \Cref{sec:notation}.}
	\label{fig:dmp-branches}
\end{figure}

The two-atom case is also a useful example for visualizing the spectral curve associated with the algebraic Stieltjes transform. Since \eqref{eq:dmp-ordinary-cubic} is cubic, each point \(z\) away from branch points has three algebraic roots, \(m_1, m_2, m_3\), displayed on the \(z\)-plane in the first row of \Cref{fig:dmp-branches}. Only one of these roots is the physical Stieltjes branch (here, \(m_1\)), selected by the Herglotz property and the normalization at infinity. The other two roots (\(m_2\) and \(m_3\)) are nonphysical branches, but they are not extraneous: they encode the analytic continuation of the physical branch across the real branch cuts. For the physical branch, these branch cuts coincide with the support intervals of the density on the real axis. The lower panels of \Cref{fig:dmp-branches} show the half-plane gluings that describe how the physical sheet continues through the cuts.

\begin{figure}[t]
	\centering
	\includegraphics[width=0.67\textwidth]{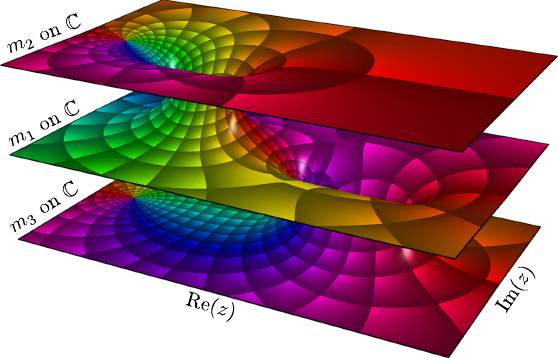}
	\caption{Three-sheet visualization of the cubic deformed Marchenko--Pastur spectral curve from \Cref{fig:dmp-branches}. The middle sheet represents the physical branch \(m_1\), while the upper and lower sheets represent the nonphysical branches \(m_2\) and \(m_3\). The sheets are vertically separated for visualization and are brought together along shared branch cuts to illustrate analytic continuation between branches. This geometric view explains why representing the Stieltjes transform as one sheet of a larger algebraic curve is useful: crossing a branch cut does not terminate the function, but moves the continuation to another sheet.}
	\label{fig:dmp-glue-sheets}
\end{figure}

The same information can be viewed as a three-sheeted algebraic surface, shown schematically in \Cref{fig:dmp-glue-sheets}. In this representation, the branch cuts are not endpoints of the Stieltjes transform, but seams where the physical sheet is attached to nonphysical sheets. Thus, crossing a branch cut does not terminate the analytic continuation; it moves the continuation from one sheet of the spectral curve to another. In this larger geometric picture, the physical Stieltjes branch is only one visible part of a connected algebraic object, and the continuation can be followed smoothly across sheets by passing through the appropriate cuts. This viewpoint is central to the polynomial method, since the algebraic relation keeps track of all branches simultaneously, even though only one branch represents the physical Stieltjes transform on \(\mathbb{C}^{+}\).


\paragraph{Relation to compound free Poisson.}
The deformed Marchenko--Pastur law is closely related to the compound free Poisson law used in free decompression; see \citep[p. 218]{NICA-2006} and \citep[Section I.1]{AMELI-2026}. In the finite-atom case, the companion law associated with \(\nu_c = \nu_{\gtens{\Sigma}} \boxtimes \mathrm{MP}_c\) is precisely the compound free Poisson law with jump distribution \(\nu_{\gtens{\Sigma}}\) and rate \(c\). This relation is most transparent at the matrix level. The nonzero eigenvalues of \(n^{-1} \tens{X} \tens{X}^{\intercal}\) and \(n^{-1} \tens{X}^{\intercal} \tens{X}\) agree, but the rectangular dimensions determine how many zero eigenvalues are added to one companion relative to the other. In the regime \(c < 1\), the companion law has an additional atom at zero \citep[Section 22.5.4]{SPEICHER-2015}. More precisely, if \(\underline{\nu}_c\) denotes the companion law associated with \(\nu_c\), then
\begin{equation*}
	\underline{\nu}_c = c \nu_c + (1 - c) \delta_0.
\end{equation*}
Hence, for \(c < 1\), the absolutely continuous part of \(\underline{\nu}_c\) has the same nonzero support as \(\nu_c\), but its density is scaled by the factor \(c\), and the remaining mass \(1 - c\) is placed at the origin.

\begin{figure}[t]
	\centering
    \ifjournal
	       \includegraphics[width=\textwidth]{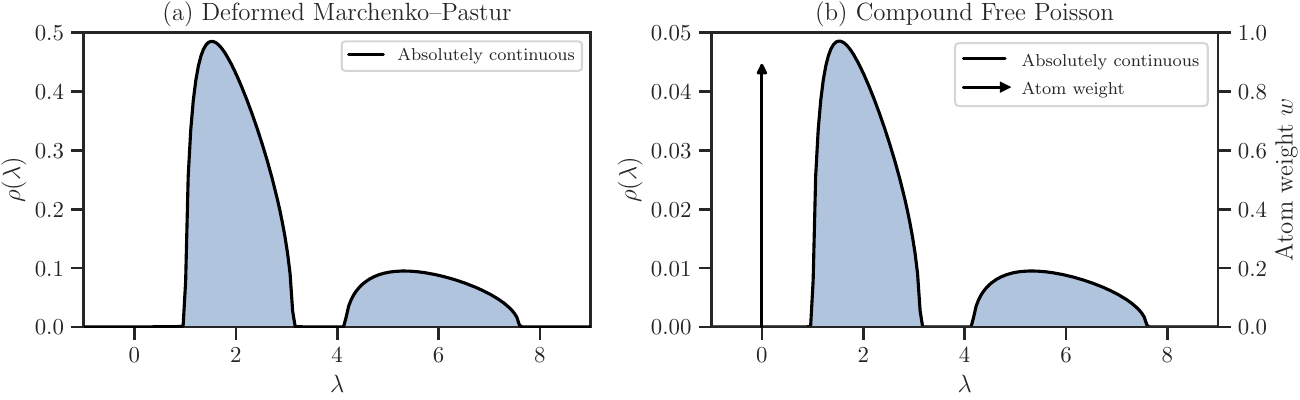}
    \else
	       \includegraphics[width=0.9\textwidth]{\figdir/dmp-density}
    \fi
	\caption{Comparison between the deformed Marchenko--Pastur law and its compound-free-Poisson companion for \((t_1, t_2) = (2, 11 / 2)\), \((w_1, w_2) = (3 / 4, 1 / 4)\), and \(c = \lambda = 1 / 10\). Panel (a) shows the deformed Marchenko--Pastur density \(\nu_c\), whose absolutely continuous part has total mass one and no atom at the origin. Panel (b) shows the corresponding compound free Poisson law. Its absolutely continuous part has the same nonzero support and the same profile scaled by \(c\), while the remaining mass \(1 - c = 0.9\) appears as an atom at the origin. In panel (b), the left ordinate corresponds to the continuous density, and the right ordinate corresponds to the atom weight indicated by the vertical arrow at the origin.}
	\label{fig:dmp-density}
\end{figure}

This relationship is illustrated in \Cref{fig:dmp-density} for the same two-atom parameters used above. The deformed Marchenko--Pastur law has two absolutely continuous bulks with total mass one and no atom at the origin, since \(c = 0.1 < 1\). Its compound-free-Poisson companion has the same two nonzero bulks after a vertical rescaling, but those bulks carry only mass \(c = 0.1\); the remaining mass \(1 - c = 0.9\) appears as an atom at zero. Thus, visually, the two nonzero density profiles look the same up to the scale of the vertical axis, even though the measures are different.

Despite this close algebraic and visual relationship, the two models belong to different decompression geometries. The compound free Poisson law is governed by the free-decompression \(R\)-transform invariance, while the deformed Marchenko--Pastur law is governed by the companion Silverstein structure used in deformed decompression. This distinction is not only formal: it changes the entire evolution path from a sampled matrix to the target matrix. Even when the final large-matrix laws are companion-related, the sampled submatrices used to initialize FD and DD generally have different spectra. Consequently, their initial supports and edge locations can be different, and the spectral edges follow different trajectories before reaching the final companion-related nonzero support.

\begin{figure}[t]
	\begin{minipage}[b]{0.37\textwidth}
        \includegraphics[width=\textwidth]{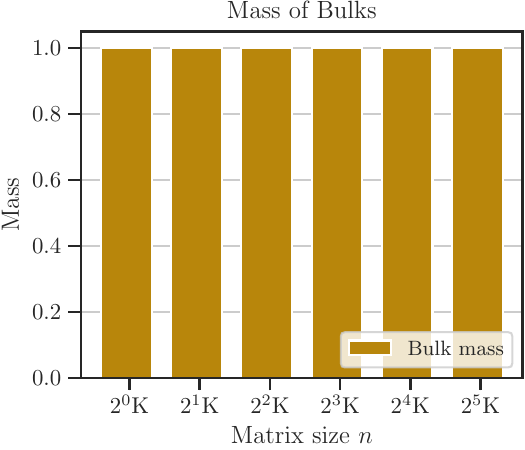}
    \end{minipage}%
    \hfill%
    \begin{minipage}[b]{0.61\textwidth}
        \includegraphics[width=\textwidth]{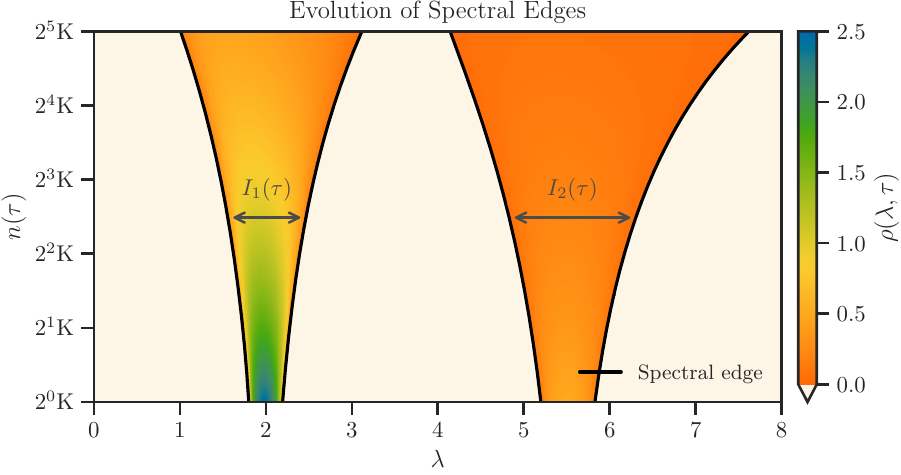}
    \end{minipage}
	\caption{Mass and edge evolution for the two-atom deformed Marchenko--Pastur example under deformed decompression from \(n = 2^0\mathrm{K}\) to \(n = 2^5\mathrm{K}\). Panel (a) shows that the full mass remains in the absolutely continuous bulks throughout the DD flow; unlike the compound-free-Poisson companion under free decompression, no mass is transferred to an atom at the origin. Panel (b) shows the evolution of the four spectral edges over the same range of matrix sizes. The color field shows the DD-evolved density \(\rho(\lambda, n)\), while the light background marks spectral gaps. The edge curves are computed independently from the algebraic branch-point evolution in \Cref{prop:dd-branch-point}, not extracted from the density plot. Their agreement with the visible boundary of the density support confirms the consistency between the evolved density and the DD edge dynamics.}
	\label{fig:dmp-atom}
\end{figure}

\Cref{fig:dmp-atom} shows this behavior for the deformed Marchenko--Pastur example. Panel (a) tracks the mass throughout the DD flow from \(n = 2^0\mathrm{K}\) to \(n = 2^5\mathrm{K}\). In contrast to the compound-free-Poisson FD example in \citep[Figure 1]{AMELI-2026}, where mass is transferred between the absolutely continuous bulks and an atom at the origin, the deformed MP law here has no atom for \(c < 1\), and the full mass remains in the absolutely continuous component throughout the evolution. Panel (b) tracks the four spectral edges of the two-bulk density over the same range of matrix sizes. The color field shows the DD-evolved density \(\rho(\lambda, \tau)\), while the light background indicates spectral gaps where the density vanishes. The edge curves are not extracted from the plotted density; they are evolved separately using the algebraic branch-point evolution equation in \Cref{prop:dd-branch-point}. Their agreement with the visible boundary of the density support provides a consistency check between the evolved density and the independently evolved DD edges.

\paragraph{Large-Ratio Numerical Validation.}
To assess the numerical accuracy of DD independently of finite-matrix sampling, we initialize the method from an analytically specified two-atom deformed Marchenko--Pastur model and compare the evolved distributions against their exact limiting laws for decompression ratios \(\tau = 2^{2j}\), \(j = 0, \ldots, 5\). \Cref{tab:dd-large-tau} reports the normalized Wasserstein-1 distance, maximum mean discrepancy (MMD), and errors in the first moment and standard deviation.

\begin{table}[t]
    \caption{Numerical validation of deformed decompression against exact deformed Marchenko--Pastur laws at increasing decompression ratios. All errors are percentages. For \(\tau = 2^{10}\), the laws have an atom at zero, so the reported errors compare their normalized absolutely continuous components.}
    \label{tab:dd-large-tau}
    \vspace{1mm}
    \centering
    \small
    \begin{tabular}{lrrrr}
        \toprule
        \(\tau\) & \(W_1/L\) (\%) & MMD (\%) & Mean error (\%) & Std error (\%) \\
        \midrule
        \(1    \)  & \(4.1 \times 10^{-7}\) & \(7.2 \times 10^{-7}\) & \(5.7 \times 10^{-7}\) & \(7.3 \times 10^{-7}\) \\
        \(2^2 \)    & \(3.7 \times 10^{-7}\) & \(7.2 \times 10^{-7}\) & \(5.8 \times 10^{-7}\) & \(7.1 \times 10^{-7}\) \\
        \(2^4 \)  & \(3.3 \times 10^{-7}\) & \(7.8 \times 10^{-7}\) & \(6.5 \times 10^{-7}\) & \(6.7 \times 10^{-7}\) \\
        \(2^6 \)  & \(3.9 \times 10^{-7}\) & \(1.1 \times 10^{-6}\) & \(1.1 \times 10^{-6}\) & \(5.9 \times 10^{-7}\) \\
        \(2^8 \)  & \(1.4 \times 10^{-5}\) & \(4.5 \times 10^{-5}\) & \(6.5 \times 10^{-5}\) & \(2.4 \times 10^{-6}\) \\
        \addlinespace[1mm]  
        \arrayrulecolor{gray!30}\cline{1-5} 
        \addlinespace[1mm]
        \arrayrulecolor{black}
        \(2^{10}\) & \(4.9 \times 10^{-8}\) & \(1.3 \times 10^{-7}\) & \(1.3 \times 10^{-7}\) & \(9.6 \times 10^{-8}\) \\
        \bottomrule
    \end{tabular}
\end{table}

The discrepancies remain extremely small across all tested ratios, with the largest reported error below \(7 \times 10^{-5}\%\). The nonzero errors at \(\tau = 1\) already reflect numerical fitting and Stieltjes inversion, rather than error from the DD evolution. These results demonstrate the numerical consistency of the algebraic DD implementation with its exact limiting laws over large decompression ratios. They do not measure the finite-sample errors incurred when initializing DD from an empirical submatrix.


\subsection{Hessian of Logistic Regression Model}\label{sec:hessian-example}

We next consider Hessian matrices from trained logistic regression models, following the generalized sample-covariance setting studied by \citet{LIAO-2021a}. In this model, the Hessian has the form
\begin{equation*}
	\tens{H} = \frac{1}{N}\tens{X}\tens{D}\tens{X}^{\intercal},
\end{equation*}
where \(\tens{X}\in\mathbb{R}^{p\times N}\) is the feature matrix, \(N\) is the number of samples, \(p\) is the feature dimension, and \(\tens{D} = \operatorname{diag}(D_{11}, \dots, D_{NN})\) is a diagonal matrix of curvature weights. This structure is a natural test case for deformed decompression, since the spectrum depends not only on the feature covariance but also on the empirical law of the diagonal weights in \(\tens{D}\).

We generated two Hessian ensembles with different ambient sizes. The first has \(N=\num{100000}\) samples and \(p=\num{32000}\) features, with results given in \Cref{sec:results}. The second, reported here, has \(N=\num{120000}\) samples and \(p=\num{16000}\) features. For both experiments, the columns \(\vect{x}_i\) of \(\tens{X}\) were sampled with zero-mean Gaussian entries, with half of the features having variance \(1\) and the other half variance \(4\). Ground-truth weights \(\vect{w}^{\star}\in\mathbb{R}^p\) were sampled from a standard normal distribution and used to define the separating hyperplane \(\{\vect{x} \mid \vect{x}^{\intercal} \vect{w}^{\star}=0\}\). Labels were first generated as \(y_i = \mathbf{1}_{\{\vect{x}_i^{\intercal} \vect{w}^{\star}>0\}}\), and then \(10\%\) of the labels were randomly flipped.

In both experiments, the trainable weights \(\vect{w}_0\) were initialized as \(\mathcal{N}(\tens{0}, 10^{-2}\tens{I})\). We trained the logistic regression model with cross-entropy loss using the Adam optimizer with learning rate \(0.1\), zero weight decay, and \(2000\) iterations. After training, the predictive probabilities were computed from the fitted logits, and the diagonal matrix \(\tens{D}\in\mathbb{R}^{N\times N}\) was formed with entries \(D_{ii} = \sigma_i (1 - \sigma_i)\). The Hessian matrix was then explicitly formed as \(\tens{H} = N^{-1} \tens{X} \tens{D} \tens{X}^{\intercal}\), giving a \(p \times p\) matrix. Although the curvature weights in $\tens{D}$ depend on the same data used to form $\tens{X}$, \citet{LIAO-2021a} show that, under their Gaussian model assumptions with fixed model parameters, this dependence does not alter the limiting spectral distribution of the corresponding independent-weight model. Here, the model parameters are learned from the same data, so we assess the accuracy of DD empirically.

\begin{figure}[!t]
    \centering
    \includegraphics[width=0.4\textwidth]{\figdir/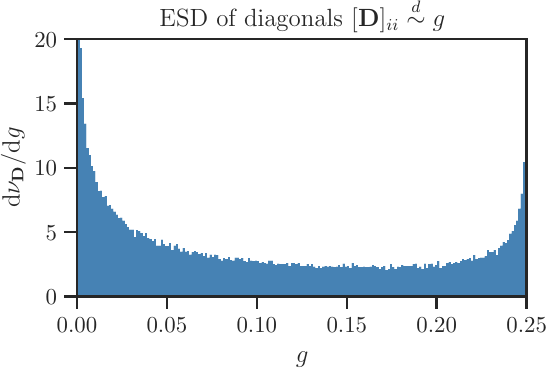}
    \caption{Empirical distribution of the diagonal curvature weights \(D_{ii}\) for the logistic-regression Hessian model. These weights are samples of the random curvature \(g=\ell''(y,h)\) in the generalized Hessian model of \citet{LIAO-2021a}. The Hessian is \(\tens{H} = N^{-1}\tens{X}\tens{D}\tens{X}^{\intercal}\), with \(N=\num{120000}\) samples and feature dimension \(p = \num{16000}\).}
    \label{fig:hessian-diag}
\end{figure}

\Cref{fig:hessian-diag} shows the empirical distribution of the diagonal curvature weights \(D_{ii}\). These weights are the samples of the random curvature term \(g=\ell''(y,h)\) in the generalized Hessian model of \citet{LIAO-2021a}. The distribution is nonconstant, so the Hessian is not merely a rescaled sample covariance matrix; rather, its spectrum reflects the interaction between the feature covariance and the curvature profile induced by the trained logistic model.

\begin{figure}[!t]
    \centering
    \includegraphics[width=\textwidth]{\figdir/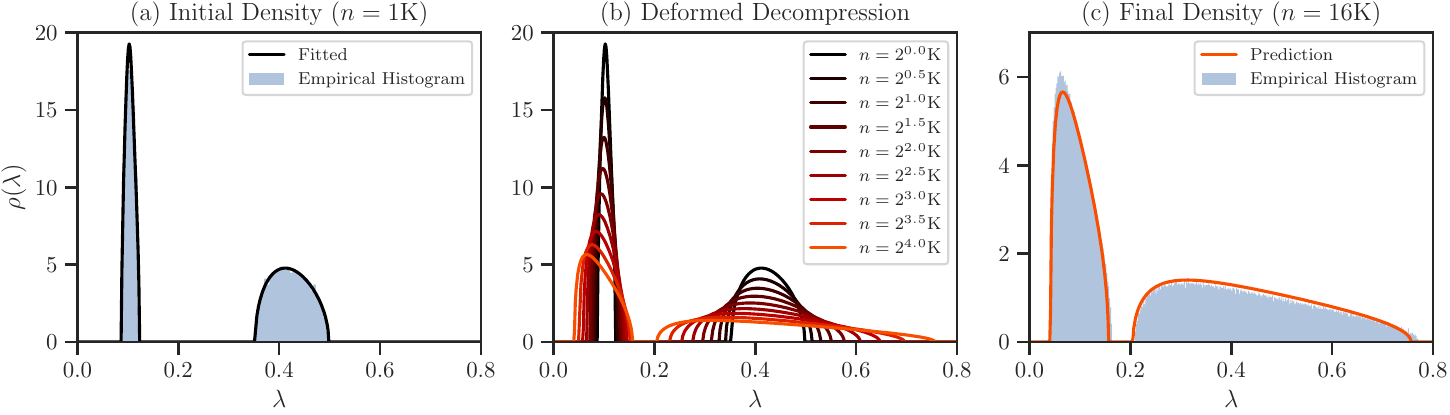}
    \caption{Deformed-decompression reconstruction of the Hessian spectrum for the logistic-regression model with target matrix size \(n = \num{16000}\). (a) ESD of the sampled \(n_0 \times n_0\) principal submatrix with \(n_0=\num{1000}\) (histogram), together with the fitted algebraic density (solid curve). (b) DD evolution from \(n_0\) to \(n\). (c) Final DD-predicted density at size \(n = \num{16000}\), compared with the empirical spectrum obtained by direct eigendecomposition of the full Hessian.}
    \label{fig:hessian-flow}
\end{figure}

\begin{figure}[!t]
    \centering
    \includegraphics[width=0.65\textwidth]{\figdir/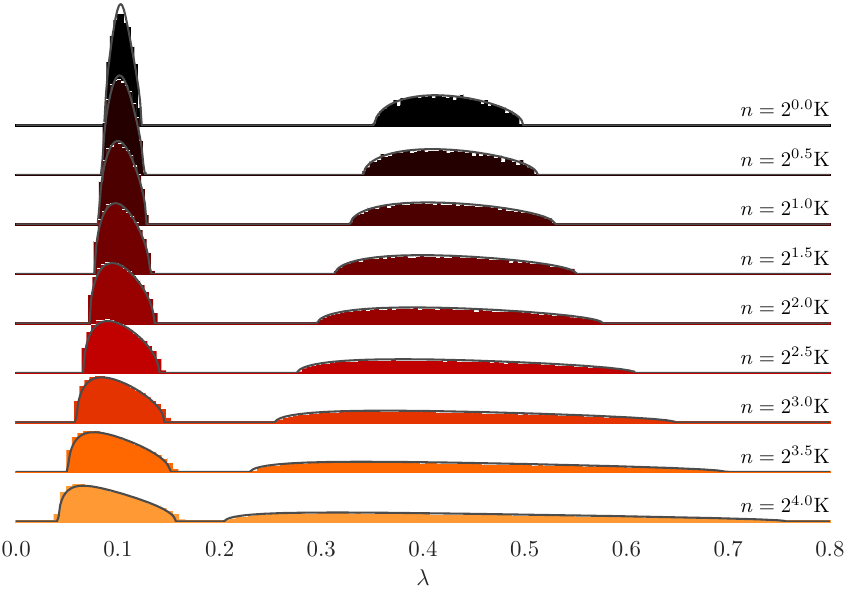}
    \caption{Ridge-plot view of the deformed-decompression evolution in \Cref{fig:hessian-flow}. Each row corresponds to an intermediate matrix size; solid curves show the DD-predicted spectral densities, while filled histograms show the empirical spectra of sampled submatrices at the same sizes. The stacked view highlights how the full density evolves from the \(n_0 = \num{1000}\) submatrix to \(n = \num{16000}\)}
    \label{fig:hessian-ridge}
\end{figure}

For the DD reconstruction, the evolving matrix size is the Hessian dimension. Thus, in \Cref{fig:hessian-flow,fig:hessian-ridge}, \(n=p=\num{16000}\), while \(n_0=\num{1000}\) denotes the sampled principal-submatrix size. \Cref{fig:hessian-flow} shows the deformed-decompression reconstruction for the \(n = \num{16000}\) Hessian. Starting from the spectrum of a sampled \(n_0 = \num{1000}\) principal submatrix, we fit an algebraic spectral curve and evolve it to the target size \(n = \num{16000}\). The initial fitted density, the intermediate DD flow, and the final comparison with the directly computed empirical spectrum show that the predicted density tracks the full-size Hessian spectrum accurately.

\Cref{fig:hessian-ridge} presents the same evolution as a ridge plot, making the gradual deformation of the spectral density across matrix sizes visible. This view emphasizes that DD does not simply shift the density, but evolves the full spectral shape as the effective aspect ratio changes.

Finally, \Cref{tab:hessian-complexity} reports runtime and accuracy metrics for the \(N=\num{100000}\), \(n=p=\num{32000}\) Hessian example described in \Cref{sec:results}. Direct eigendecomposition becomes rapidly more expensive as the matrix size grows, while DD only requires the eigendecomposition of the initial \(2^0\mathrm{K}\)-sized submatrix followed by a comparatively inexpensive spectral evolution. The distributional distances, moment errors, and edge errors remain small across the tested range, indicating that DD captures both the bulk density and the lower-dimensional spectral quantities of the Hessian.

\begin{table}[t!]
\caption{Comparison of direct eigendecomposition with deformed decompression (DD) for the \(n = \num{32000}\) logistic-regression Hessian matrix described in \Cref{sec:results}. The Direct column reports the process time for computing the eigenvalues at each matrix size. In the DD column, the first term is the time to compute eigenvalues of the initial \(2^{0}\mathrm{K}\)-sized submatrix, and the second term is the DD evolution time. Distributional distances compare the empirical spectrum with the DD reconstruction, with \(W_1/L\) denoting the Wasserstein-1 distance normalized by the spectral range \(L\), and MMD denoting the Gaussian-kernel maximum mean discrepancy on eigenvalues. Moment errors are relative errors of the first moment and standard deviation. Edge errors compare empirically detected spectral edges with DD-evolved edges. All reported errors are shown as percentages.}
\label{tab:hessian-complexity}
\vspace{1mm}
\centering
\begin{tabular}{lrrrrrrrr}
    \toprule
    Size & \multicolumn{2}{c}{Process Time (sec)} & \multicolumn{2}{c}{Distributional Distance} & \multicolumn{2}{c}{Moments Rel. Error} & \multicolumn{2}{c}{Edges Error} \\
    \cmidrule(lr){2-3} \cmidrule(lr){4-5} \cmidrule(lr){6-7} \cmidrule(lr){8-9}
    \(n\) & Direct & DD (ours) & \(W_1/L\) & MMD & \(\Delta \mu_1 /\mu_{1}\) & \(\Delta \sigma / \sigma\) & Mean & Max \\
    \midrule
    \(2^{0}\)K & \(    3.6\) & \(3.6 +  \phantom{0}0.0\) & \(0.00\%\) & \(0.04\%\) & \(0.00\%\) & \(0.03\%\) & \(0.1\%\) & \(0.2\%\) \\
    \(2^{1}\)K & \(   13.1\) & \(3.6 +  \phantom{0}1.0\) & \(0.02\%\) & \(1.01\%\) & \(2.15\%\) & \(0.07\%\) & \(0.1\%\) & \(0.3\%\) \\
    \(2^{2}\)K & \(   75.3\) & \(3.6 +  \phantom{0}2.3\) & \(0.03\%\) & \(1.19\%\) & \(2.86\%\) & \(0.05\%\) & \(0.2\%\) & \(0.5\%\) \\
    \(2^{3}\)K & \(  576.7\) & \(3.6 +  \phantom{0}4.7\) & \(0.05\%\) & \(1.42\%\) & \(4.36\%\) & \(0.45\%\) & \(0.4\%\) & \(1.1\%\) \\
    \(2^{4}\)K & \( 4221.8\) & \(3.6 +  \phantom{0}9.2\) & \(0.05\%\) & \(1.19\%\) & \(4.14\%\) & \(0.37\%\) & \(1.3\%\) & \(2.5\%\) \\
    \(2^{5}\)K & \(27897.1\) & \(3.6 +            18.6\) & \(1.31\%\) & \(5.63\%\) & \(4.24\%\) & \(0.47\%\) & \(2.8\%\) & \(5.4\%\) \\
    \bottomrule
\end{tabular}
\end{table}


\subsection{STL-10 Image Dataset} \label{sec:stl10}

We next consider an empirical Gram covariance matrix formed from natural images in the STL-10 dataset \citep{COATES-2011}. STL-10 contains \(96 \times 96\) RGB images and includes a large unlabeled split, making it a convenient source of real high-dimensional observations for testing deformed decompression. We vectorize each image into a row vector of length \(p = 96 \times 96 \times 3 = \num{27648}\), preprocess the rows by removing per-image brightness and contrast, and then standardize the resulting features across the sampled images.

Let \(\tens{X} \in \mathbb{R}^{n \times p}\) denote the resulting data matrix. We form the sample Gram covariance matrix \(\tens{A} \coloneqq \frac{1}{p} \tens{X} \tens{X}^{\intercal}\). This example differs from the closed-form deformed Marchenko--Pastur model in \Cref{sec:deformed-mp}: the underlying population law is not known analytically and the fitted algebraic curve is obtained directly from the empirical spectrum of a smaller principal submatrix. Starting from \(n_0 = \num{1000}\), we apply deformed decompression to predict the spectral density at larger matrix sizes up to \(n = \num{8000}\).

\Cref{tab:stl-complexity} summarizes the corresponding runtime and distributional errors. The direct baseline computes the empirical spectrum separately at each matrix size, while DD uses only the spectrum of the initial \(n_0 = 2^{0}\mathrm{K}\) submatrix and then evolves the fitted algebraic curve to larger aspect ratios. Because the DD reconstruction is a continuous density and the empirical eigenvalues form a discrete measure, we compare them using probability metrics rather than pointwise density errors. The normalized Wasserstein distance \(W_1/L\) measures discrepancy on the log-spectral axis without requiring kernel density estimation, while MMD provides a complementary kernel-based comparison of the same log-eigenvalue distributions \cite{GIBBS-2002,PANARETOS-2019}. Across the tested sizes, DD substantially reduces the cost relative to direct eigendecomposition while maintaining small \(W_1/L\), MMD, and first-moment errors.

\begin{figure}[t!]
    \centering
    \includegraphics[width=0.47\textwidth]{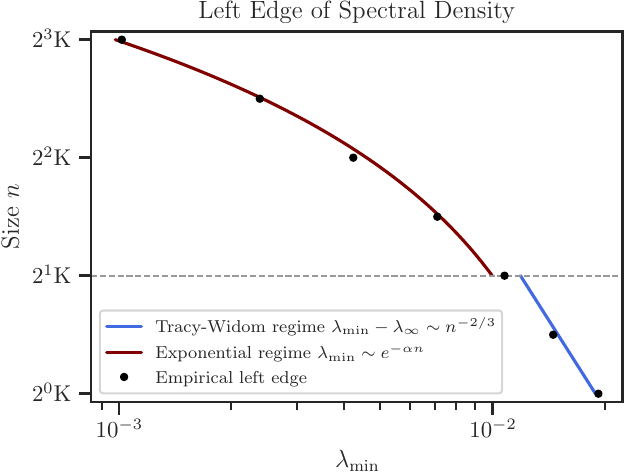}
    \caption{Empirical (black) smallest eigenvalue $\lambda_{\min}$ of the STL-10 image dataset example, with Tracy--Widom (blue) predicted decay at rate $\asymp n^{-2/3}$ and exponential (red) predicted decay at rate $\asymp e^{-\alpha n}$ for fitted $\alpha$. }
    \label{fig:stl-left-edge}
\end{figure}

The behavior of the left edge across matrix sizes $n$ is plotted in \Cref{fig:stl-left-edge}. It is particularly interesting that the same Tracy-Widom-then-exponential scaling regime described in \citet[Appendix H]{AMELI-2026} is present here also. We remark that the justification provided there, in the form of \citet[Proposition H.2]{AMELI-2026}, is not dependent on the free decompression operation, and describes a universal left-edge behavior for positive-definite matrices where $\lambda_{\min}$ is approaching zero.


\section{Implementation and Reproducibility Guide} \label{app:software}

All numerical procedures in this paper are provided through the \texttt{\freealg.AlgebraicForm} interface in the open-source Python package \texttt{\freealg}.\footnote{\texttt{\freealg} is available for installation from PyPI (\FreeAlgPypiUrl), the documentation can be found at \FreeAlgDocUrl, and the source code is available at \FreeAlgGithubUrl.} The package includes tools for fitting algebraic spectral curves from empirical spectra, evaluating the associated Stieltjes transform, evolving densities under deformed decompression, and tracking lower-dimensional spectral quantities such as edges and cusps. It also provides closed-form free-probability distributions, including deformed Marchenko--Pastur laws, which are useful for controlled validation experiments.

\Cref{list:freealg1} gives a minimal setup for the synthetic deformed Marchenko--Pastur experiment used in this paper. The example constructs the distribution, visualizes its spectral-curve branches, generates a large matrix realization, and extracts a smaller principal submatrix. \Cref{list:freealg2} then applies the \texttt{\freealg.AlgebraicForm} workflow to the smaller matrix: it fits an algebraic curve, reconstructs the initial density, evolves the density to a larger matrix size using deformed decompression  by setting \texttt{kind='deformed'}, and tracks the corresponding edges and cusp points. Further details on the package interface, supported input formats, and Stieltjes-transform fitting routines are given in \citet{AMELI-2025b} and \citet{AMELI-2026}.

The scripts and notebooks used to generate the figures and tables in this paper are included in the code repository. Additional details on function arguments, class options, and reproducibility settings are provided in the package documentation. The experiments were run on a consumer-grade workstation with an AMD Ryzen 7 5800X processor, 128GB DDR4 RAM, and Fedora Linux.

\begin{lstlisting}[
    style=mystyle,
    language=Python,
    caption={Synthetic deformed Marchenko-Pastur setup using \texttt{\freealg}. The distribution object is used only to generate a controlled matrix example and visualize the associated spectral-curve branches.},
    label={list:freealg1},
    float=ht!]
# Install ;#\freealg#; with ;#"\texttt{pip install \freealg}"#;
import freealg as fa
import numpy as np

# Create an object for the deformed Marchenko--Pastur distribution (see ;# \Cref{sec:deformed-mp}#;)
dist = fa.distributions.DeformedMarchenkoPastur(t=[2.0, 5.5], w=[0.75, 0.25], c=0.1)

# Visualize all branches and their gluing as in ;# \Cref{fig:dmp-branches} #;
x = np.linspace(-1, 9, 300)
y = np.linspace(-3, 3, 300)
dist.plot_branches(x, y)

# Generate a matrix realization of size ;#$ n=\num{8000} $#; for this distribution
A_large = dist.matrix(8000, seed=0)

# Sample the above matrix to a smaller submatrix of size ;#$ n_0 = \num{1000} $#;
A_small = fa.submatrix(A_large, 1000, seed=0)
\end{lstlisting}

\ifjournal
    \clearpage
\else
    \clearpage
\fi

\begin{lstlisting}[
    style=mystyle,
    language=Python,
    caption={Deformed decompression workflow using \texttt{\freealg.AlgebraicForm}. Starting from the smaller matrix \(A_{\mathrm{small}}\), the code fits an algebraic spectral curve, estimates the density, decompresses to a larger matrix size, and tracks edges and cusps.},
    label={list:freealg2},
    float=ht!]
# Create an algebraic-form object for the small matrix
ratio = dist.ratio * (A_small.shape[0] / A_large.shape[0])
af = fa.AlgebraicForm(A_small, ratio=ratio, support=None)

# Fit the polynomial ;#$ P $#; with ;#$ \deg_m(P) = 3 $#; and ;#$ \deg_z(P) = 2 $#;
af.fit(deg_m=3, deg_z=2)

# Estimate the density of the small matrix from ;#$ P $#;
x = np.linspace(0, 10, 200)
rho_small = af.density(x, plot=True)

# Deformed decompression of density and atoms corresponding to a larger matrix of
# size ;#$ n = \num{8000} $#; using ;# \Cref{alg:deform} #; (similar to ;# \Cref{fig:dmp-flow} #;). 
rho_large = af.decompress(size=8000, x=x, kind='deformed', plot=True)

# Evolve spectral edges similar to ;# \Cref{fig:dmp-atom}#; using ;# \Cref{alg:edge} #;
# This also finds cusp points, if any occur.
sizes = np.linspace(1000, 8000, 100)
t = np.log(sizes / 1000)
edges, cusps = af.edge(t, kind='deformed')
\end{lstlisting}


\end{appendices}

\end{document}